%% file: freq_sens_arxiv.tex
\documentclass[11pt
]{article} 
\usepackage[letterpaper,margin=1in,footskip=0.25in]{geometry}
\usepackage{stix2}

\usepackage[utf8]{inputenc} 
\usepackage[T1]{fontenc}    
\usepackage{hyperref}       
\usepackage{url}            
\usepackage{longtable,booktabs}       
\usepackage{nicefrac}       
\usepackage{microtype}      
\usepackage{xcolor}         
\usepackage{tabularx}

\usepackage{mathtools,amssymb,amsthm,amsfonts}
\usepackage[nameinlink]{cleveref}
\usepackage{tikz-cd,extarrows}
\usepackage{graphicx}
\usepackage{subcaption}

\usepackage[shortlabels,inline]{enumitem}
\usepackage[style=alphabetic,sortcites]{biblatex}
\usepackage{listings}
\hypersetup{
colorlinks=true,
bookmarks=true,
bookmarksdepth=2
}
\usetikzlibrary{decorations.pathmorphing}
\usepackage{array}

\input{cgmacros.tex}

\newcommand{\saveforarxiv}[1]{#1}

\title{
Testing predictions of representation cost theory with CNNs}

\author{Charles Godfrey, Elise Bishoff, Myles Mckay, Davis Brown, \\
Grayson Jorgenson, Henry Kvinge \& Eleanor Byler 
\\
Pacific Northwest National Lab\\
\texttt{\{first\}.\{last\}@pnnl.gov} \\
}
\date{}

\begin{document}

\maketitle
\begin{abstract}
   \input{abstract.tex}
\end{abstract}

\input{main_body.tex}

\section{Acknowledgements} 
The research described in this paper was conducted under the Laboratory
Directed Research and Development Program at Pacific Northwest National
Laboratory, a multiprogram national laboratory operated by Battelle for the U.S.
Department of Energy.

We are grateful to Bobak Kiani for pointing out the reference to
\cite{bhatiaMatrixAnalysis1996} used in the proof of
\cref{lem:non-com-generalized-holder}, and to Sara Fridovich-Keil, Bhavya
Kailkhura and Hannah Lawrence and Daniel Soudry for helpful discussions. We
thank Michael Rawson for reviewing an earlier draft of this paper.

\printbibliography
\appendix

\input{appendix.tex}

\end{document}

%% file: cgmacros.tex
\newcommand{\RR}{\mathbb{R}}
\newcommand{\CC}{\mathbb{C}}
\newcommand{\NN}{\mathbb{N}}
\newcommand{\ZZ}{\mathbb{Z}}
\newcommand{\nrm}[1]{\lvert #1 \rvert}
\newcommand{\nnrm}[1]{\lVert #1 \rVert}

\newcommand{\cD}{\mathcal{D}}
\newcommand{\cL}{\mathcal{L}}

\DeclareMathOperator{\relu}{ReLU}
\DeclareMathOperator{\diag}{diag}

\DeclareMathOperator{\sign}{sign}

\numberwithin{equation}{section}

\theoremstyle{plain}
\newtheorem{theorem}[equation]{Theorem}
\newtheorem{lemma}[equation]{Lemma}
\newtheorem{corollary}[equation]{Corollary}
\newtheorem{proposition}[equation]{Proposition}

\theoremstyle{definition}
\newtheorem{definition}[equation]{Definition}

\theoremstyle{remark}
\newtheorem{remark}[equation]{Remark}
\newtheorem{remarks}[equation]{Remarks}

%% file: abstract.tex
It is widely acknowledged that trained convolutional neural networks (CNNs) have
different levels of sensitivity to signals of different frequency. In
particular, a number of empirical studies have documented CNNs sensitivity to
low-frequency signals. In this work we show with theory and experiments that
this observed sensitivity is a consequence of the frequency distribution of
natural images, which is known to have most of its power concentrated in
low-to-mid frequencies. Our theoretical analysis relies on representations of
the layers of a CNN in frequency space, an idea that has previously been used to
accelerate computations and study implicit bias of network training algorithms,
but to the best of our knowledge has not been applied in the domain of model
robustness.

%% file: main_body.tex



\section{Introduction}
\label{sec:intro}

Since their rise to prominence in the early 1990s, convolutional neural networks
(CNNs) have formed the backbone of image and video recognition, object detection,
and speech to text systems
(\cite{fukushimaNeocognitronSelforganizingNeural1980,lecunBackpropagationAppliedHandwritten1989,krizhevskyImageNetClassificationDeep2012,collobertWav2LetterEndtoEndConvNetbased2016,karpathyLargescaleVideoClassification2014}). The success of CNNs has largely been
attributed to their "hard priors" of spatial translation invariance and local
receptive fields \cite[\S 9.3]{Goodfellow-et-al-2016}. On the other hand, more
recent research has revealed a number of less desirable and potentially
data-dependent biases of CNNs, such as a tendency to make predictions on the
basis of texture features (\cite{geirhos2018imagenettrained}). Moreover, it has
been repeatedly observed that CNNs are sensitive to perturbations in targeted
ranges of the Fourier frequency spectrum
(\cite{guoLowFrequencyAdversarial2019,sharmaEffectivenessLowFrequency2019}) and
further investigation has shown that these frequency ranges are dependent on
training data (\cite{yin2019,berhnard2021,abello2021,maiya2022a}). In this work, we provide a
\emph{mathematical explanation} for these frequency space phenomena, showing with theory and
experiments that neural network training causes CNNs to be most sensitive to
frequencies that are prevalent in the training data distribution.

Our theoretical results rely on representing an idealized CNN in frequency
space, a strategy we borrow from \cite{Gunasekar2018}. This representation is
built on the classical convolution theorem,
\begin{equation}
   \label{eq:conv-to-ptwise-mlt}
   \widehat{w \ast x} = \hat{w} \cdot \hat{x}
\end{equation}
where \(\hat{x}\) and $\hat{w}$ denote the Fourier transform of \(x\) and $w$
respectively, and \(\ast \) denotes a convolution. Equation
\ref{eq:conv-to-ptwise-mlt} demonstrates that a Fourier transform converts
convolutions into products. As such, in a ``cartoon'' representation of a CNN in
frequency space, the convolution layers become \emph{coordinate-wise
multiplications} (a more precise description is presented in
\cref{sec:DFT-CNN}). This suggests that in the presence of some form of weight
decay, the weights \(\hat{w}\) for high-power frequencies in the training data
distribution will grow during training, while weights corresponding to low-power
frequencies in the training data will be suppressed. The resulting uneven
magnitude of the weights \(\hat{w}\) across frequencies can thus account for the
observed uneven perturbation-sensitivity of CNNs in frequency space. We
formalize this argument for linear CNNs (without biases) in
\cref{sec:DFT-CNN,sec:reg-CNN-training}.

One interesting feature of the framework set up in \cref{sec:reg-CNN-training}
is that the discrete Fourier transform (DFT) representation of a linear CNN is
\emph{precisely} a feedforward network with block diagonal weight matrices,
where each block corresponds to a frequency index. We show in
\cref{thm:generalized-matrix-holder} that a learning objective for such a
network of depth \(L\) with an \(\ell_2\)-norm penalty on weights is equivalent
to an objective for the associated linear model with an \(\ell_p\) penalty on
the singular values of each of its blocks, i.e. each frequency index --- this
result is new for CNNs with multiple channels and outputs. In
particular, the latter penalty is highly sparsity-encouraging, suggesting as
depth increases these linearly-activated CNNs have an even stronger incentive to
prioritize frequencies present in the training data.


It has long been known that the frequency content of natural images is
concentrated in low-to-mid frequencies, in the sense that the power in Fourier
frequency \(f\) is well-described by \(1/|f|^\alpha\) for a coefficient \(\alpha
\approx 1\) (\cite{leeOcclusionModelsNatural2001}). Hence, when specialized to
training data distributions of natural images, our results explain findings that
CNNs are more susceptible to low frequency perturbations in practice
(\cite{guoLowFrequencyAdversarial2019,sharmaEffectivenessLowFrequency2019}). 

\begin{figure}
   \centering
   \begin{subfigure}{0.8\linewidth}
      \centering
      \includegraphics[width=\linewidth]{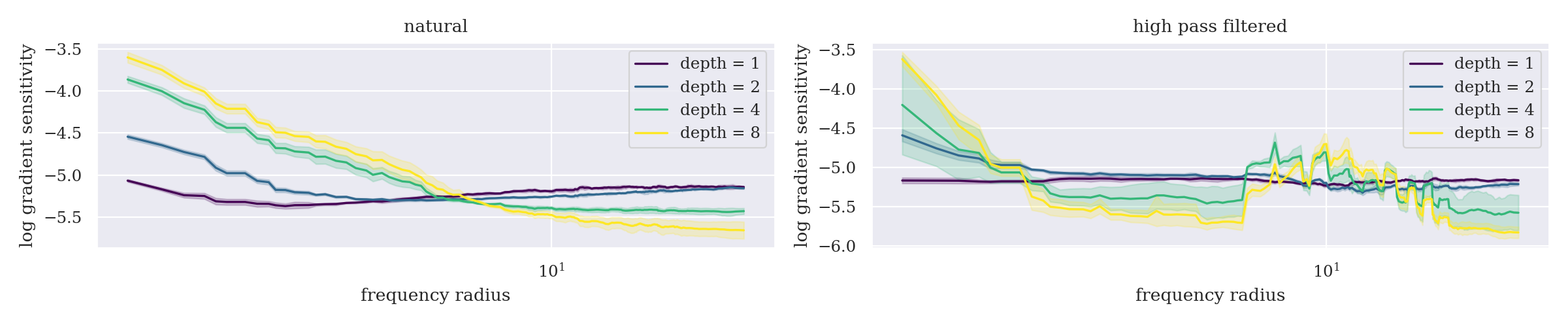}
      \caption{\footnotesize ConvActually models of varying depth.}
   \end{subfigure}
   \begin{subfigure}{0.8\linewidth}
      \centering
      \includegraphics[width=\linewidth]{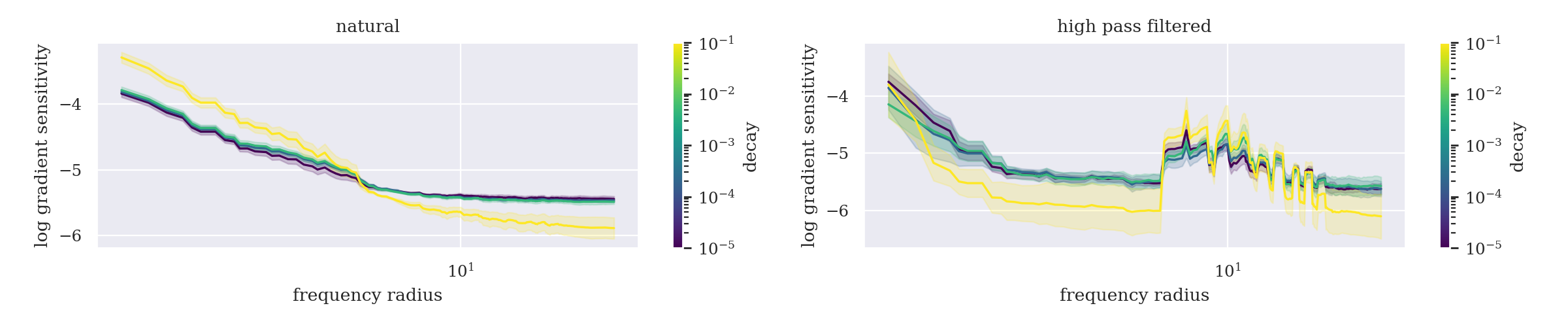}
      \caption{\footnotesize ConvActually models of depth 4 trained with varying weight decay.}
   \end{subfigure}
   \begin{subfigure}{0.8\linewidth}
      \centering
      \includegraphics[width=\linewidth]{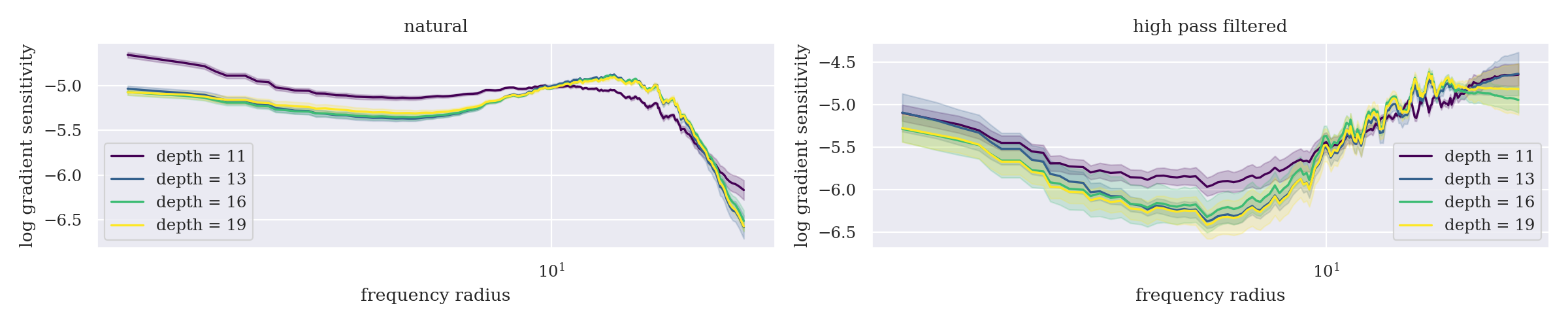}
      \caption{\footnotesize VGG models of varying depth.}
   \end{subfigure}
   \begin{subfigure}{0.8\linewidth}
      \centering
      \includegraphics[width=\linewidth]{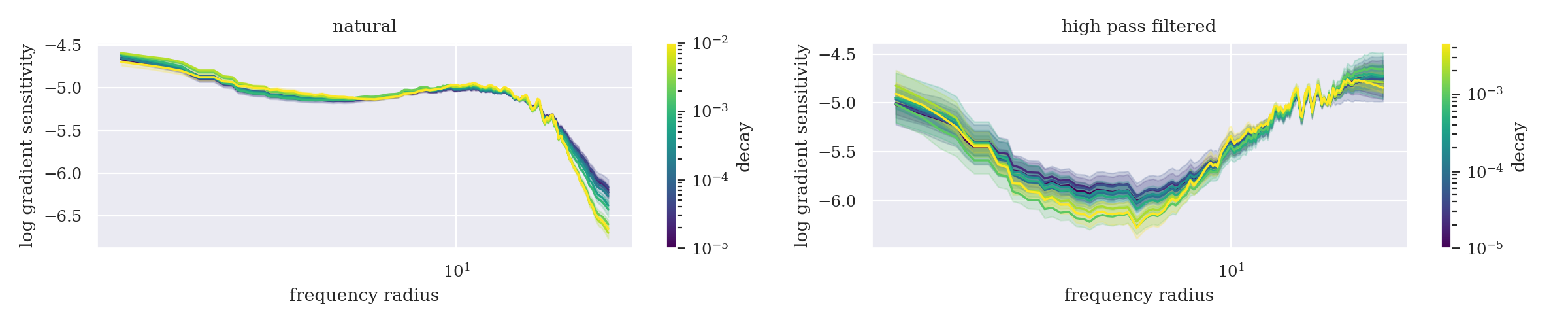}
      \caption{\footnotesize VGG 11 models trained with varying weight decay.}
   \end{subfigure}
   \begin{subfigure}{0.8\linewidth}
      \centering
      \includegraphics[width=\linewidth]{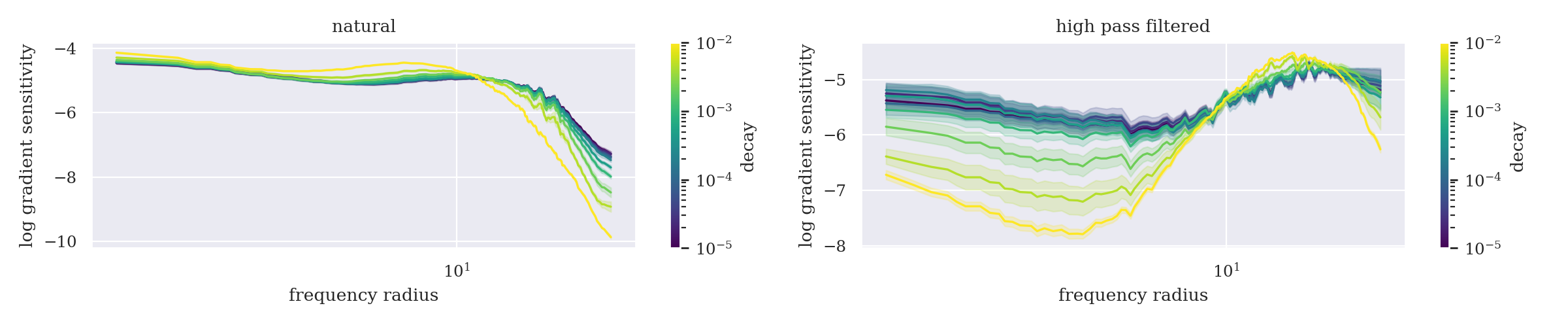}
      \caption{\footnotesize Myrtle CNN models trained with varying weight decay.}
   \end{subfigure}
   \begin{subfigure}{0.4\linewidth}
      \centering
      \includegraphics[width=\linewidth]{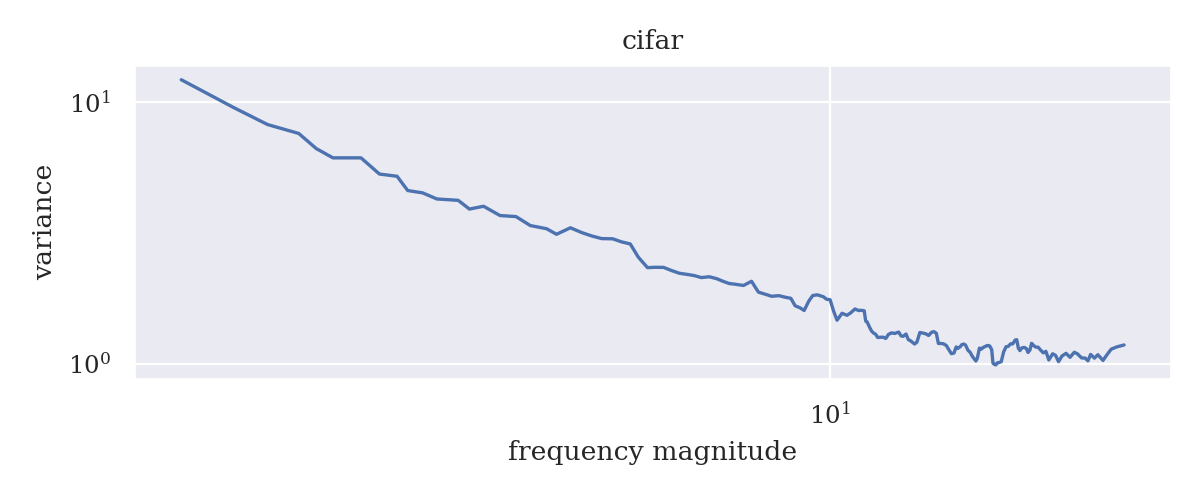}
   \end{subfigure}
   \begin{subfigure}{0.4\linewidth}
      \centering
      \includegraphics[width=\linewidth]{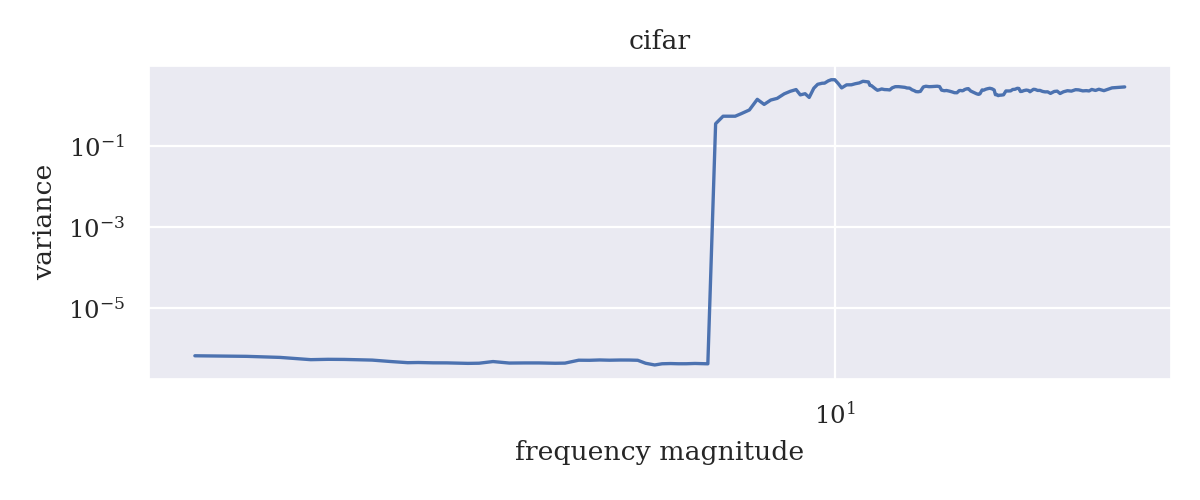}
   \end{subfigure}
   \caption{\footnotesize Radial averages \(E[\nrm{\nabla_{x}f(x)^T \hat{e}_{cij}}\, | \,
   \nrm{(i, j)} = r]\) of frequency sensitivities of CNNs trained on (hpf-)CIFAR10, post-processed by dividing each curve by its
   integral, smoothing by averaging with \(3\) neighbors
   on either side and taking logarithms. \textbf{Bottom row}: frequency statistics of (hpf-)CIFAR10 for comparison. See \cref{sec:experiments} for
   further details.}\label{fig:cifar-experiments}
\end{figure}

We use our theoretical results to derive specific
predictions: CNN frequency sensitivity aligns with the frequency content of
training data, and deeper  models, as well as  models trained with substantial
weight decay, exhibit frequency sensitivity more closely reflecting the
statistics of the underlying images. We confirm these predictions for
\emph{nonlinear} CNNs trained on the CIFAR10 and ImageNette datasets.
\Cref{fig:cifar-experiments} shows our experimental results for a variety of CNN
models trained on CIFAR10 as well as a variant of CIFAR10 preprocessed with high
pass filtering (more experimental details will be provided in
\cref{sec:experiments}).

To the best of our knowledge, ours is the first work to connect the following
research threads (see \cref{sec:rw} for further discussion):
\begin{itemize}[nosep]
   \item equivalences between linear neural networks and sparse linear models
   (implicit bias and representation cost),
   \item classical data-dependent ``shrinkage'' properties of sparse linear models, 
   \item statistical properties of natural images, and
   \item sensitivity of CNNs to perturbations in certain frequency ranges.
\end{itemize}
It seems likely that a similar analysis could provide insight into models
trained on data from a domain other than images; hence our
work may serve as a template for combining results on implicit bias and/or
representation cost with domain knowledge of input data statistics to understand
the behavior of deep learning models. 


\section{Related work}
\label{sec:rw}

\textbf{CNN sensitivity to Fourier frequency components}: \cite{jo} computed transfer accuracy of image classifiers trained on data
preprocessed with various Fourier filtering schemes (e.g. train on low pass
filtered images, test on unfiltered images or vice versa). They found
significant generalization gaps, suggesting that models trained on images with
different frequency content learned different patterns.

\cite{guoLowFrequencyAdversarial2019} proposed algorithms for generating
adversarial perturbations constrained to low frequency Fourier components,
finding that they allowed for greater query efficiency and higher
transferability between different neural networks.
\cite{sharmaEffectivenessLowFrequency2019} demonstrated empirically that
constraining to high or or midrange frequencies did \emph{not} produce similar
effects, suggesting convolutional networks trained on natural images exhibit a particular sensitivity to
low frequency perturbations.

\cite{yin2019} showed different types of corruptions of natural images (e.g.
blur, noise, fog) have  different effects when viewed in frequency space, and
models trained with different augmentation strategies (e.g. adversarial
training, gaussian noise augmentation) exhibit different sensitivities to
perturbations along Fourier frequency components. \cite{diffenderfer2021a}
investigates the relationship between frequency sensitivity and natural
corruption robustness for models compressed with various weight pruning
techniques, and introduces ensembling algorithms where the frequency statistics of a test image are compared to those of various image
augmentation methods, and models trained on the augmentations most
spectrally similar to the test image are used in inference. \cite{fouriermix}
designs an augmentation procedure that explicitly introduces variation in both
the amplitude and phase of the DFT of input images, finding it improves
certified robustness and robustness and common corruptions.
\cite{berhnard2021} investigated the extent to which constraining models to use
only the lowest (or highest) Fourier frequency components of input data provided
perturbation robustness, also finding significant variability across datasets.
\cite{abello2021} tested the extent to which CNNs relied on various frequency
bands by measuring model error on inputs where certain frequencies were removed,
again finding a striking amount of variability across datasets.
\cite{maiya2022a} analyzed the sensitivity of networks to perturbations in
various frequencies, finding significant variation across a variety of datasets
and model architectures. All of these works suggest that model frequency
sensitivity depends heavily on the underlying training data.
--- our work began as an attempt to explain this phenomenon mathematically.

\textbf{Implicit bias and representation cost of CNNs}: Our analysis of
(linear) convolutional networks leverages prior work on implicit bias and
representational cost of CNNs, especially \cite{Gunasekar2018}. There it was
found that for a linear one-dimensional convolutional network where inputs and
all hidden layers have one channel (in the notation of \cref{sec:DFT-CNN}, \(C = C_1 = \dots
=C_{L-1} = K= 1\)) trained on a binary linear
classification task with exponential loss, with linear effective predictor
\(\beta\), the Fourier transformed predictor \(\hat{\beta}\)  converges in
direction to a first-order stationary point of an optimization problem of the
form
\begin{equation}
   \min_{\hat\beta} \frac{1}{2} \nrm{\hat{\beta}}_{2/L} \text{  such that  } y^n \hat{\beta}^T x^n \geq 1 \text{  for all  } n.
\end{equation}
A generalization to arbitrary group-equivariant CNNs (of which
the usual CNNs are a special case) appears in
\cite[Thm. 1]{lawrenceImplicitBiasLinear2022} --- while we suspect that some of
our results generalize to more general equivariant networks we leave that to
future work. For generalizations in different directions see
\cite{lyuGradientDescentMaximizes2020b,yunUnifyingViewImplicit2021}, and for
additional follow up work see \cite{pmlr-v178-jagadeesan22a}. Our
general setup in \cref{sec:DFT-CNN} closely follows these authors', and our
\cref{thm:generalized-matrix-holder} partially confirms a suspicion of \cite[\S
6]{Gunasekar2018} that  ``with multiple outputs, as more layers are added, even
fully connected networks exhibit a shrinking sparsity penalty on the singular
values of the effective linear matrix predictor ...''

While the aforementioned works study the \emph{implicit} regularization imposed
by gradient descent, we instead consider \emph{explicit} regularization imposed
by auxiliary \(\ell_2\) norm penalties in objective functions, and prove
equivalences of minimization problems. In this sense our analysis is 
more closely related to that of \cite{dai2021representation}, which considers
parametrized families of functions \(f(x, w)\) and defines the
\emph{representation cost} of a function \(g(x)\) appearing in the parametric
family as 
\begin{equation}
   R(g) := \min_{w}\{\nrm{w}_2^2 \, | \, f(x, w) = g(x) \text{ for all } x \}.
\end{equation}
While this approach lacks the intimate connection with the gradient descent
algorithms used to train modern neural networks, it comes with some benefits:
for example, results regarding representation cost are agnostic to the choice of
per-sample loss function (in particular they apply to both squared error and
cross entropy loss). In the case where the number of channels \(C = C_1 = \dots
=C_{L-1} = 1\) (but the number of output dimensions may be \(>1\)),
\cref{thm:generalized-matrix-holder} can be deduced from \cite[Thm.
3]{dai2021representation}.  

\textbf{Data-dependent bias}: while in this paper we focus on \emph{spatial frequency} properties of
image data, there is a large and growing body of work on the impact of frequency
properties of training data more broadly interpreted. \cite{Rahaman2019OnTS}
gave a formula for the \emph{continuous} Fourier transform of a ReLU network
\(f: \RR^n \to \RR\), and showed in a range of experiments that ReLU networks
learn low frequency modes of input data first. \cite{pmlr-v178-xiao22a} proves theoretical
results on low frequency components being learned first for 
networks $f: \prod_i S^{n_i} \to \mathbb{R}$ on products of spheres,
where the role of frequency is played by spherical harmonic indices (see also
\cite{xiaoLearningCurves} for some related results). 

Perhaps the work most closely related to ours is that of \cite{hacohen} on
\emph{principal component (PC)  bias}, where it is shown that rates of convergence in
deep (and wide) linear networks are governed by the spectrum of the input data
covariance matrix.\footnote{They also prove a result for shallow ReLU networks.}
\cite{hacohen} also includes experiments connecting PC bias with spectral bias
(learning low frequency modes first, as in the preceding paragraph) and a
phenomenon known as \emph{learning order consistency}. However, it is worth
noting that in their work there is no explicit theoretical analysis of CNNs and
no consideration of the statistics of natural images in Fourier frequency space.

\textbf{Other applications of Fourier transformed CNNs}:
\cite{Zhu2021GoingDI,Pratt2017FCNNFC,mathieu2014,vasilache2015} all, in one way
or another, leverage frequency space representations of convolutions to
accelerate computations, e.g. neural network training. Since this is not our
main focus, we omit a more detailed synopsis.

\section{The discrete Fourier transform of a CNN}
\label{sec:DFT-CNN}
In this section we fix the notation and structures we will be working with. We define a class of idealized, linear convolutional networks and derive a useful representation of these networks in the frequency space via the discrete Fourier transform.

Consider a linear, feedforward, multichannel 2D-CNN \(f(x)\) of the form
\begin{equation}
   \label{eq:simpleCNN}
   \RR^{C\times H \times W} \xrightarrow{w^1 \ast -} \RR^{C_1 \times H \times W} \xrightarrow{w^2 \ast -} \RR^{C_2 \times H \times W} \xrightarrow[]{w^3\ast -} \cdots \xrightarrow[]{w^{L-1} \ast -} \RR^{C_{L-1} \times H \times W} \xrightarrow{w^{L, T} -} \RR^K
\end{equation}
where \(w^l \ast x\) denotes the convolution operation between tensors \(w^l \in \RR^{C_l \times H \times W \times C_{l-1}}\) and \( x \in \RR^{C_{l-1} \times H \times W} \), defined by
\begin{equation}
   \label{eq:conv-layer}
   (w^l \ast x)_{c i j} = \sum_{m+m' = i, n+ n' =j} \Big(\sum_{d} w^l_{cmnd} x_{dm' n'}\Big)
\end{equation}
and \(w^{L, T} x\) denotes a contraction (a.k.a. Einstein summation) of the tensor  \( w^{L} \in \RR^{K  \times H \times W \times C_{L-1}}\) with  the tensor \(x \in \RR^{C_{L-1} \times H \times W}\) over the last 3 indices (the \((-)^T\) denotes a transpose operation described momentarily). Explicitly,
\begin{equation}
\label{eq:contraction}
    (w^{L, T} x)_k = \sum_{l, m, n} w^{L}_{kmnl} x_{lmn}.
\end{equation}
Thus, the model \cref{eq:simpleCNN} has weights $w_l \in \RR^{C_l \times H \times W \times C_{l-1} }$
for $l = 1, \dots, L-1$ and $w_{L} \in \RR^{K  \times H \times W \times  C_{L-1}}$.
\begin{remarks}
For tensors with at least 3 indices (such as \(x \) and the weights \(w_l\)
above) we will always use the transpose notation \(-^T\) to denote reversing the
second and third tensor indices, which will always be the 2D spatial indices.
For matrices and vectors it will be used according to standard practice. In
\cref{eq:contraction} the transpose ensures that the indices in Einstein sums
move from ``inside to out'' as is standard practice.

Equivalently, \(w^{L,T} x\) can be described as a usual matrix product \( \tilde{w}_{L} \mathrm{vec}(x) \) where \(\mathrm{vec}(x) \) is the vectorization (flattening) of \( x \) and \(    \tilde{w}_{L}\) is obtained by flattening the last 3 tensor indices of \(w_{L} \) (compatibly with those of \(x\) as dictated by \cref{eq:contraction}). Hence it represents a typical ``flatten and then apply a linear layer'' architecture component. Our reason for adopting the tensor contraction perspective is that it is more amenable to the Fourier analysis described below.
\end{remarks}

Note that in this network the number of channels is allowed to vary but the
heights and widths remain fixed, and that we use full \(H \times W\)
convolutions throughout as opposed to the local (e.g. \(3 \times 3\))
convolutions often occurring in practice. 

Given an array \(x \in \RR^{C_l \times H \times W}\), we may consider its
discrete Fourier transform (DFT) \(\hat{x}\), whose entries are computed as
\begin{equation}
   \label{eq:DFT}
   \hat{x}_{cij} = \frac{1}{\sqrt{H W} } \sum_{m, n} x_{cmn} \exp(- \frac{2 \pi \imath}{H} m i - \frac{ 2 \pi \imath}{W} n j).
\end{equation}
Similarly, for an array \(w \in \RR^{C_l  \times H \times W \times C_{l-1}}\)
the DFT \(\hat{w}\) is defined to be
\begin{equation}
   \label{eq:DFT2}
   \hat{w}_{cijd} = \frac{1}{\sqrt{H W} } \sum_{m, n} w_{cmnd} \exp(- \frac{ 2 \pi \imath}{H} m i - \frac{ 2 \pi \imath}{W} n j).
\end{equation}
In what follows the DFT will always be taken with respect to the two spatial dimensions and no others. The mapping \(x \mapsto \hat{x}\) defines an orthogonal linear transformation of
\(\RR^{C_l \times H \times W}\). In addition, it satisfies the following two
properties: 
\begin{lemma}[{Cf. \cite[Lem. C.2]{kiani2022projunn}}]
   \label{lem:parseval-convolution}
   \begin{align}
      \widehat{w \ast x} &= \hat{w} \cdot \hat{x} \text{ for } w \in \RR^{C_l \times H \times W \times C_{l-1} }, x \in \RR^{C_l \times H \times W}  \text{  (convolution theorem) } \label{eq:convthm} \\
      w^T x &= \hat{w}^T \hat{x} \text{  for  } w \in \RR^{C_l \times H \times W \times C_{l-1} }, x \in \RR^{C_{l-1} \times H \times W} \text{  (Parseval's theorem). } \label{eq:parseval}
   \end{align}
\end{lemma}
Explicitly, the products on the right hand sides of
\cref{eq:convthm,eq:parseval} are defined as
\begin{equation}
   (\hat{w} \cdot \hat{x})_{cij} = \sum_d \hat{w}_{cijd} \hat{x}_{dij} \text{  and  } (\hat{w}^T \hat{x})_{c} = \sum_{d, i, j} \hat{w}_{cijd} \hat{x}_{dij} \text{ respectively.}
\end{equation}
\begin{remark}
   In the terminology of CNNs, this says that the DFT converts full \(H \times
   W\) convolutions to  spatially pointwise matrix multiplications, and preserves dot products.
\end{remark}
Our first lemma is a mild generalization of \cite[Lem. 3]{Gunasekar2018}; we defer all proofs to \cref{sec:pfs}.
\begin{lemma}
   \label{lem:DFTofCNN}
   The CNN \(f(x)\) is functionally equivalent to the network
   \(\hat{f}(\hat{x})\) defined as
   \begin{equation}
      \label{eq:simpleCNN-DFT}
      \CC^{C\times H \times W} \xrightarrow{\hat{-}} \CC^{C\times H \times W} \xrightarrow{\hat{w}^1 \cdot -} \CC^{C_1 \times H \times W}  \cdots \xrightarrow[]{\hat{w}^{L-1} \cdot -} \CC^{C_{L-1} \times H \times W} \xrightarrow{\hat{w}^{L,T} -} \CC^K ,
   \end{equation}
   where the first map \(\hat{-}\) denotes the DFT \(x \mapsto \hat{x}\).
\end{lemma}

\section{Regularized CNN optimization problems in frequency space}
\label{sec:reg-CNN-training}

Consider a learning problem described as follows: given a dataset $\cD = \{(x^n,
y^n) \in \RR^{C \times H \times W} \times \RR^K \, | \, n = 1, \dots, N\}$, and
a loss function \(\cL: (\RR^K)^N \times (\RR^K)^N \to \RR\), we seek weights
\(w\) solving the \(\ell_2 \)-regularized  minimization problem
\begin{equation}
   \label{eq:deep-ridg-obj}
   \begin{split}
      &\min_{w} \cL((f(x^n))_{n=1}^N, (y^n)_{n=1}^N) + \lambda \sum_{l} \nrm{w^l}_2^2 \text{ or equivalently given \cref{lem:DFTofCNN},} \\
      &\min_{\hat{w}} \cL((\hat{f}(\hat{x}^n))_{n=1}^N, (y^n)_{n=1}^N) + \lambda \sum_{l} \nrm{\hat{w}^l}_2^2
   \end{split}
\end{equation}
where $f$ and $\hat{f}$ depend on $w$ and $\hat{w}$ respectively.
This setup allows a wide variety of loss functions. In the experiments of \cref{sec:experiments,sec:exp-det} we
consider two important special case: first, supervised-style empirical risk minimization with respect to a
sample-wise loss function \(\ell: \RR^K \times \RR^K \to \RR\) (in our
experiments, cross entropy), where
\begin{equation}
   \cL((f(x^n))_{n=1}^N, (y^n)_{n=1}^N) = \frac{1}{N} \sum_{n=1}^N \ell(f(x^n), y^n)
\end{equation}
Second, \emph{contrastive} losses such as the alignment and uniformity objective
of \cite{pmlr-v119-wang20k}, which encourages the features \(f(x^n)\) to be
closely aligned with their corresponding ``labels'' \(y^n\), and the \emph{set}
of features \(\{f(x^n)\}_{n=1}^N\) to be uniformly distributed on the unit
sphere \(S^{K-1} \subset \RR^K\).

According to \cref{lem:DFTofCNN},
\begin{equation}
   \label{eq:simp-conn}
   \hat{f}(\hat{x}) =  \hat{w}^{L,T}\big( \hat{w}^{L-1} \cdots \hat{w}^1 \cdot \hat{x} \big).
\end{equation}
In the case where the numbers of
channels \(C, C_1, \dots, C_{L-1}\) are all 1 and the number of classes \(K=1\), networks of this form were studied in
\cite{tibshiraniEquivalencesSparseModels} where they were termed ``simply
connected.'' In the case where the the number of classes \(K=1\) but the numbers of channels \(C_l\) may be larger, such networks were studied in \cite{Gunasekar2018}. Of course, \cref{eq:simp-conn} is just an over-parametrized linear
function. We can describe it more succinctly by introducing a new tensor
\(\hat{v} \in \RR^{K \times H \times W \times C} \) such that \(\hat{f} (\hat{x}) = \hat{v}^T \hat{x}\). With a little manipulation of \cref{eq:simp-conn}, we can obtain a formula for \(\hat{v} \) in terms of the \(\hat{w}_l\) for
\(l= 1, \dots, L\).
\begin{lemma}
   \label{lem:bilinearity}
   \begin{equation}
    \label{eq:simp-conn2}
    \hat{v} = \hat{w}^{L,T} \cdot \hat{w}^{L-1}  \cdots \hat{w}^1
\end{equation}
\end{lemma}

\saveforarxiv{\begin{remark}
   One can view \cref{lem:bilinearity} an analogue of a much simpler identity for functions \(f^1, \dots, f^{L}, g\) on the real line:
   \[ \int f^{L}(t)  (f^{L-1} \cdots f^1 \cdot g)(t) \, dt = \int f^{L}(t) \cdot f^{L-1}(t) \cdots f^1(t) \cdot g(t) \, dt = \int (f^{L} \cdot f^{L-1} \cdots f^1)(t) g(t) \, dt \]
\end{remark}}

The following
theorem shows that the regularization term of \cref{eq:deep-ridg-obj}, which
penalizes the \(\ell_2\)-norms of the
factors \(\hat{w}^l\), is equivalent to a penalty using more
sparsity-encouraging norms of
\(\hat{v}\). To state
it we need a definition.

\begin{definition}[{Schatten \(p\)-norms, cf. \cite[\S IV]{bhatiaMatrixAnalysis1996}}]
   \label{def:cp-norm}
   Let \(A =(a_{ij}) \in M(m\times n, \mathbb{C})\) be a matrix and let \(A =
   UDV^T\) be a singular value decomposition of \(A\), where \(U \in U(m), V \in
   U(n)\) are unitary matrices and \(D = \diag(\lambda_i)\) is a non-negative
   diagonal matrix with diagonal entries \(\lambda_1, \dots, \lambda_{\min \{m, n\}} \geq 0\).
   For any \(p>0\) the \textbf{Schatten \(p\)-norm of \(A\)} is
   \begin{equation}
      \label{eq:def-cp-norm}
      \nnrm{A}_p^S = (\sum_{i=1}^{\min \{m, n\}} \nrm{\lambda_i}^p)^{\frac{1}{p}}.
   \end{equation}
\end{definition}

\begin{remark}
   \label{rmk:definitions-agree}
   In the case \(p=2\), the \(2\)-norm of \cref{def:cp-norm} agrees with the
   usual Euclidean \(2\)-norm \((\sum_{i, j} \nrm{a_{ij}}^2)^{\frac{1}{2}}\),
   since left (resp. right) multiplication by a unitary matrix \(U^T\) (resp. \(V\)) preserves
   the Euclidean \(2\)-norms of the columns (resp. rows) of \(A\).
\end{remark}

\begin{theorem}
   \label{thm:generalized-matrix-holder}
   The optimization problem \cref{eq:deep-ridg-obj} is equivalent to an
   optimization problem for \(\hat{v}\) of the form
   \begin{equation}
      \label{eq:deep-ridge-alt}
      \min_{\hat{v}} \cL((\hat{v}^T \hat{x}^n)_{n=1}^N, (y^n)_{n=1}^N) + \lambda L \sum_{i, j}(\nnrm{ \hat{v}_{ij}}_{\frac{2}{L}}^S)^{\frac{2}{L}},
   \end{equation}
   where\footnote{by an abuse of notation for which we beg your forgiveness.}
   \(\hat{v}_{ij} \) denotes the \(K \times C \) matrix obtained by fixing the
   spatial indices of \(\hat{v} = (\hat{v}_{cijd}) \), and the \(\min\) is taken over the space of tensors \(\hat{v}\) such that each matrix \(\hat{v}_{ij} \) has rank at most \(\min\{C, C_1, \dots, C_{L-1}, K \} \).
\end{theorem}
The essential ingredient of our proof is a generalized non-commutative H\"older inequality.
\begin{lemma}
   \label{lem:non-com-generalized-holder}
   If \(B \in M(m \times n, \CC)\) is a matrix with complex entries, \(A_1,
   \dots, A_{k}\)  is a composable sequence of complex matrices such that
   \(A_L \cdots A_1 = B\)  and \(\sum_i
   \frac{1}{p_i} = \frac{1}{r}\) where \(p_1,
   \dots, p_{L}, r >0\) are positive real numbers, then
   \begin{equation}
       \nnrm{B}_{r}^S \leq \prod_i \nnrm{A_i}_{p_i}^S
   \end{equation}
 \end{lemma}
Such inequalities are not new: in the case \(r=1\), \cref{lem:non-com-generalized-holder} follows from
 \cite[Thm. 6]{dixmierFormesLineairesAnneau1953a}, and in the case \( L=2\) it is an exercise in \cite{bhatiaMatrixAnalysis1996}. However, we suspect (and our proof of \cref{thm:generalized-matrix-holder} suggests) that  \cref{lem:non-com-generalized-holder} underpins many existing results on implicit bias and representation costs of (linear) neural networks, such as those of \cite{Gunasekar2018,yunUnifyingViewImplicit2021,dai2021representation,lawrenceImplicitBiasLinear2022,pmlr-v178-jagadeesan22a}.

In the case where the numbers of channels \(C, C_1, \dots, C_{L-1}\) are all
\(1\) and and the number of outputs \(K = 1\), \emph{and} where the loss \(\ell \) is squared error, the
problem \cref{eq:deep-ridge-alt} reduces to
\begin{equation}
   \label{eq:deep-ridge-alt-tibs}
   \min_{\hat{v}} \frac{1}{N} \sum_n \nrm{y^n - \sum_{i, j} \hat{v}_{ij}  \hat{x}_{ij}}_2^2 + \lambda L \sum_{i, j} \nrm{\hat{v}_{ij}}^{\frac{2}{L}}.
\end{equation}
The sum in the regularization term of \cref{eq:deep-ridge-alt-tibs} is
\(\nrm{\hat{v}}_p^p\) where \(p = \frac{2}{L}\) --- in particular when \(L=1\), \cref{eq:deep-ridge-alt-tibs}  is a ridge regression problem (\cite{Hastie2001TheEO,tibshiraniSparsityLassoFriendsa})  and when \(L=2\)
(the one hidden layer case) \cref{eq:deep-ridge-alt-tibs} is a LASSO problem. We
can analyze these two tractable cases to obtain qualitative predictions which
will be tested empirically in \cref{sec:experiments}. Since the qualitative
predictions from both cases are similar, we devote the following section to
ridge, and defer LASSO to \cref{sec:lasso}.

\subsection{\(L = 1\): ridge regression}
\label{sec:1layer}

In this case, \cref{eq:deep-ridge-alt-tibs} is the usual ridge regression
objective (see e.g. \cite{Hastie2001TheEO}); the closed-form solution is
\begin{equation}
   \label{eq:ridge-clsd-frm}
   (\lambda + \frac{1}{N} \hat{X}^T \hat{X}) \hat{v} = \frac{1}{N} \hat{X}^T Y
\end{equation}
where \(\hat{X}\) is a \(N \times C \times H \times W\) batch tensor with ``rows'' the \(\hat{x}^n\) and the entries of \(Y\) are
the \(y^n\) (see e.g. \cite{Hastie2001TheEO}). When \(\lambda = 0\) this reduces to the usual (unpenalized) least
squares solution \(\hat{v}_{\mathrm{LS}} := (\hat{X}^T \hat{X})^{-1}\hat{X}^T Y\), and substituting \(\hat{X}^T Y = \hat{X}^T \hat{X} \hat{v}_{\mathrm{LS}} \) in \cref{eq:ridge-clsd-frm} we obtain
\begin{equation}
   \label{eq:clsdfrm-ridge}
   (\lambda + \frac{1}{N} \hat{X}^T \hat{X}) \hat{v} = \frac{1}{N} \hat{X}^T \hat{X} \hat{v}_{\mathrm{LS}}
\end{equation}
where strictly speaking \(\hat{X}^T \hat{X}\) is a tensor product of \(\hat{X}\)
with itself in which we contract over the batch index of length \(N\), hence it
is of shape \(W \times H\times C\times C \times H \times W\).

The frequency properties of images enter into the structure of the symmetric
tensor \(\frac{1}{N} \hat{X}^T \hat{X}\), which (if the dataset \(\cD\) is
centered, i.e. preprocessed by subtracting the mean \(\frac{1}{N}X^T
\mathbf{1}_N\)) serves as a generalized \emph{covariance matrix}  for the
frequency space representation of \(\cD\). To ease notation, let \(\Sigma =
\frac{1}{N} \hat{X}^T \hat{X}\), and \emph{suppose} that
\begin{equation}
   \label{eq:diagonality-assumption}
   \Sigma_{whcc'h'w'} \approx \begin{cases}
      \tau_{cij} & \text{ if } (c, i, j) = (c', i', j') \\
      0 \text{  otherwise.}
   \end{cases}
\end{equation}
In other words, proper covariances between distinct frequency components are
negligible and we retain only the \emph{variances}, i.e. the diagonal entries of
the covariance matrix.
In \cref{fig:cifar-covar} we demonstrate that this
assumption is not unrealistic in the case where \(X\) is a dataset of natural images.

With the assumption of \cref{eq:diagonality-assumption}, \cref{eq:clsdfrm-ridge} reduces to
\begin{equation}
   \label{eq:shrinkage}
   \hat{v}_{cij} \approx \frac{\tau_{cij}}{\lambda + \tau_{cij}} \hat{v}_{\mathrm{LS}, cij} = \frac{1}{1 + \frac{\lambda}{\tau_{cij}}} \hat{v}_{\mathrm{LS},cij} \text{  for all  } cij.
\end{equation}
\Cref{eq:shrinkage} is an instance of the classic fact that ridge regression
shrinks coefficients more in directions of low input variance. In words,
\(\tau_{cij}\) is the variance of training images in Fourier component \(cij\),
and \cref{eq:shrinkage} says \(\nrm{\hat{v}_{cij}}\) shrinks more when the
variance \(\tau_{cij}\) is low; in the limiting case \(\tau_{cij} \to 0\) the
coefficient \(\hat{v}_{cij} \to 0\) as well.

Returning to the subject of frequency sensitivity, observe that
\(\hat{v}_{cij}\) is the directional derivative of \(f\) with respect to the
\(cij\)-th Fourier component.
\begin{proposition}[{Data-dependent frequency sensitivity, \(L=1\)}]
   \label{prp:data-dep-freq}
   With the notations and assumptions introduced above, the magnitude of the directional
   derivative of \(f\) with respect to the \(cij\)-th Fourier component scales
   with \(\lambda\) and \( \tau\) according to \(\frac{1}{1 + \frac{\lambda}{\tau_{cij}}} \).
\end{proposition}
Empirically it has been found that for natural distributions of images the
variances \(\tau_{cij}\) follow a power law of the form \(\tau_{cij} \approx
\frac{\gamma}{\nrm{i}^\alpha + \nrm{j}^\beta}\)
(\cite{leeOcclusionModelsNatural2001,baradad2021learning}).\footnote{In particular, under this approximation \(\tau_{cij}\) is independent of \(c\).} Under this model, \cref{eq:shrinkage} becomes
\begin{equation}
   \label{eq:dead-leaves-shrinkage}
   \hat{v}_{cij} \approx \frac{1}{1 + \frac{\lambda}{\gamma} (\nrm{i}^\alpha + \nrm{j}^\beta)} \hat{v}_{\mathrm{LS},cij} \text{  for all  } cij,
\end{equation}
that is, sensitivity is monotonically decreasing with respect to both frequency
magnitude and the regularization coefficient \(\lambda\). This is consistent with
findings that CNNs trained on natural images are vulnerable to low frequency
perturbations
(\cite{guoLowFrequencyAdversarial2019,sharmaEffectivenessLowFrequency2019}).

\section{Experiments}
\label{sec:experiments}


When \(L> 2\) (that is, when there are more than 1 convolutional layers), the
\(p\) ``norm'' of \cref{eq:deep-ridge-alt} is non-convex (hence the quotes), and
in the limit as \(L \to \infty\) the norms appearing in \cref{eq:deep-ridge-alt}
converge to the Schatten 0-``norm,'' which is simply the number of non-0
singular values of a matrix.\footnote{In the special case where \(K=1\) the
regularization term in \cref{eq:deep-ridge-alt} is the
penalty of the \emph{subset selection} problem in the field of sparse linear
models.} Moreover, we see that the regularization coefficient of
\cref{eq:deep-ridge-alt} is effectively multiplied by \(L\).

\begin{figure}
   \centering
   \begin{subfigure}{0.45\linewidth}
      \centering
      \includegraphics[width=\linewidth]{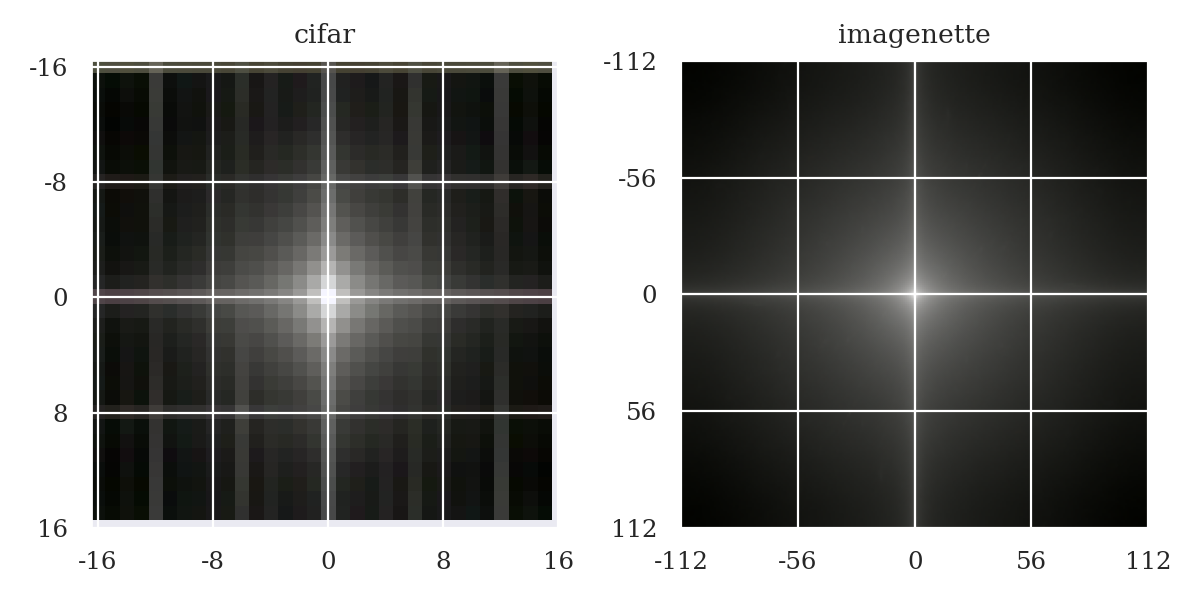}
      \caption{}\label{fig:datasets-powermaps}
   \end{subfigure}
   \begin{subfigure}{0.45\linewidth}
      \centering
      \includegraphics[width=\linewidth]{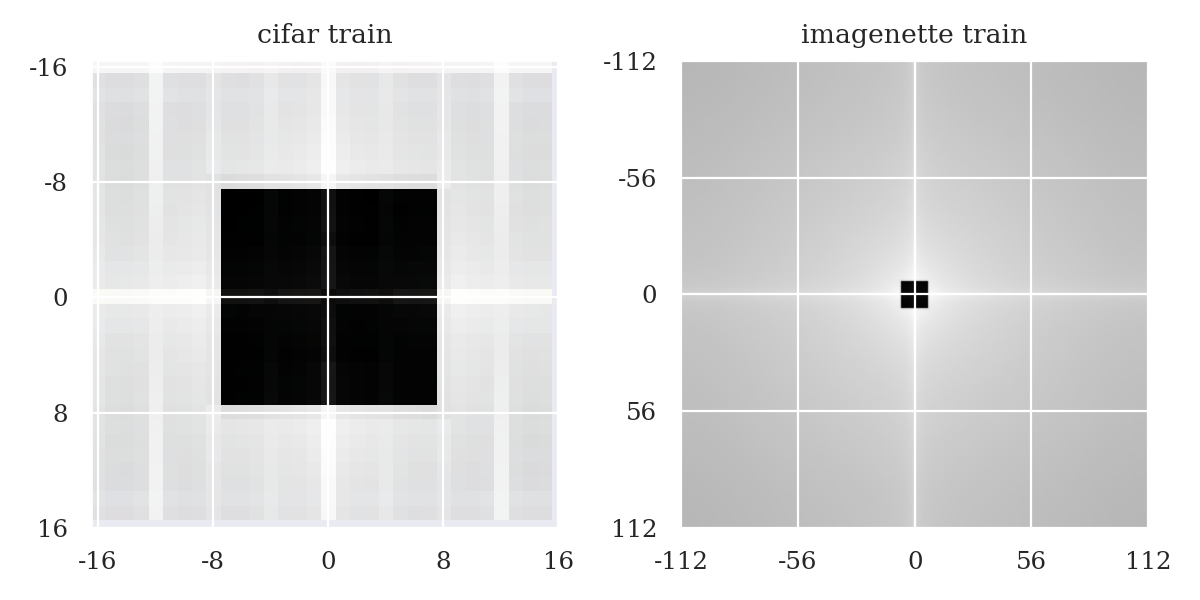}
      \caption{}\label{fig:datasets-powermaps-hp}
   \end{subfigure}
   \caption{\textbf{(a)} Standard deviations of image datasets along Fourier
   basis vectors: in the notation above, these are the
   \(\sqrt{\Sigma_{jiccij}}\) (viewed in log scale as RGB images). The origin
   corresponds to the lowest frequency basis vectors (i.e. constant images).
   \textbf{(b)} Same as (a), but with the addition of high pass filtering
   removing frequency indices \((i, j)\) with \(\nrm{i}, \nrm{j} \leq 8\) (we
   also experiment with a higher frequency cutoff on ImageNette in \cref{sec:exp-det}).}
\end{figure}

Even in the case where \(K=1\) so that \(\hat{v}\) is a vector, it is known that solving \cref{eq:deep-ridge-alt} for \(L>1\) is NP-hard
\cite{pmlr-v70-chen17d}, so we have no hope of finding closed form solutions as
in \cref{sec:1layer}. However, we can use the analysis in \cref{sec:reg-CNN-training} to derive three testable hypotheses:
\begin{enumerate}[I.,nosep]
   \item \label{hyp:content} CNN frequency
   sensitivity depends on the frequency content of training data (\cref{prp:data-dep-freq}).
   \item \label{hyp:depth} The fact that the regularization term of
   \cref{eq:deep-ridge-alt} becomes more sparsity-encouraging as \(L \) increases
   suggests that the data-dependent frequency sensitivity observed in
   \cref{sec:1layer,sec:lasso} becomes \emph{even more pronounced} as the number of
   convolutional layers increases.
   \item \label{hyp:decay}  Moreover, the functional forms of
   \cref{eq:shrinkage,eq:particularly-nice} suggest that the data-dependent frequency
   sensitivity will increase monotonically with the weight decay parameter
   \(\lambda\).
\end{enumerate}
We empirically validate these hypotheses with experiments using CNNs trained on
multiple datasets. The datasets used in these experiments are CIFAR10
\cite{Krizhevsky09learningmultiple},
ImageNette (the 10 easiest classes
from ImageNet \cite{deng2009imagenet}) \cite{imagenette}, and synthetic images generated using the
wavelet marginal model (WMM) of \cite{baradad2021learning}. The later dataset is of
interest since the generative model is \emph{explicitly} designed to capture the
frequency statistics of natural images, and allows for varying the exponents
\(\alpha\) and \(\beta\) in the power law \(\tau_{cij} \approx
\frac{\gamma}{\nrm{i}^\alpha + \nrm{j}^\beta}\) described above.
\Cref{fig:datasets-powermaps} display the variances CIFAR10, ImageNette and
their high pass filtered variants in frequency space; for those of the WMM datasets see \cref{fig:image-stats-dataset-details}.
In addition, we experiment with \emph{high pass filtered} versions of the
CIFAR10 and ImageNette datasets, which we refer to as hpf-CIFAR10 and hpf-ImageNette respectively; the frequency space statistics of these are
shown in \cref{fig:datasets-powermaps-hp}.\footnote{This was inspired by the
experiments of \cite{jo} and \cite{yin2019}.} For implementation details we refer
to \cref{sec:exp-det}. We can further summarize the data displayed in
\cref{fig:datasets-powermaps-hp,fig:datasets-powermaps} by averaging over circles with varying
radii \(r\), i.e. computing expectations \(E[\sqrt{\Sigma_{jiccij}} \, | \, \nrm{(i, j)} = r]\), to obtain the frequency \emph{magnitude} statistics curves shown in
\cref{fig:cifar-experiments} (see \cref{sec:cov-mats} for implementation details). From this point forward we
focus on such curves with respect to frequency magnitude. 

The CNNs used in these experiments are 
\begin{description}[nosep]
   \item[Linear CNNs] A family of CNNs that closely approximate
   \cref{eq:simpleCNN}, the only
   difference being that we include biases. This
   is accomplished by applying convolutions with \(32\times32\) kernels (i.e.,
   kernels of the same size as the input images) with circular padding.
   \item[ConvActually] A family of nonlinear CNNs obtained from the linear CNNs by
   including \(\relu\) non-linearities after each convolutional layer. 
   \item[Myrtle CNN] A small feed-forward CNN known to achieve high performance on CIFAR10
   obtained from \cite{pageHowTrainYour2018}.
   This CNN has small kernels, \(\relu\) non-linearities and \(\max\)-pooling,
   as well as exponentially varying channel dimension.
   \item[VVG] A family of CIFAR10 VGGs \cite{Simonyan15} obtained from \cite{vggcifar} and ImageNet VGGs from \cite{torchvision}.
   \item[ResNet] A family of CIFAR10 ResNets \cite{He2016DeepRL} obtained from
   \cite{mosaicml2022composer}.
   \item[AlexNet] A small AlexNet adapted to contrastive training obtained from \cite{baradad2021learning}.
\end{description}

For more detailed descriptions of datasets and model training we refer to
\cref{sec:dataset-details,sec:model-arch}.

We measure frequency sensitivity of a CNN \(f\) in terms of the magnitudes of
the directional derivatives \(\nabla_{x}f(x)^T \hat{e}_{cij}\), where
\(\hat{e}_{cij}\) is the \(cij\)-th Fourier basis vector. These magnitudes are
averaged over all the input images \(x\) in the relevant validation set, and as
in the case of image statistics we can average them over circles of varying
radii, i.e. compute expectations \(E[\nrm{\nabla_{x}f(x)^T \hat{e}_{cij}}\, | \,
\nrm{(i, j)} = r]\) (see \cref{sec:gradsensims} for implementation details). By ``data-dependent frequency sensitivity'' we mean
the extent to which the frequency sensitivity of a CNN \(f\) reflects the
statistics of the images it was trained on.
\Cref{fig:datasets-powermaps,fig:cifar-experiments,fig:vggi-experiments,fig:image-stats-dataset-details}
show that the variance of DFTed CIFAR10, ImageNette and WMM images is heavily
concentrated at low frequencies, in agreement with the power law form
\(\tau_{cij} \approx \frac{\gamma}{\nrm{i}^\alpha + \nrm{j}^\beta}\) described
in \cref{sec:reg-CNN-training}.\footnote{In the case of WMM this is by design.}
Hence in the absence of any modifications to the underlying images, we expect
that training on this data will emphasize sensitivity of \(f\) to perturbations
along the lowest frequency Fourier basis vectors, with the effect increasing
along with model depth and the weight decay parameter \(\lambda\). On the other
hand,  the variance of hpf-CIFAR10 and hpf-ImageNette is
concentrated in mid-range frequencies, and so here we expect training will
emphasize sensitivity of \(f\) to perturbations along mid range Fourier basis
vectors (again with more pronounced effect as depth and \(\lambda\) increase).

\subsection{Frequency sensitivity and depth}
\label{sec:freq-sens-depth}

\Cref{fig:cifar-experiments}a  shows sensitivity of ConvActually models of
varying depth to perturbations of varying DFT frequency magnitudes. These curves
illustrate that as depth increases, frequency sensitivity
\[E[\nrm{\nabla_{x}f(x)^T \hat{e}_{cij}}\, | \, \nrm{(i, j)} = r]\] more and
more closely matches the frequency magnitude statistics
\[E[\sqrt{\Sigma_{jiccij}} \, | \, \nrm{(i, j)} = r]\] of the training data set,
\emph{both} in the case of natural and high pass filtered images --- hence,
these empirical results corroborate hypotheses \ref{hyp:content} and
\ref{hyp:depth}. \Cref{fig:cifar-experiments}c shows results of a similar
experiment with VGG models of varying depth; models trained on natural images
have sensitivity generally decreasing with frequency magnitude, whereas those
trained on high pass filtered data have ``U'' shaped sensitivity curves with
minima near the filter cutoff, corroborating hypothesis \ref{hyp:content}. Here
it is not clear whether models trained on natural images follow that pattern
predicted by \ref{hyp:depth} (the deepest models are less sensitive to both low
and high frequencies), however when trained on high pass filtered images deeper
models do seem to most closely follow the frequency statistics of the dataset.
\Cref{fig:vggi-experiments} includes a similar experiment with VGG models of
varying depth trained on (hpf-)ImageNette. 

\subsection{Frequency sensitivity and weight decay}
\label{sec:freq-sens-decay}

\Cref{fig:cifar-experiments}b shows radial
frequency sensitivity curves for ConvActually models of depth 4 trained on
(hpf-)CIFAR10 with varying weight
decay coefficient \(\lambda\). We see that as  \(\lambda\)
increases, model frequency sensitivity more and more closely reflects the
statistics of the training data images,
corroborating hypotheses \ref{hyp:content} and \ref{hyp:decay}.
\Cref{fig:cifar-experiments}b shows results for a similar experiment with Myrtle
CNNs, with a similar conclusion.
\Cref{fig:vggi-experiments}
shows results for VGG models trained with varying weight
decay on (hpf-)ImageNette.

\subsection{Sensitivity to low frequencies, even after training on hpf data}
\label{sec:sens2low}

One aspect of our experimental results that is \emph{not} predicted by the
analysis of \cref{sec:DFT-CNN,sec:reg-CNN-training} is sensitivity of models
trained on hpf-CIFAR10 and hpf-ImageNette to the lowest Fourier frequencies.
This can be seen in \cref{fig:cifar-experiments}(a-d) as well as
\cref{fig:vggi-experiments,fig:vggi-higher-cutoff}. 

It is worth noting that in all of these figures, the \(y\)-axes for the
frequency sensitivities of models trained on natural and high pass filtered
images are distinct, and both are on a logarithmic scale. In general, we do see
that the models trained on high pass filtered images exhibit less sensitivity to
the lowest Fourier frequencies than their counterparts trained on natural
images, however inspection of confidence intervals shows this effect is not
always statistically significant.\footnote{The lack of statistical significance
is in part due to the large variance of sensitivity to low frequencies of models
trained on high pass filtered images.}

There are a number of possible causes of the persistence of sensitivity
to low Fourier frequencies even after training on high pass filtered images. The
first and most obvious is that the CNNs appearing in our experiments differ from
the idealized linear CNNs of \cref{sec:DFT-CNN} in a variety of ways
(including nonlinearities, small kernels and pooling). Another possibility is that our
analysis in \cref{sec:reg-CNN-training} does not account in any way for dynamics
of stochastic gradient descent. Although existing work on implicit bias
\cite{Gunasekar2018,lawrenceImplicitBiasLinear2022,lyuGradientDescentMaximizes2020b,yunUnifyingViewImplicit2021}
suggests that a statement analogous to \cref{thm:generalized-matrix-holder}
remains true, it is also known that gradient descent can proceed at dramatically
different rates in directions determined by training data statistics
\cite{pmlr-v178-xiao22a,xiaoLearningCurves,hacohen}. In the simple case of a
linear regression problem of the form 
\begin{equation}
   \min_\beta \frac{1}{2} \nrm{X \beta - Y}_2^2
\end{equation}
these directions are the principal components  of the matrix\footnote{If the
inputs are mean-centered, this matrix is \(N \) times the input data covariance matrix.} \(X^T X\) and the exponential rates of converge are the
corresponding eigenvectors, in the sense that if \(X^T X = V \diag(\lambda_i)
V^T\) where \(V\) is an orthogonal matrix with columns \(v_i\), and if
\(\beta_\infty = (X^T X)^{-1} X^T Y\), then gradient flow initialized at
\(\beta(0) = 0\) follows the path
\begin{equation}
   \label{eq:princ-comp-converg}
   \beta(t) = V(I - \diag(e^{-\lambda_i t})) V^T \beta_\infty.
\end{equation}
In particular, convergence is slowest in directions of low input data variance,
and in our experiment with high pass filtered images, there is \emph{zero}
variance along low frequency Fourier basis vectors. In an attempt to test this
possible explanation in terms of gradient descent dynamics, we compute radial
frequency sensitivity curves of all model architectures studied \emph{at random
initialization} in \cref{fig:untrained-nets}, finding that with a few notable
exceptions (the deeper ImageNette VGGs and ResNet models) they are relatively
flat (i.e. constant across frequency magnitudes). We also find that in many
cases for small frequency magnitudes \(r \) the unnormalized radially averaged
gradient norms \(E[\nrm{\nabla_{x}f(x)^T \hat{e}_{cij}}\, | \, \nrm{(i, j)} =
r]\) of randomly initialized models are \emph{smaller} than those of the models
in \cref{fig:cifar-experiments} trained on the high pass filtered datasets. This
suggests that gradient descent dynamics are \emph{not} the source of lingering
low-frequency sensitivity of the CNNs trained on hpf-\{CIFAR10, ImageNette\},
since it rules out the possibility that lingering sensitivity to low Fourier
frequencies in the models trained on high pass filtered data was present at
initialization and simply failed to shrink to zero due to uneven rates of
gradient descent as occur in \cref{eq:princ-comp-converg}. Thus we view the presence
of nonlinearities, small kernels and pooling as a more likely explanation,
however theoretically analyzing their impact on frequency sensitivity is a non-trivial
problem (see for example the appendices of \cite{pmlr-v178-jagadeesan22a}), and
we do not attempt to solve it in this work.

\subsection{Impact of the learning objective}
So far, our analysis and experiments have only shown that the regularization
term in \cref{eq:deep-ridg-obj} \emph{encourages} CNN gradients with respect to spatial
Fourier basis vectors to reflect the frequency statistics of the training data.
It is of course possible that the first term of \cref{eq:deep-ridg-obj} defining
the learning objective overwhelms the regularization term resulting in different
model frequency sensitivity. In \cref{fig:lwn-alpha} we show this occurs in
CNNs trained on WMM synthetic data with an alignment and uniformity contrastive
loss; see \cref{sec:more-exp-res} for a possible explanation in the framework of \cref{sec:DFT-CNN,sec:reg-CNN-training}.

\section{Limitations and open questions}
\label{sec:limits}

In order to obtain an optimization problem with some level of analytical
tractibility, we made many simplifying assumptions in
\cref{sec:DFT-CNN,sec:reg-CNN-training}, most notably omitting nonlinearities
from our idealized CNNs. While the experimental results of
\cref{sec:experiments} illustrate that multiple predictions derived from
\cref{sec:DFT-CNN,sec:reg-CNN-training} hold true for CNNs more closely
resembling those used in practice trained with supervised learning, \cref{fig:resnet-ctrex} shows that
hypotheses I-III can fail in the presence of residual connections --- see
\cref{sec:more-exp-res} for further discussion. As previously mentioned,
\cref{fig:lwn-alpha} shows that a contrastive alignment and uniformity learning
objective results in far different CNN representations.

Perhaps more significantly, it must be emphasized that model sensitivity as
measured by gradients represents a very small corner of a broader picture of
model robustness (or lack therof). For example, it does not encompass model
behavior on corruptions (see e.g. \cite{hendrycks2018benchmarking}) or shifted
distributions (see e.g. \cite{pmlr-v97-recht19a}).

%% file: appendix.tex
\section{Experimental details}
\label{sec:exp-det}

\subsection{Datasets}
\label{sec:dataset-details}

\begin{figure}
   \centering
   \begin{subfigure}{0.99\linewidth}
      \centering
      \includegraphics[width=\linewidth]{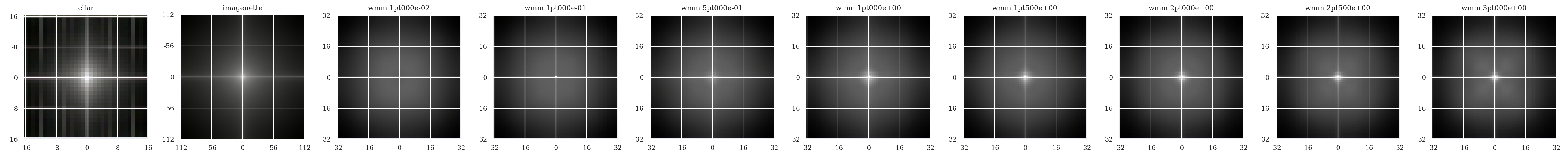}
   \end{subfigure}
   \begin{subfigure}{0.99\linewidth}
      \centering
      \includegraphics[width=\linewidth]{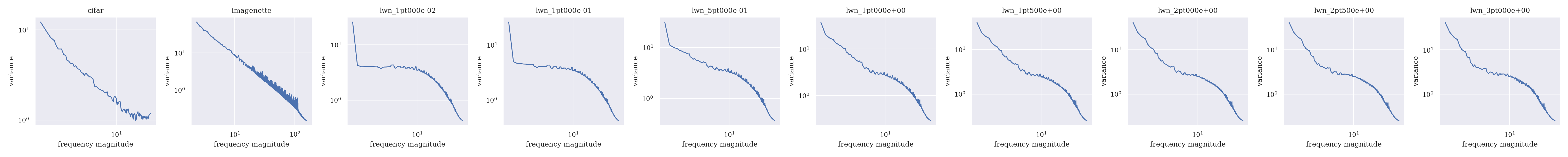}
   \end{subfigure}
   \caption{\textbf{Top row:} Standard deviations of image datasets (including
   those generated by the wavelet marginal model) along Fourier
   basis vectors: in the notation above, these are the
   \(\sqrt{\Sigma_{jiccij}}\) (viewed in log scale as RGB images). The origin
   corresponds to the lowest frequency basis vectors (i.e. constant images).
   \textbf{Bottom row:} Same data as the top row, further processed by
   taking radial averages \(E[\sqrt{\Sigma_{jiccij}} \, | \, \nrm{(i, j)} =
   r]\), and smoothing by averaging with three nearest neighbors on either side; shown in log-log-scale.}\label{fig:image-stats-dataset-details}
\end{figure}

\Cref{fig:image-stats-dataset-details} illustrates the frequency content of the
image datasets used in this paper. 

For CIFAR10 we use canonical train/test splits (imported using \cite{torchvision}).

As described at \cite{imagenette},
\begin{quote}
   Imagenette is a subset of 10 easily classified classes from Imagenet (tench, English springer, cassette player, chain saw, church, French horn, garbage truck, gas pump, golf ball, parachute).
\end{quote}
We use the ``full size'' version of the dataset with standard (224-by-224)
ImageNet preprocessing.

We implement high pass filtering by
passing each image \(x\) through the following preprocessing steps (before any
other preprocessing other than loading a JPEG image file as a tensor): \begin{enumerate*}[(i)]
   \item take the DFT \(\hat{x}\),
   \item multiply with a \emph{mask} \(m\) where \(m_{cij} = 0 \) if
   \(\nrm{i},\nrm{j} \leq \) some fixed threshold, and \(1\)
   otherwise,\footnote{Here we assume that the lowest frequency component is at
   the origin, so the low frequency components are zeroed out and the high
   frequency ones pass through.}
   \item applying the inverse DFT. 
\end{enumerate*}
For \emph{both} CIFAR10 and ImageNette, our threshold is 8 pixels. This means
that while 25\% of frequency indices are filtered out for CIFAR10, only about
0.13\% are filtered for ImageNette. The motivation for this approach was to
remove a similar amount of absolute frequency content in both cases; we also
experimented with a threshold of 112 in the case of ImageNette, resulting in
removal of 25\% of frequency indices in both cases, and results of these
experiments are shown in \cref{fig:vggi-higher-cutoff}.

We use the wavelet marginal model dataset generated using the implementation
in \cite{baradad2021learning}
--- details of this model are described in \cite[\S 3.3]{baradad2021learning}.
The generated images are of resolution 128-by-128; they are preprocessed with
downsampling to 96-by-96 and cropping to 64-by-64. For further preprocessing details we refer
to \cite[\S 4]{baradad2021learning}.

All pipelines described in this paper implement the standard preprocessing step
of subtracting the mean RGB value of the training dataset, and dividing by the
standard deviation of the training set RGB values.\footnote{For example, in the
case of ImageNet these are the canonical \texttt{[0.485, 0.456, 0.406]} and 
\texttt{[0.229, 0.224, 0.225]} respectively.} Note that this does \emph{not}
flatten the variances of the image distributions in frequency space displayed in
\cref{fig:cifar-experiments,fig:vggi-experiments}
--- in fact, it (provably) only impacts the variance of the 0th Fourier
component, corresponding to constant images. Ensuring that the variance in each
Fourier component is (approximately) 1 would require the far less standard
preprocessing step to each image \(x\) consisting of \begin{enumerate*}[(i)]
   \item apply the DFT to get \(\hat{x}\),
   \item subtract \(\hat{\mu}\), where \(\hat{\mu}\) is the mean of the DFTed
   images in the training dataset (\emph{not} just the mean of their RGB
   values),\footnote{Equivalently \(\hat{\mu}\) is the DFT of the mean training
   image.} 
   \item divide by the standard deviation \(\hat{\sigma}\) of the DFTed
   images in the training dataset, and finally
   \item apply the inverse DFT.
\end{enumerate*}
It should be emphasized that \(\hat{\mu} \) and \(\hat{\sigma}\) have the same
shape as \(x\), e.g. \((3,32,32)\) for the CIFAR10 dataset. The only work we are
aware of that implements such preprocessing on images is \cite[see \S
4]{hacohen}, though of course there may be others.

\subsection{Covariance matrices of DFTed image datasets}
\label{sec:cov-mats}

\begin{figure}
   \centering
      \begin{subfigure}{0.4\linewidth}
         \centering
         \includegraphics[width=\linewidth]{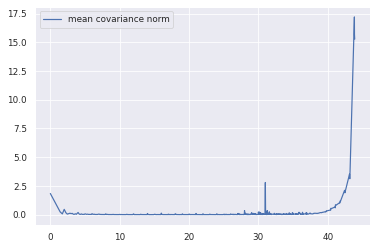}
         \caption{}\label{fig:cifar-covar-a}
      \end{subfigure}
      \begin{subfigure}{0.4\linewidth}
         \centering
         \includegraphics[width=\linewidth]{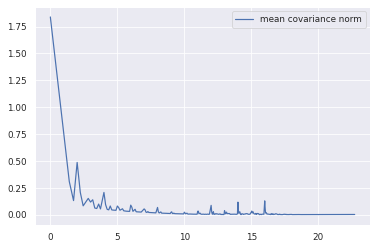}
         \caption{}\label{fig:cifar-covar-b}
      \end{subfigure}
      \caption{Testing the assumption of \cref{eq:diagonality-assumption} on the
      CIFAR10 dataset. \textbf{(a)} Average covariance entry norm
      \(E[\nrm{\Sigma_{jicc'i'j'}} \, | \, \nrm{(c, i, j) - (c, i', j')} = r]\)
      plotted with respect to \(r = \nrm{(i, j) - (i', j')}\). \textbf{(b)} Same as
      (a), except the \(i, j\) part of the distance \(r\) is computed
      \emph{modularly}, i.e. on the discrete torus \(\ZZ /H \times \ZZ/W\) instead
      of the discrete square \(\{0, \dots, H-1\} \times \{0, \dots, W-1\}\). For
      more details see \cref{sec:exp-det}.
      }\label{fig:cifar-covar}
\end{figure}

In this section we provide further details on our computations of (co)variances and standard deviations of DFTed image datasets.

To compute the covariances \(\Sigma_{jicc'i'j'}\) for \cref{fig:cifar-covar}, we begin with a dataset of natural images, say \(X \). We subtract its mean RGB pixel value (a vector in \(\RR^3 \)) and divide by the standard deviation of RGB pixel values (also a vector in \(\RR^3 \)) as is standard. We next apply the DFT to every image in \(X \), to obtain a DFTed dataset \(\hat{X}\). We then compute the mean \(\hat{\mu}\) of the \emph{images} in \(\hat{X}\), not the RGB pixel values --- this is a \(3 \times H \times W \) tensor, where \(H, W \) are the heights/widths of the images in \(X \) (e.g., 32 for CIFAR10). This mean is then subtracted from \(\hat{X}\) to obtain a centered dataset. Next, we sample batches of size \(B \), say \(\hat{x}_1, \dots, \hat{x}_B \), from \(\hat{X}\) (these are tensors of shape \(B \times C \times H \times W\)). For each batch, we subtract the mean, contract over the batch index and divide by \(B \) to obtain an estimate for \(\Sigma \), say \(\tilde{\Sigma}\) like so:
\begin{equation}
   \tilde{\Sigma} = \frac{1}{B} \sum_{b=1}^B (\hat{x}_b -\hat{\mu})^T \otimes (\hat{x}_b -\hat{\mu}); \text{ explicitly  }  \tilde{\Sigma}_{jicc'i'j'} = \frac{1}{B} \sum_{b=1}^B (\hat{x}_{b, jic} -\hat{\mu}_{jic})(\hat{x}_{b, c'i'j'} -\hat{\mu}_{c'i'j'})
\end{equation}
Finally, we average over an entire dataset's worth of batches to get our final estimate of \(\Sigma_{jicc'i'j'}\).

To estimate the expectations \(E[\nrm{\Sigma_{jicc'i'j'}} \, | \, \nrm{(c, i, j) - (c, i', j')} = r]\), we average absolute values \(\nrm{\Sigma_{jicc'i'j'}}\) over all indices
\(jicc'i'j'\) satisfying \(\nrm{(c, i, j) - (c', i', j')} =r\). The number of
such entries varies significantly with \(r\), which is why we have not explicitly written down
the average. We compute ``modular'' distances, i.e. distances on the discrete torus, using the formulae
\begin{equation}
    \begin{split}
        d(i, j') &= \min\{\nrm{(h-h') \mathrm{mod} H}, \nrm{H-((h-h') \mathrm{mod} H)}\}\\
        d(w, w') &= \min\{\nrm{(w - w') \mathrm{mod} W}, \nrm{W-((w-w') \mathrm{mod} W)}\}
    \end{split}
\end{equation}
and finally \(  d((c,h,w), (c',h',w')) = \sqrt{(c-c')^2 + d(h, h')^2 + d(w, w')^2} \).

In \cref{fig:datasets-powermaps}, we begin with a dataset of natural images, say \(X \). We subtract its mean RGB pixel value (a vector in \(\RR^3 \)) and divide by the standard deviation of RGB pixel values (also a vector in \(\RR^3 \)) as is standard. We next apply the DFT to every image in \(X \), to obtain a DFTed dataset \(\hat{X}\). We then compute the standard deviation of the \emph{images} in \(\hat{X}\), not the RGB pixel values --- this is a \(3 \times H \times W \) tensor, where \(H, W \) are the heights/widths of the images in \(X \) (e.g., 32 for CIFAR10).

\subsection{Gradient sensitivity images}
\label{sec:gradsensims}
To compute the sensitvity curves in
\cref{fig:cifar-experiments,fig:vggi-experiments,fig:resnet-ctrex,fig:lwn-alpha},
we subsample 5,000 images from the underlying validation dataset in the case of
CIFAR10 and WMM use the entire validation dataset in the case of ImageNette. For
each such image \(x \), we compute the DFT \(\hat{x} \), and then backpropagate
gradients through the composition \( \hat{x} \to \hat{\hat{x}} = x \to f(x) \).
The result is a \(C \times H \times W \times K \) Jacobian matrix expressing the
derivative of \(f\) with respect to Fourier basis vectors. We take \(\ell_2 \)
norms over the class index (corresponding to \(K\)) to obtain a \( C \times H
\times W \) gradient norm image, with \( c, h, w \) component
\(\nrm{\nabla_{x}f(x)^T \hat{e}_{cij}}\). Finally, we average these gradient
norms over the (subsampled in the case of CIFAR10 and WMM) dataset.

Next, we average radially much as we did in \cref{fig:cifar-covar}. Given an
expected gradient norm image \(E[\nrm{\nabla_{x}f(x)^T   \hat{e}_{cij}}]\)
obtained as above, we further average over all indices \( cij \) such that
\(\sqrt{i^2 + j^2} = r\). Again, the number of such indices is highly variable.

To provide error estimates, for each set of trained model weights and each image
dataset, we compute radial frequency sensitivity curves as in the paragraph
above, apply post-processing consisting of:
\begin{itemize}
   \item Dividing the curve by its integral (to obtain a probability
   distribution)\footnote{this results in a comparison between models that is
   \emph{scale invariant}, that is, we want to compare the shapes of frequency
   sensitivity curves, not their overall magnitudes}
   \item Smoothing by averaging with 3 nearest neighbors on each side. This step
   is motivated by the aforementioned highly variable number of DFT frequency indices
   corresponding to a given radius, which means that some radius values
   correspond to an average over far fewer samples which as a result has high variance. A possible alternative would have been to
   \emph{bin} radii, thus averaging over DFT components with frequency
   magnitudes in a small interval. However, the automatic binning strategies we
   tried yielded bins that were too large, giving an undesirably low resolution view of
   frequency sensitivity curves.
   \item Taking the logarithm of the resulting probability distribution. This
   step is motivated by the empirical observation that the variance of Fourier
   transformed images follows a power law with respect to frequency magnitude.
\end{itemize}
We then repeat the entire pipeline above for 5 sets of model weights trained
from independent random initializations, to obtain 5 curves, and display the
standard deviation of their \(y\)-values. For more details on our training
procedures, see \cref{sec:model-arch}.

\subsection{More experimental results}
\label{sec:more-exp-res}

\begin{figure}[!htb]
   \centering
   \begin{subfigure}{0.9\linewidth}
      \centering
      \includegraphics[width=\linewidth]{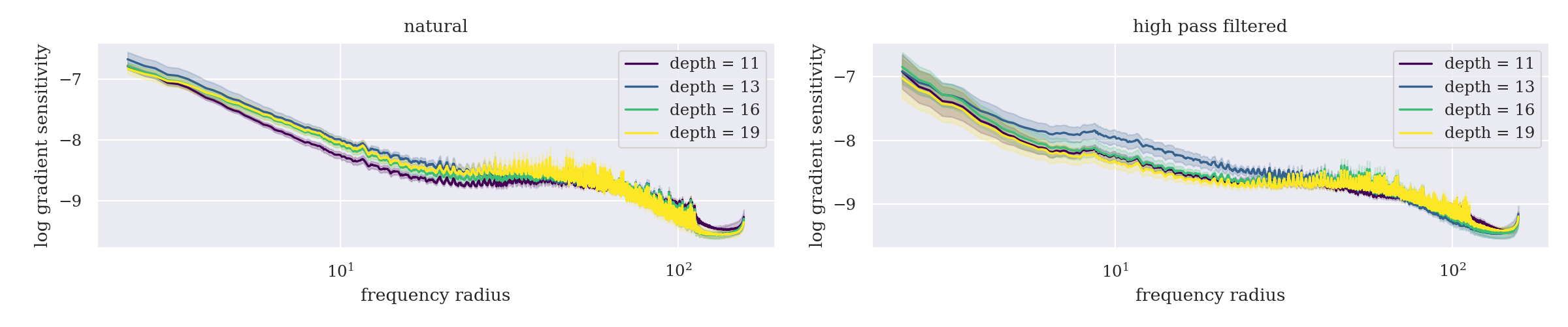}
      \caption{VGG models of varying depth.}
   \end{subfigure}
   \begin{subfigure}{0.9\linewidth}
      \centering
      \includegraphics[width=\linewidth]{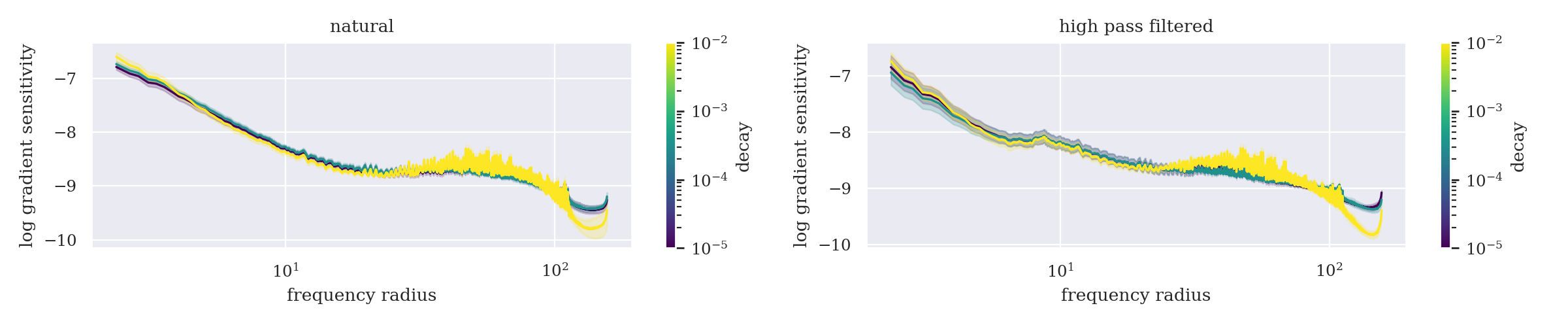}
      \caption{VGG 11 models trained with varying levels of decay.}
   \end{subfigure}
   \begin{subfigure}{0.4\linewidth}
      \centering
      \includegraphics[width=\linewidth]{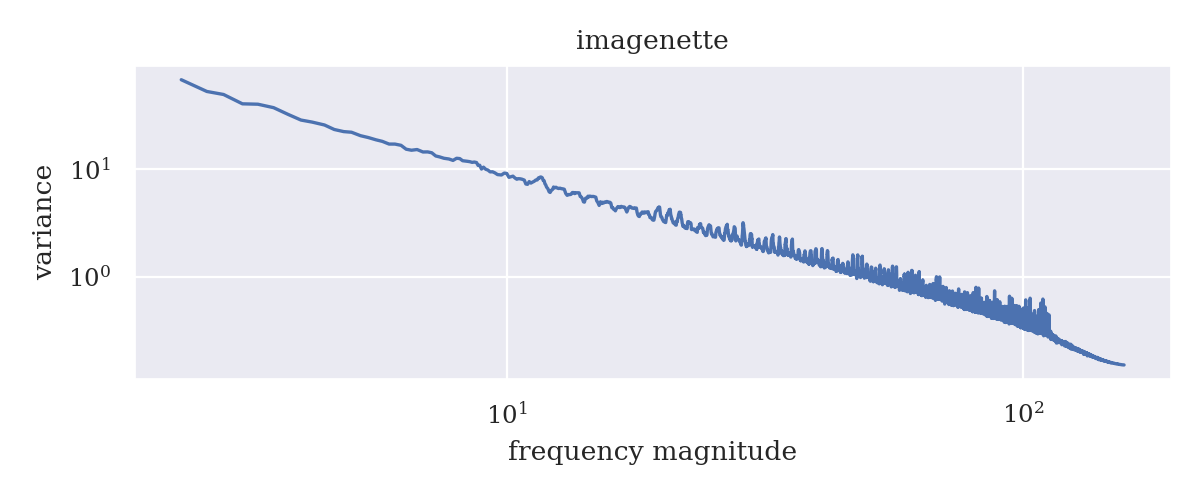}
      
   \end{subfigure}
   \begin{subfigure}{0.4\linewidth}
      \centering
      \includegraphics[width=\linewidth]{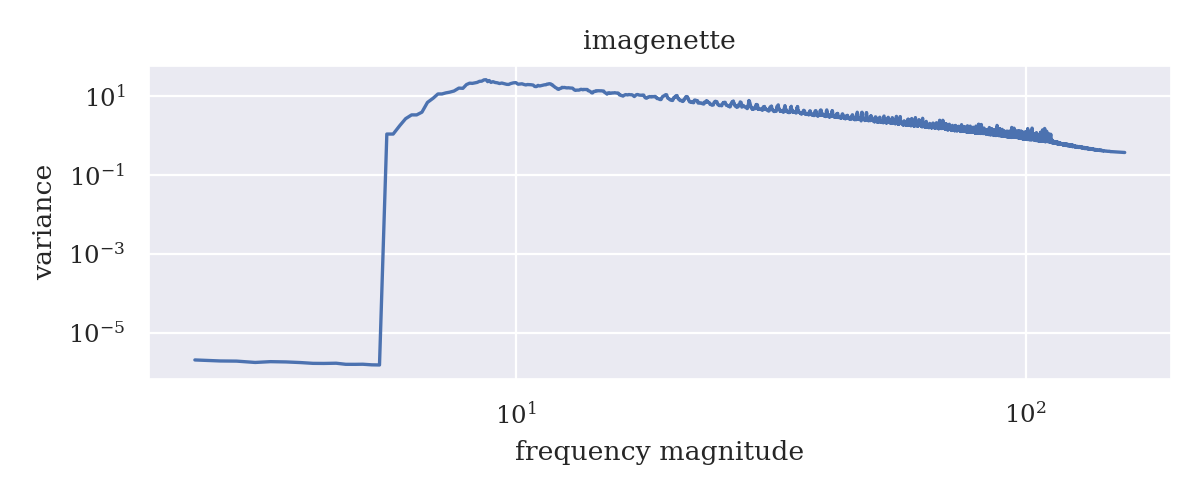}
      
   \end{subfigure}
   \caption{Radial averages \(E[\nrm{\nabla_{x}f(x)^T \hat{e}_{cij}}\, | \,
   \nrm{(i, j)} = r]\) of frequency sensitivities of VGG models trained on ImageNette and
   its high pass filtered variant, post processed as discussed in \cref{sec:gradsensims}. \textbf{Bottom row}: frequency statistics
   of ImageNette and its high pass filtered variant for comparison.}\label{fig:vggi-experiments}
\end{figure}

\Cref{fig:vggi-experiments} shows radial frequency sensitivity curves from
experiments training VGGs with variable depth and
weight decay on ImageNette, with and without high pass filtering. Here we do see
significant differences between models trained on natural vs. high pass filtered
images, including in the later case (small) local peaks in near the filter
cutoff, however the effect of filtering is not nearly as noticeable as in the
CIFAR10 experiments of \cref{fig:cifar-experiments}. 

Moreover, effects of depth and decay are not as dramatic in these experiments,
although we do see the curves corresponding to high depth/decay for VGGs trained
on natural images dropping off most severely at high frequencies, and in the
case of high pass filtered images deep VGGs are the least sensitive in the
low frequency range (where the training images have no variance).

Interestingly, we see a range of frequency radii, roughly \(r \in [10,100]\),
where the ImageNette frequency statistics exhibit significant noise around the
overall power law pattern, and it does appear that all VGG models concentrate
frequency sensitivity in this range, and more so with greater depth/decay. It
would be interesting to understand what sorts of natural image features
contribute to  the observed noise in the \(r \in [10,100]\) range, and if they
are for some reason useful to the ImageNette classification task.

\begin{figure}[!htb]
   \centering
   \begin{subfigure}{0.9\linewidth}
      \includegraphics[width=\linewidth]{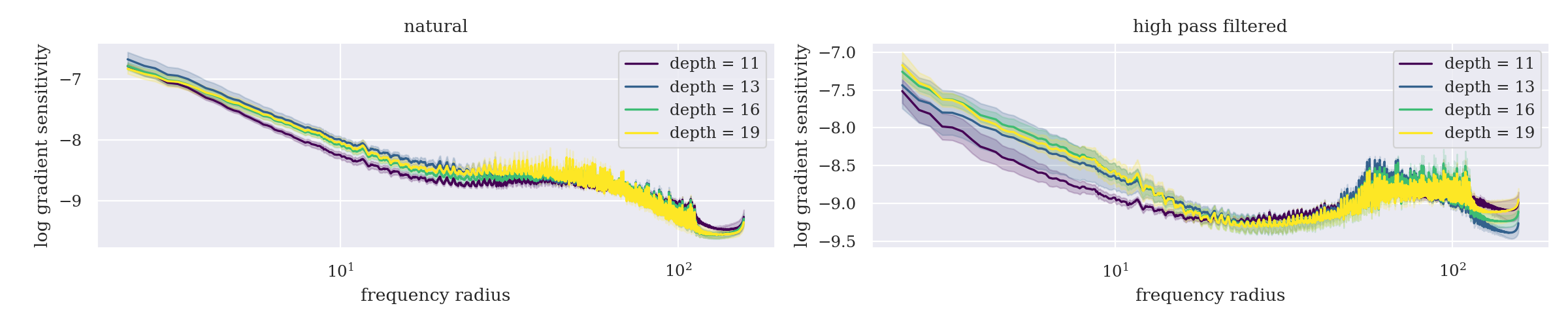}
   \end{subfigure}
   \begin{subfigure}{0.9\linewidth}
      \includegraphics[width=\linewidth]{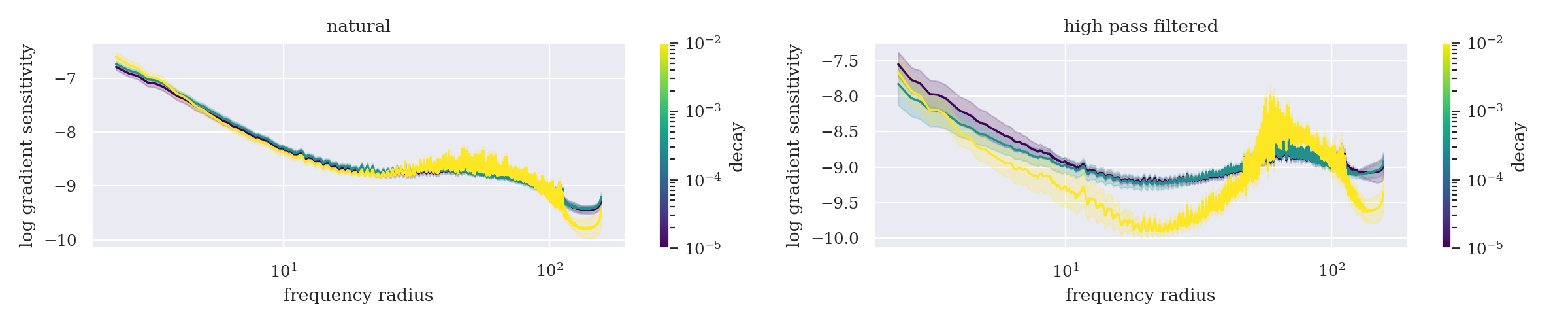}
   \end{subfigure}
   \begin{subfigure}{0.4\linewidth}
      \includegraphics[width=\linewidth]{paper_plots_normalized/pmaps_radial_loglog_imagenette.png}
   \end{subfigure}
   \begin{subfigure}{0.4\linewidth}
      \includegraphics[width=\linewidth]{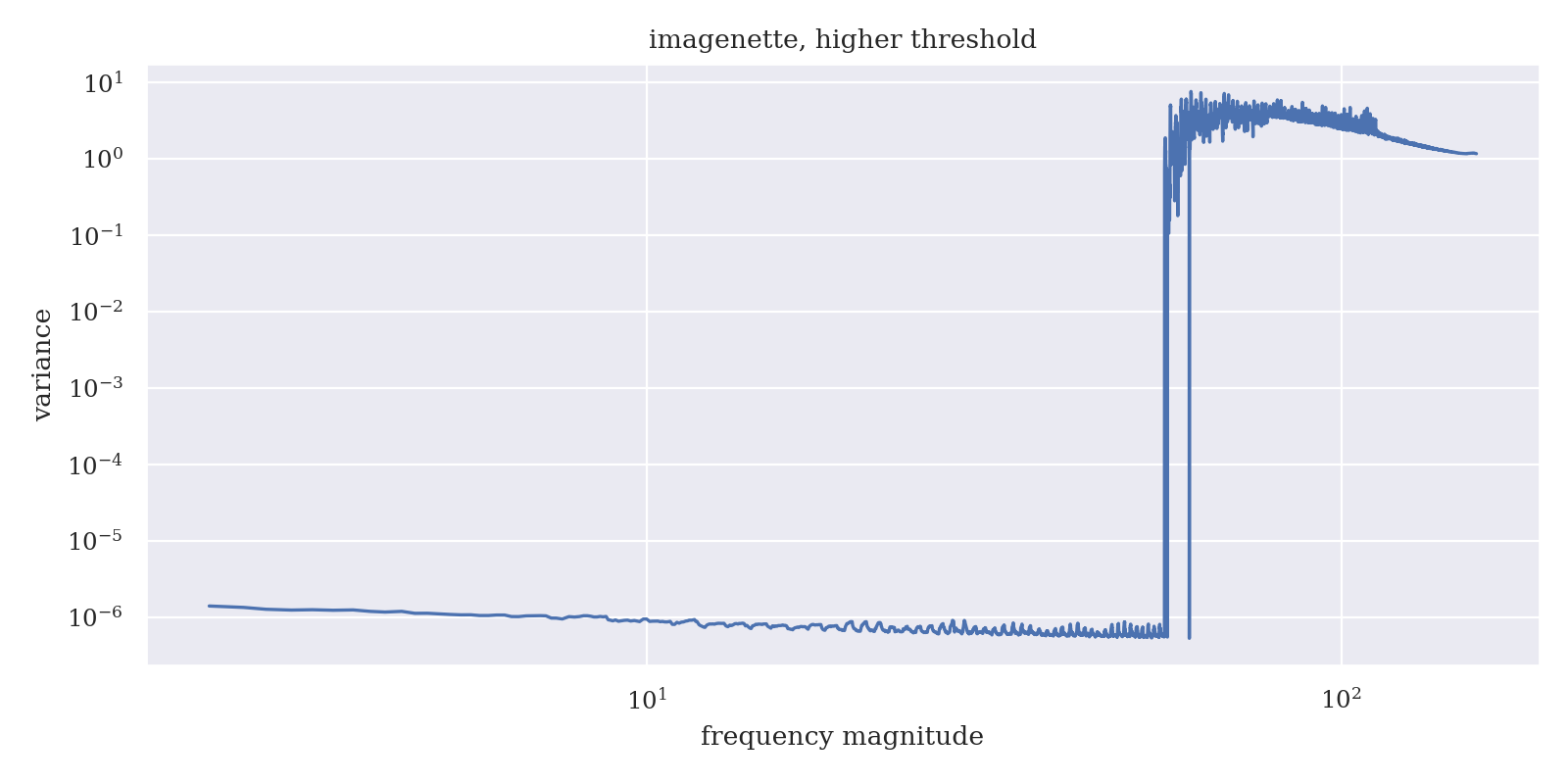}
   \end{subfigure}
   \caption{VGGs trained on hpf-ImageNette with a larger cutoff (removing 25\% of frequency indices).}\label{fig:vggi-higher-cutoff}
\end{figure}

\Cref{fig:vggi-higher-cutoff} shows a similar experiment with VGGs trained on
(hpf-)ImageNette, but with a high-pass threshold of 112, so that 25\% of
frequency indices are discarded. Here the differences between CNNs trained on natural and high
pass filtered  images are (not surprisingly) more dramatic, and the effect of
decay on VGG 11s trained on hpf-ImageNette is especially pronounced. It is (at best)
unclear whether the results for VGGs of variable depth trained on hpf-ImageNette
support hypothesis \ref{hyp:depth} (depth). 

\begin{figure}[!htb]
   \centering
   \begin{subfigure}{0.9\linewidth}
      \centering
      \includegraphics[width=\linewidth]{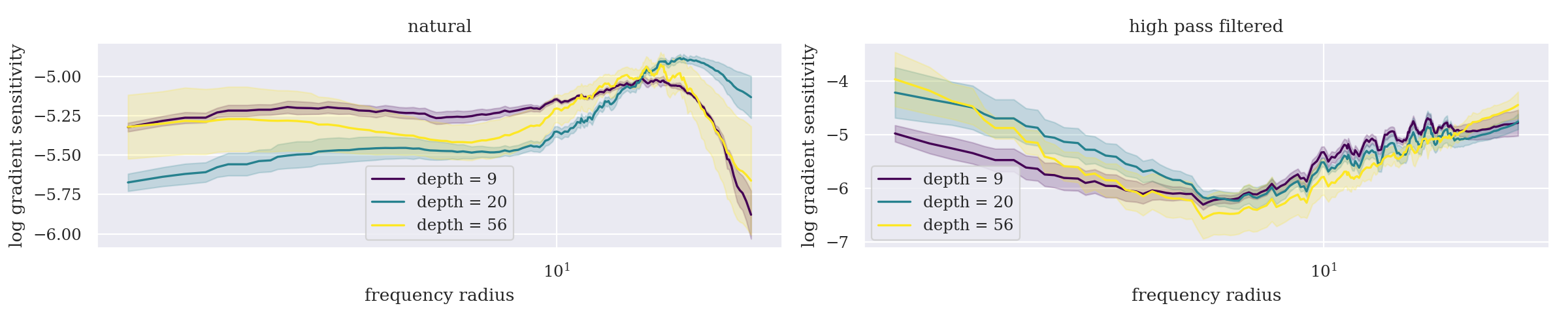}
      \caption{ResNet models of varying depth.}
   \end{subfigure}
   \begin{subfigure}{0.9\linewidth}
      \centering
      \includegraphics[width=\linewidth]{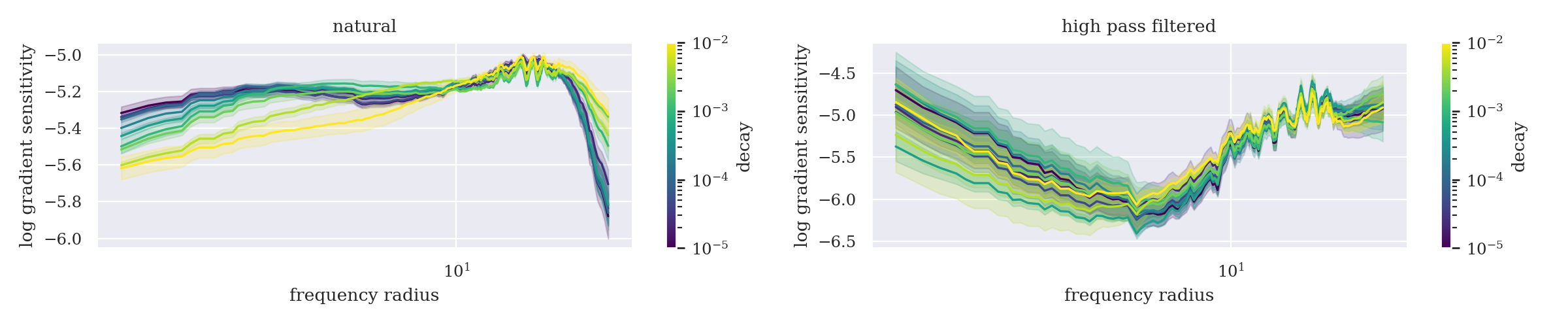}
      \caption{ResNet 9 models trained with varying levels of decay.}
   \end{subfigure}
   \begin{subfigure}{0.4\linewidth}
      \centering
      \includegraphics[width=\linewidth]{paper_plots_normalized/pmaps_radial_loglog_cifar.png}
      
   \end{subfigure}
   \begin{subfigure}{0.4\linewidth}
      \centering
      \includegraphics[width=\linewidth]{paper_plots_normalized/pmaps_hp_radial_loglog_cifar.png}
      
   \end{subfigure}
   \caption{Radial averages \(E[\nrm{\nabla_{x}f(x)^T \hat{e}_{cij}}\, | \,
   \nrm{(i, j)} = r]\) of frequency sensitivities of ResNet models of varying
   depth (top row) and decay (middle row) trained on CIFAR10 and its high pass
   filtered variant. Post processing is as discussed in
   \cref{sec:gradsensims}.}\label{fig:resnet-ctrex}
\end{figure}

\Cref{fig:resnet-ctrex} shows radial frequency sensitivity curves from experiments training ResNets with variable depth and
weight decay on CIFAR10, with and without high pass filtering. The
frequency sensitivity curves are clearly different when trained on natural
images (where they drop off at the highest frequencies) versus
high pass filtered images (where they have a ``U''-shape similar to those of the
VGG models trained on high pass filtered CIFAR10 in
\cref{fig:cifar-experiments}). These observations seems somewhat consistent with
hypothesis \ref{hyp:content} (data-dependent frequency sensitivity). However, the trends with depth and decay do \emph{not}
conform with hypotheses \ref{hyp:depth}, \ref{hyp:decay} (depth, decay): it is difficult to see trends in the
depth experiment, and in the decay experiment the frequency sensitivity curves
seem to become more \emph{increasing} as decay increases, rather than adhering
to the frequency content of the dataset. It is not immediately clear to the
authors what causes this behavior --- further investigation would be an
interesting direction for future work.

\begin{figure}
   \centering
   \includegraphics[width=0.8\linewidth]{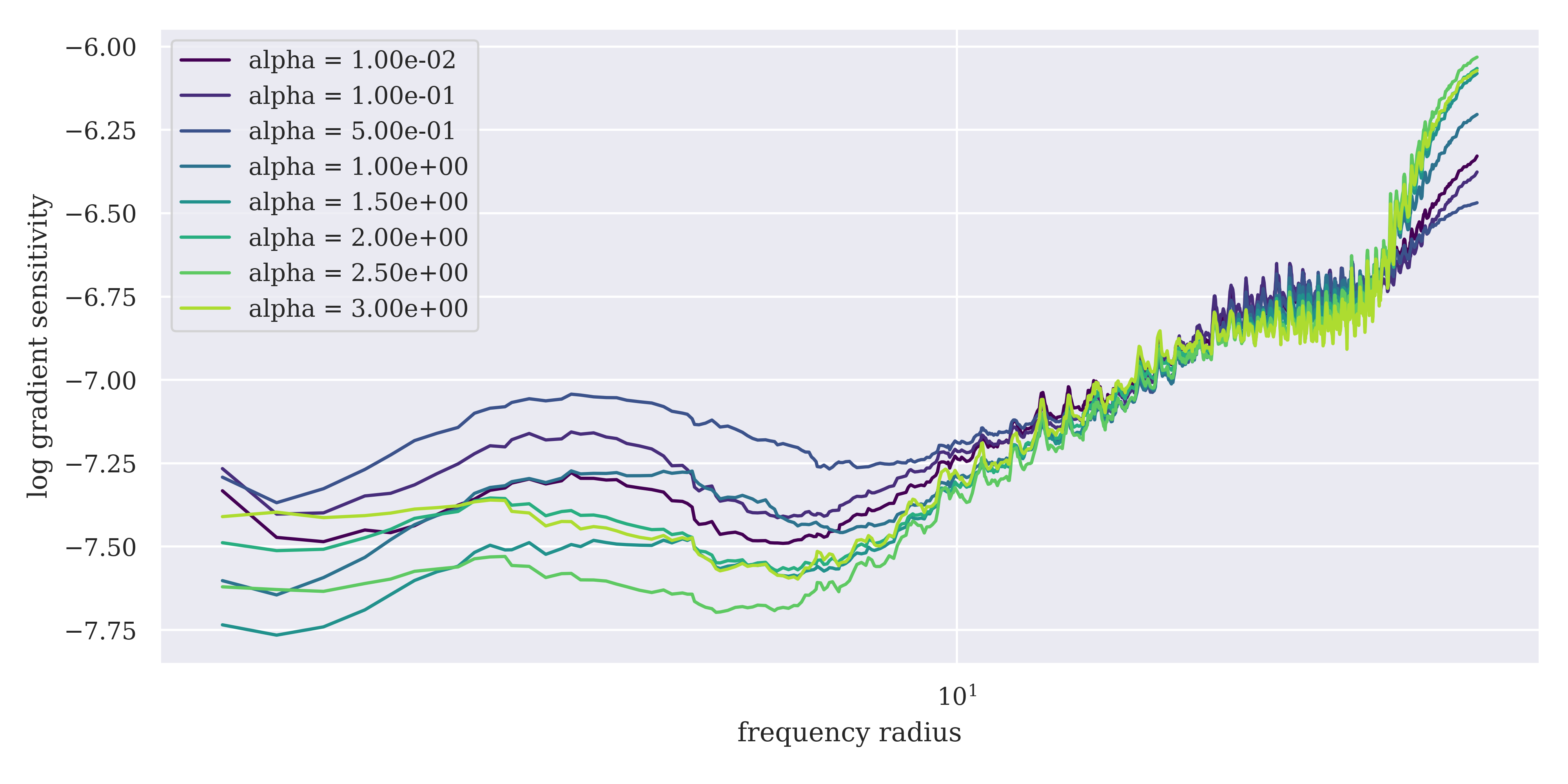}
   \caption{Radial averages \(E[\nrm{\nabla_{x}f(x)^T \hat{e}_{cij}}\, | \, \nrm{(i, j)} = r]\) of frequency sensitivities of
   unsupervised AlexNet models trained with alignment and uniformity loss on data
   generated by WMM models with varying \(\alpha\) parameter. Post-processing is
   as discussed in \cref{sec:gradsensims}.}\label{fig:lwn-alpha}
\end{figure}

Lastly, \cref{fig:lwn-alpha} shows frequency sensitivity curves for AlexNets
trained with alignment and uniformity loss \cite{pmlr-v119-wang20k} on
synthetic data generated by the wavelet marginal model (WMM) of
\cite{baradad2021learning} with varying \(\alpha\) parameter; to be specific,
for simplicity we set \(\alpha = \beta \) in the power law \(\tau_{cij} \approx
\frac{\gamma}{\nrm{i}^\alpha + \nrm{j}^\beta}\). These results initially
surprised us: for all  \(\alpha\) the variance of the synthetic data is
concentrated in low frequencies (\cref{fig:image-stats-dataset-details}), with
the level of concentration decreasing as \(\alpha\)  increases (i.e. as
\(\alpha\) increases, the frequency curve of the synthetic data spreads out).
\emph{However}, for all AlexNets trained on WMM data the frequency sensitivity
curves \emph{increase} with frequency magnitude, with slope roughly
\emph{increasing} with respect to \(\alpha\)! 

This result, which shows that CNNs trained with contrastive learning objectives
can respond quite differently to the statistics of their training data, admits a
simple explanation in terms of \cref{sec:DFT-CNN,sec:reg-CNN-training}.

Indeed, the uniformity part of the objective encourages the set of vectors 
\begin{equation}
   \{\hat{f}(\hat{x^n}) = \hat{v}^T \hat{x}^n \, | \, n=1, \dots, N \}
\end{equation}
to be uniformly distributed on the unit sphere \(S^{K-1} \subseteq \RR^K\).
Suppose now for the sake of simplicity that that the vectors \(\hat{x}^n = (
\hat{x}^n_{cij} )\) are normally distributed and their covariance matrix \(\Sigma\) is
diagonal, with diagonal entries of the form 
\begin{equation}
   \tau_{cij} \approx \frac{\gamma}{\nrm{i}^\alpha + \nrm{j}^\alpha}.
\end{equation}
Then the feature vectors \(\hat{f}(\hat{x^n}) = \hat{v}^T \hat{x}^n\) are also
normally distributed, with covariance matrix 
\begin{equation}
   \label{eq:transformed-covar}
   \hat{v}^T \Sigma \hat{v} = \sum_{ij}  \frac{\gamma}{\nrm{i}^\alpha + \nrm{j}^\alpha} \hat{v}_{ij} \hat{v}_{ij}^T.
\end{equation}
Here as above \(\hat{v}_{ij}\) is a \(K \times C\) matrix, so each
\(\hat{v}_{ij} \hat{v}_{ij}^T\) is indeed a \(K \times K\) positive
semi-definite matrix, and in cases of interest where \(C \ll K\) it will have
very low rank. Considering \cref{eq:transformed-covar}, we see for
example that if \(C \ll K\) and the \(\hat{v}_{ij}\) are of roughly constant magnitude, then the
covariance of the features \(\hat{f}(\hat{x^n})\) will be dominated by the terms 
\begin{equation}
   \frac{\gamma}{\nrm{i}^\alpha + \nrm{j}^\alpha} \hat{v}_{ij} \hat{v}_{ij}^T \text{  for small  } ij
\end{equation}
making it impossible for the \(\hat{f}(\hat{x^n})\) to be uniformly distributed
on the unit sphere. It seems that perhaps the \emph{only} way for such
uniform distribution to occur is for the magnitudes of the \(\hat{v}_{ij}\) to
increase with the frequency magnitude \(\nrm{(i,j)}\), in such a way as to
offset the denominator \(\nrm{i}^\alpha + \nrm{j}^\alpha\). We do not have a
proof of this fact (it seems such a proof would have to involve analysis of the
functional form of the contrastive loss in \cite{pmlr-v119-wang20k}, which we
have not carried out), but it does offer one potential explanation of
\cref{fig:lwn-alpha}. At the risk of being overzealous, the above discussion
would suggest that
\begin{equation}
   \nrm{\hat{v}_{ij}} \propto \nrm{i}^\alpha + \nrm{j}^\alpha,
\end{equation}
i.e. the norm of the gradient of \(f\) with respect
to the \(ij\)-th DFT basis vector, follows a power law in \(\alpha\), and indeed
in \cref{fig:lwn-alpha}, where these gradient norms are plotted with respect to
frequency magnitude on a log-log scale, we see slope roughly increasing with
\(\alpha\). 

\subsection{Frequency sensitivity of untrained networks}
\label{sec:fs-untrained}

\begin{figure}[!htb]
   \centering
   \begin{subfigure}{0.45\linewidth}
      \centering
      \includegraphics[width=\linewidth]{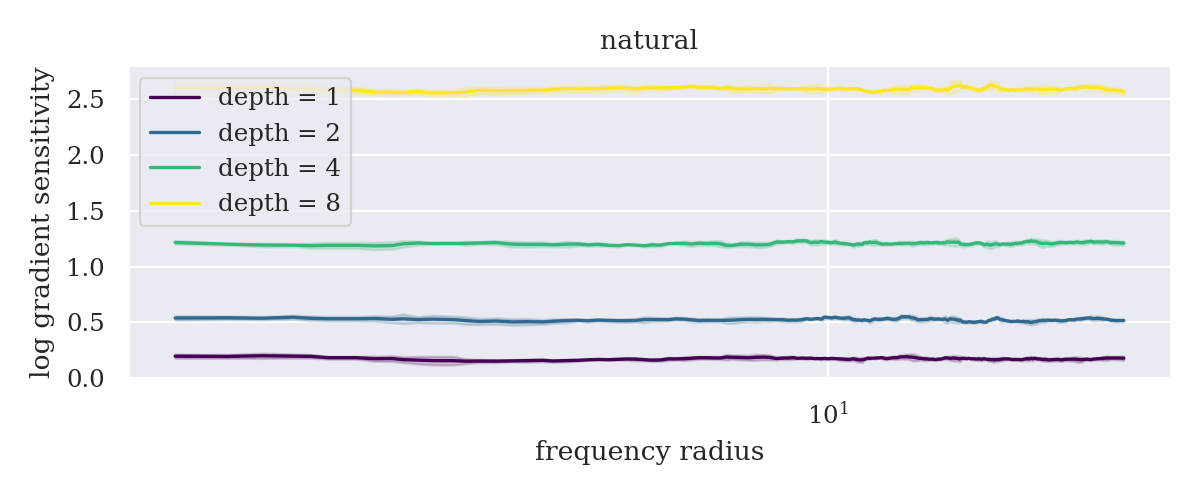}
      \caption{\footnotesize Linear CNNs of varying depth.}
   \end{subfigure}
   \begin{subfigure}{0.45\linewidth}
      \centering
      \includegraphics[width=\linewidth]{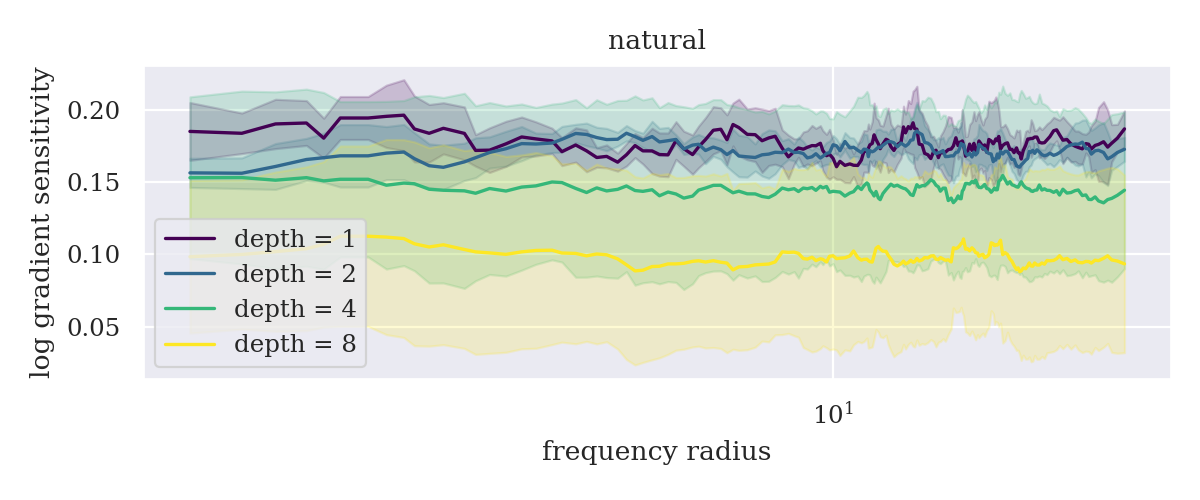}
      \caption{\footnotesize ConvActually models of varying depth.}
   \end{subfigure}
   \begin{subfigure}{0.45\linewidth}
      \centering
      \includegraphics[width=\linewidth]{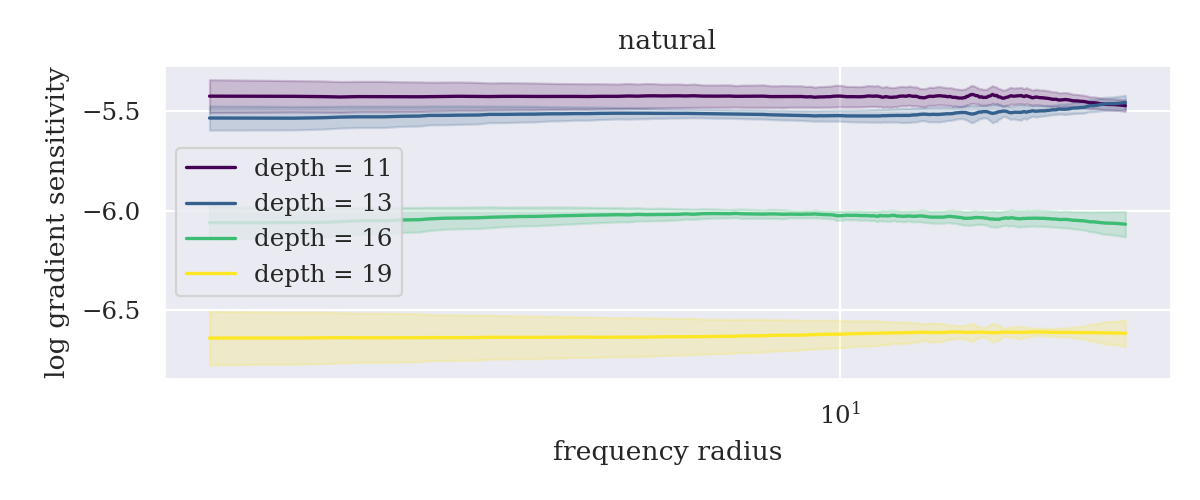}
      \caption{\footnotesize VGG models of varying depth.}
   \end{subfigure}
   \begin{subfigure}{0.45\linewidth}
      \centering
      \includegraphics[width=\linewidth]{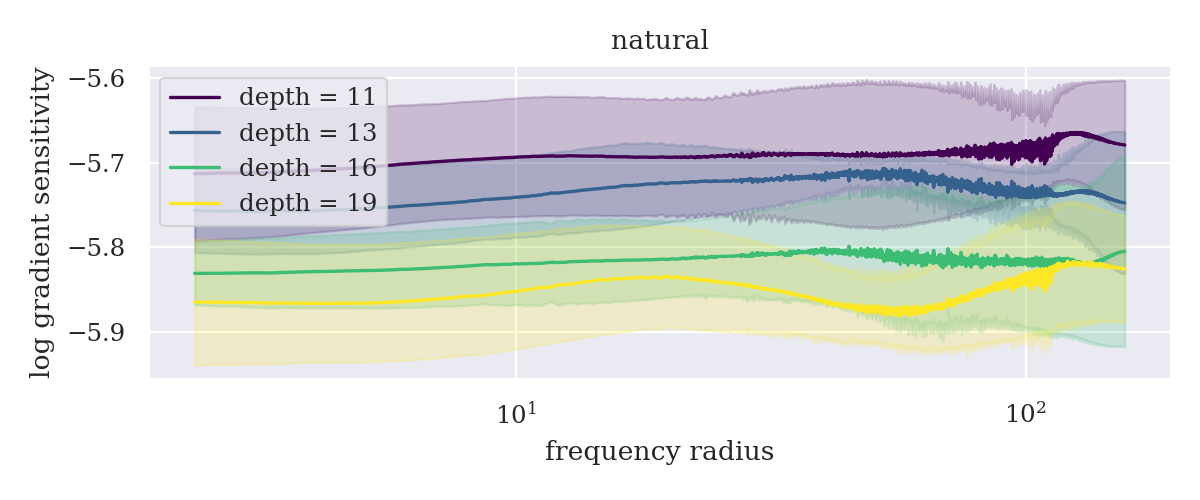}
      \caption{\footnotesize ImageNette VGG models of varying depth.}
   \end{subfigure}
   \begin{subfigure}{0.45\linewidth}
      \centering
      \includegraphics[width=\linewidth]{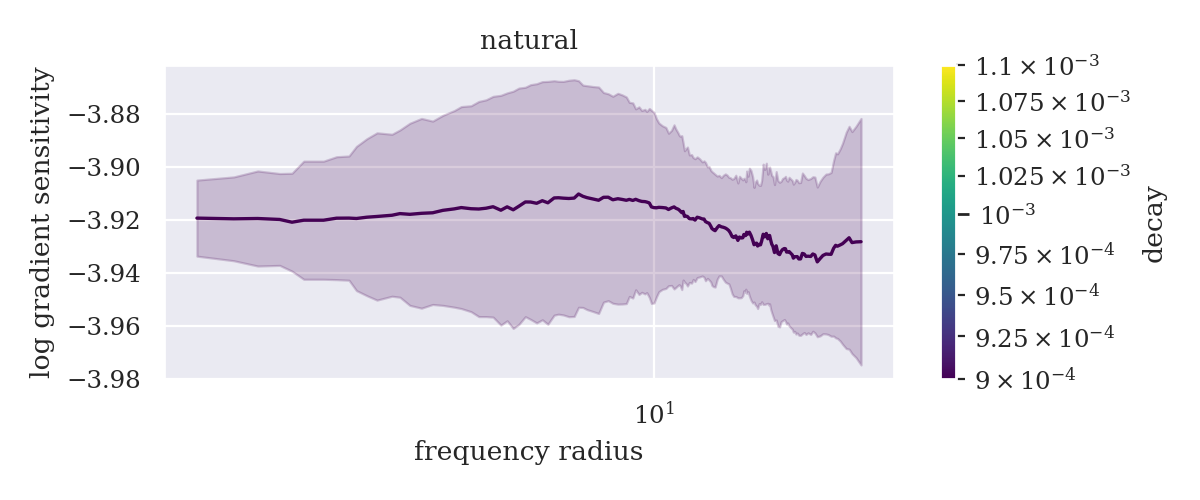}
      \caption{\footnotesize Myrtle CNN model.}
   \end{subfigure}
   \begin{subfigure}{0.45\linewidth}
      \centering
      \includegraphics[width=\linewidth]{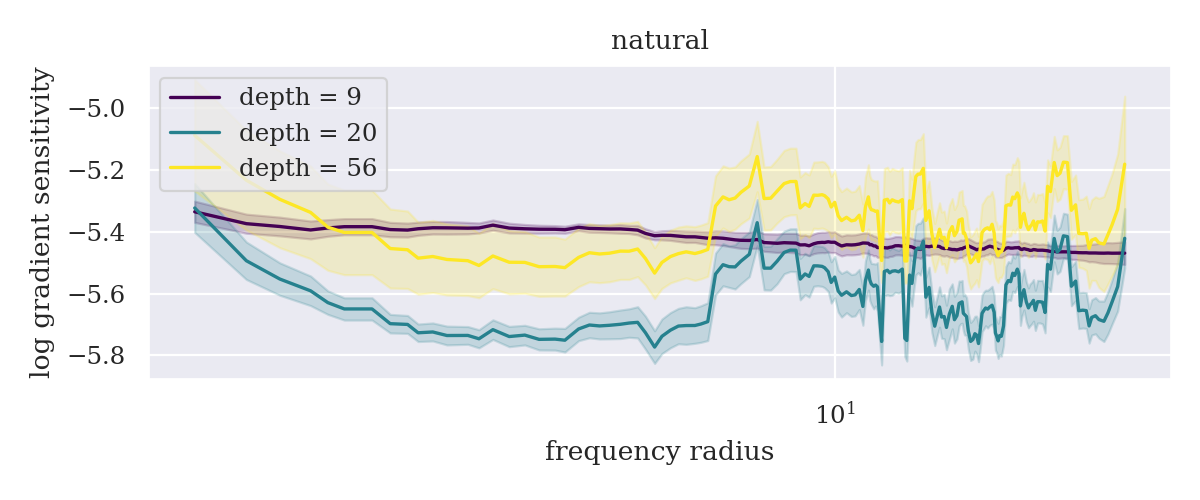}
      \caption{\footnotesize ResNet models of varying depth..}
   \end{subfigure}
   \caption{\footnotesize Radial averages \(E[\nrm{\nabla_{x}f(x)^T \hat{e}_{cij}}\, | \,
   \nrm{(i, j)} = r]\) of frequency sensitivities of randomly initialized CNNs,
   post-processed by smoothing by averaging with \(3\)
   neighbors
   on either side and  taking logarithms. In particular, we do \emph{do not} divide \(E[\nrm{\nabla_{x}f(x)^T \hat{e}_{cij}}\, | \,
   \nrm{(i, j)} = r]\) by its integral with respect to \(r\).}\label{fig:untrained-nets}
\end{figure}

\begin{figure}[!htbp]
   \centering
   \begin{subfigure}{0.45\linewidth}
      \centering
      \includegraphics[width=\linewidth]{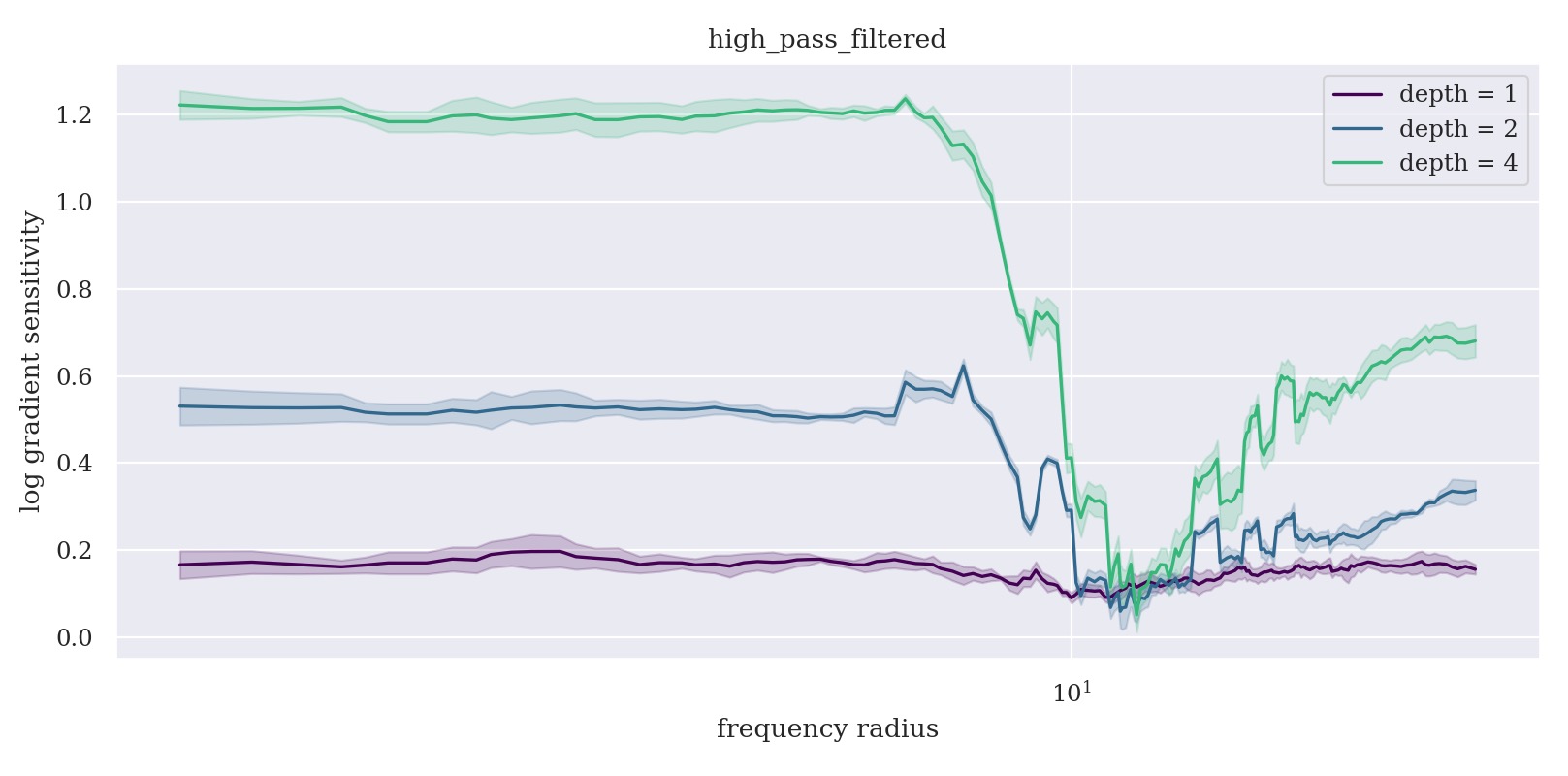}
      \caption{\footnotesize Linear CNNs of varying depth.}
   \end{subfigure}
   \begin{subfigure}{0.45\linewidth}
      \centering
      \includegraphics[width=\linewidth]{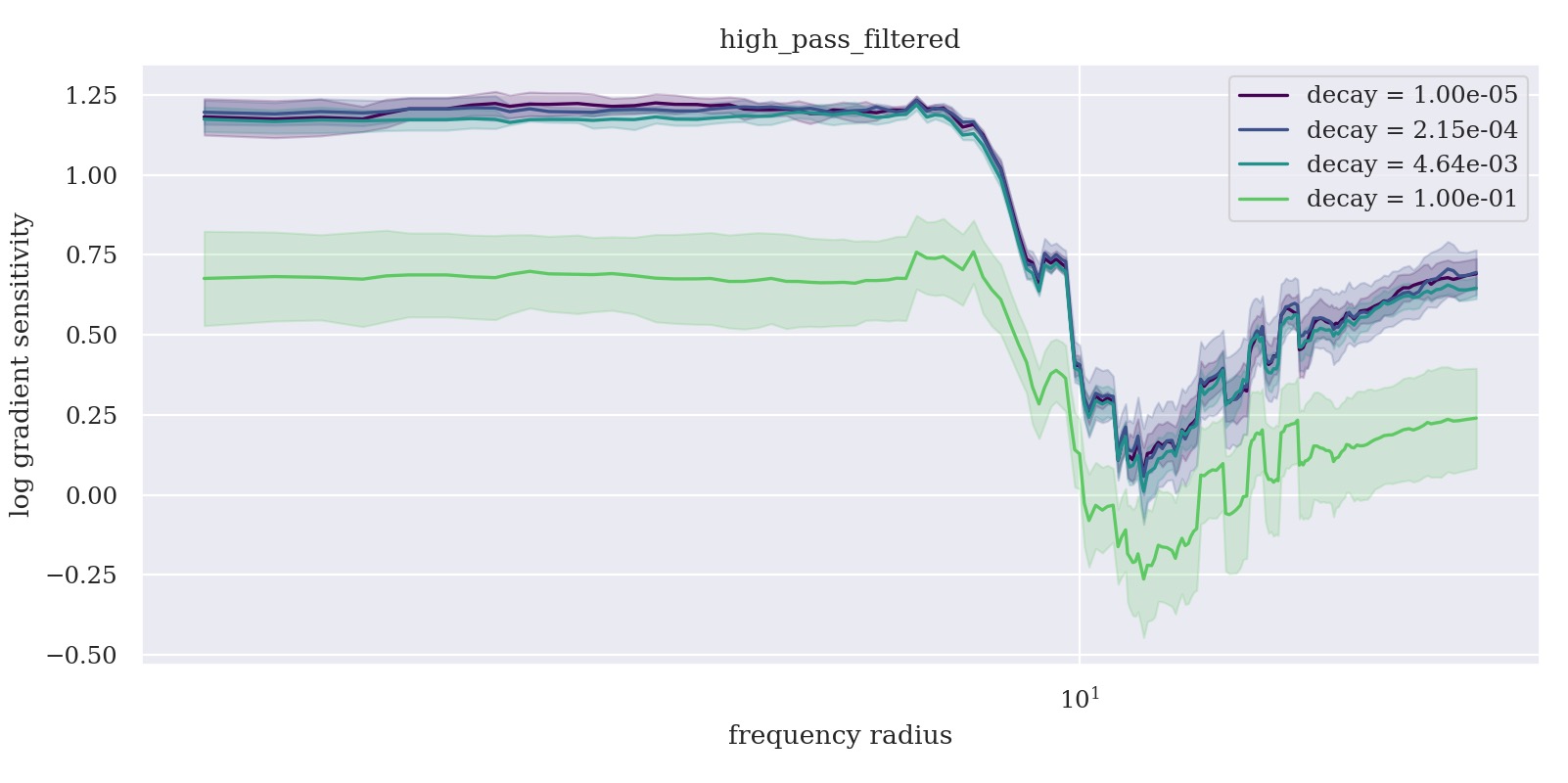}
      \caption{\footnotesize Linear CNNs trained with varying weight decay.}
   \end{subfigure}
   \begin{subfigure}{0.45\linewidth}
      \centering
      \includegraphics[width=\linewidth]{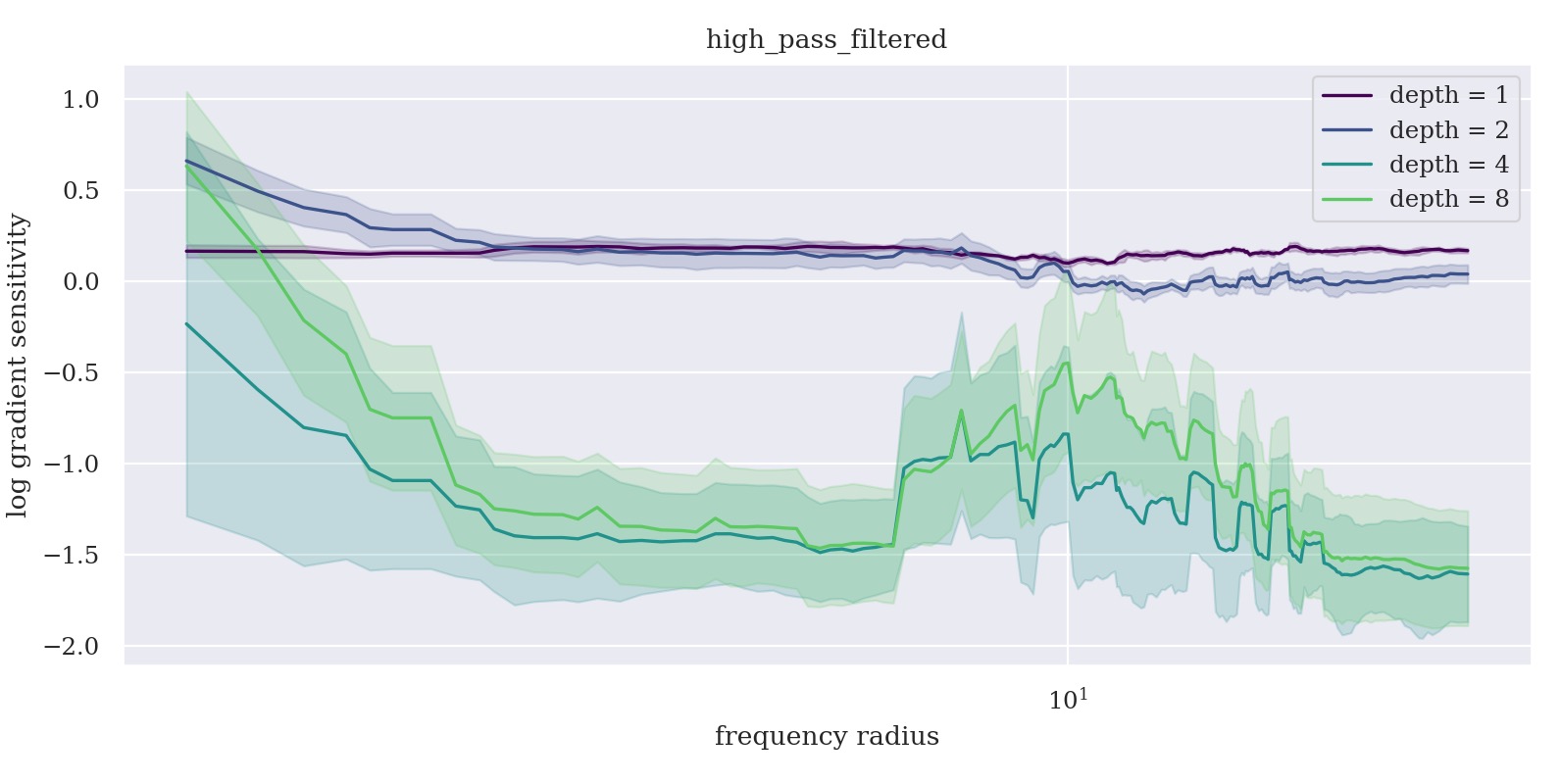}
      \caption{\footnotesize ConvActually models of varying depth.}
   \end{subfigure}
   \begin{subfigure}{0.45\linewidth}
      \centering
      \includegraphics[width=\linewidth]{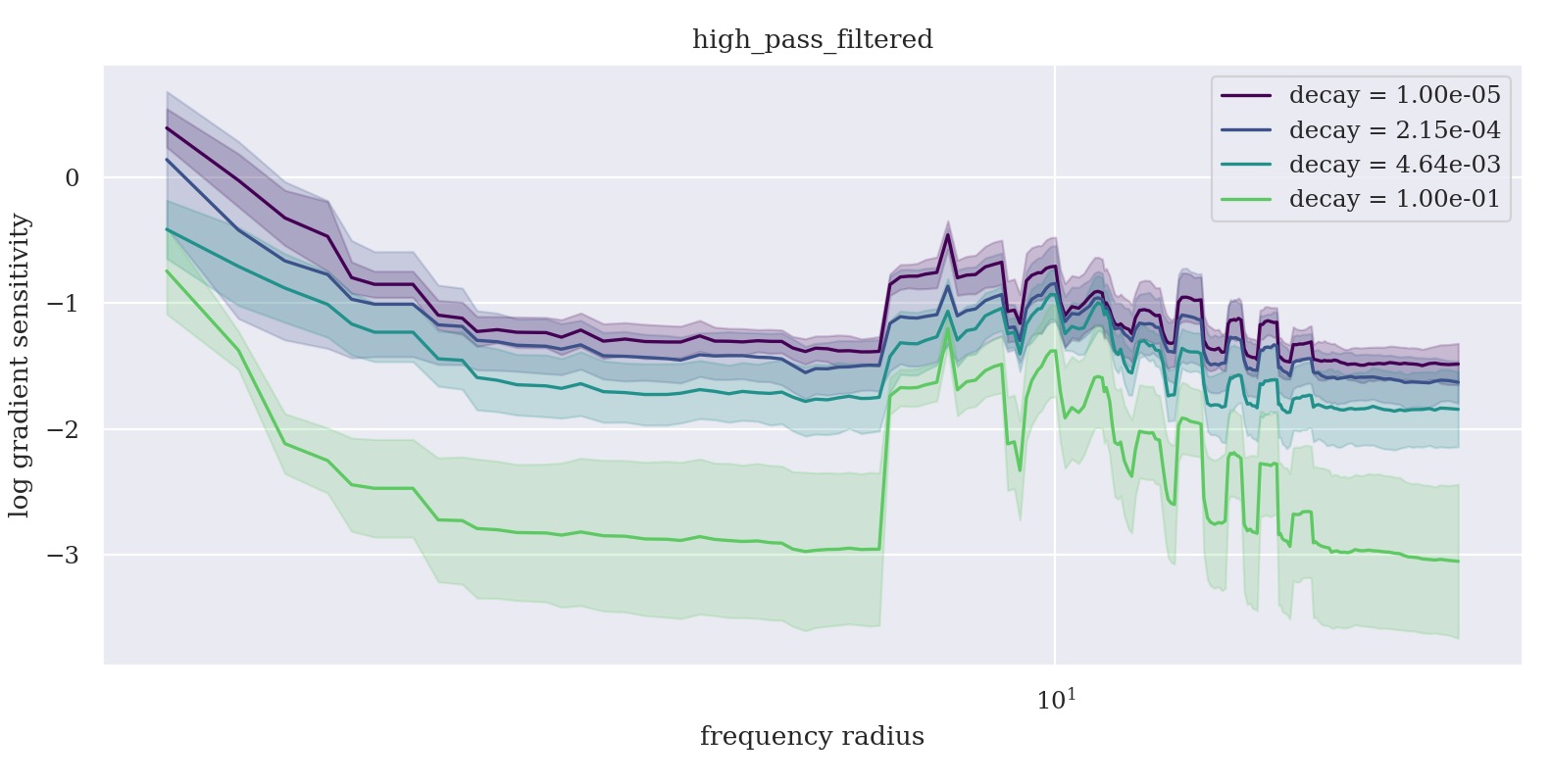}
      \caption{\footnotesize ConvActually models of depth 4 trained with varying weight decay.}
   \end{subfigure}
   \begin{subfigure}{0.45\linewidth}
      \centering
      \includegraphics[width=\linewidth]{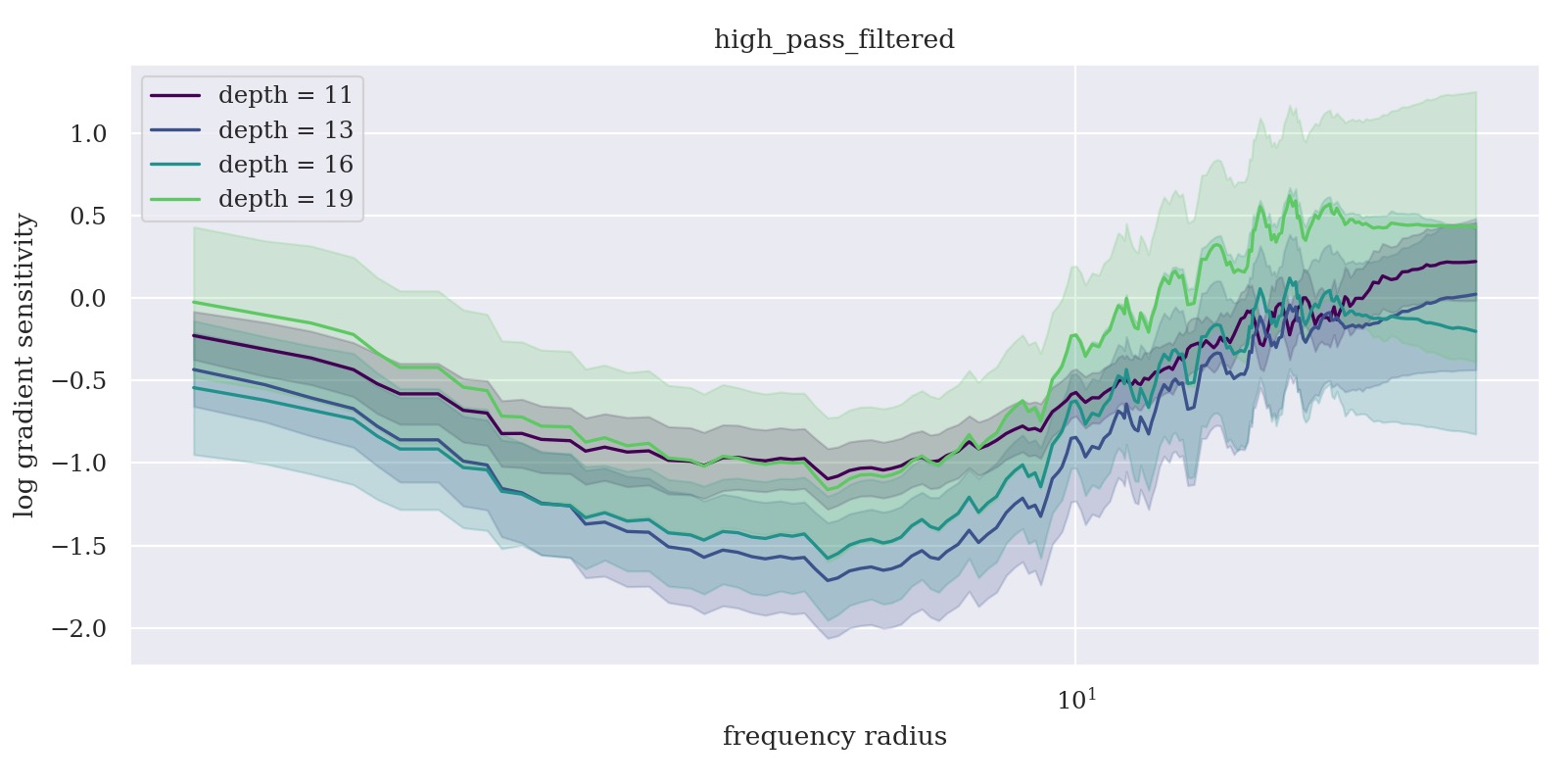}
      \caption{\footnotesize VGG models of varying depth.}
   \end{subfigure}
   \begin{subfigure}{0.45\linewidth}
      \centering
      \includegraphics[width=\linewidth]{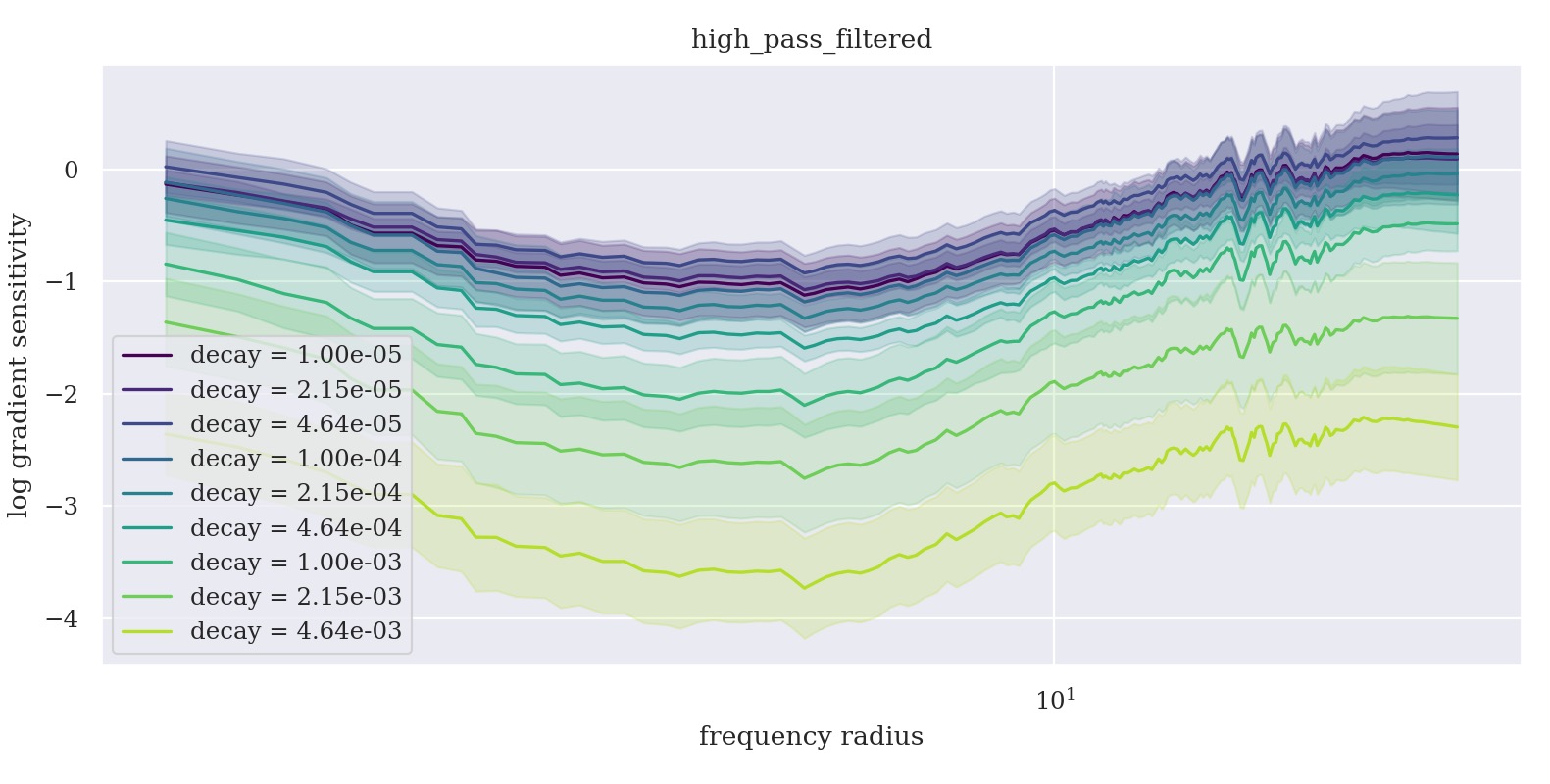}
      \caption{\footnotesize VGG 11 models trained with varying weight decay.}
   \end{subfigure}
   \begin{subfigure}{0.45\linewidth}
      \centering
      \includegraphics[width=\linewidth]{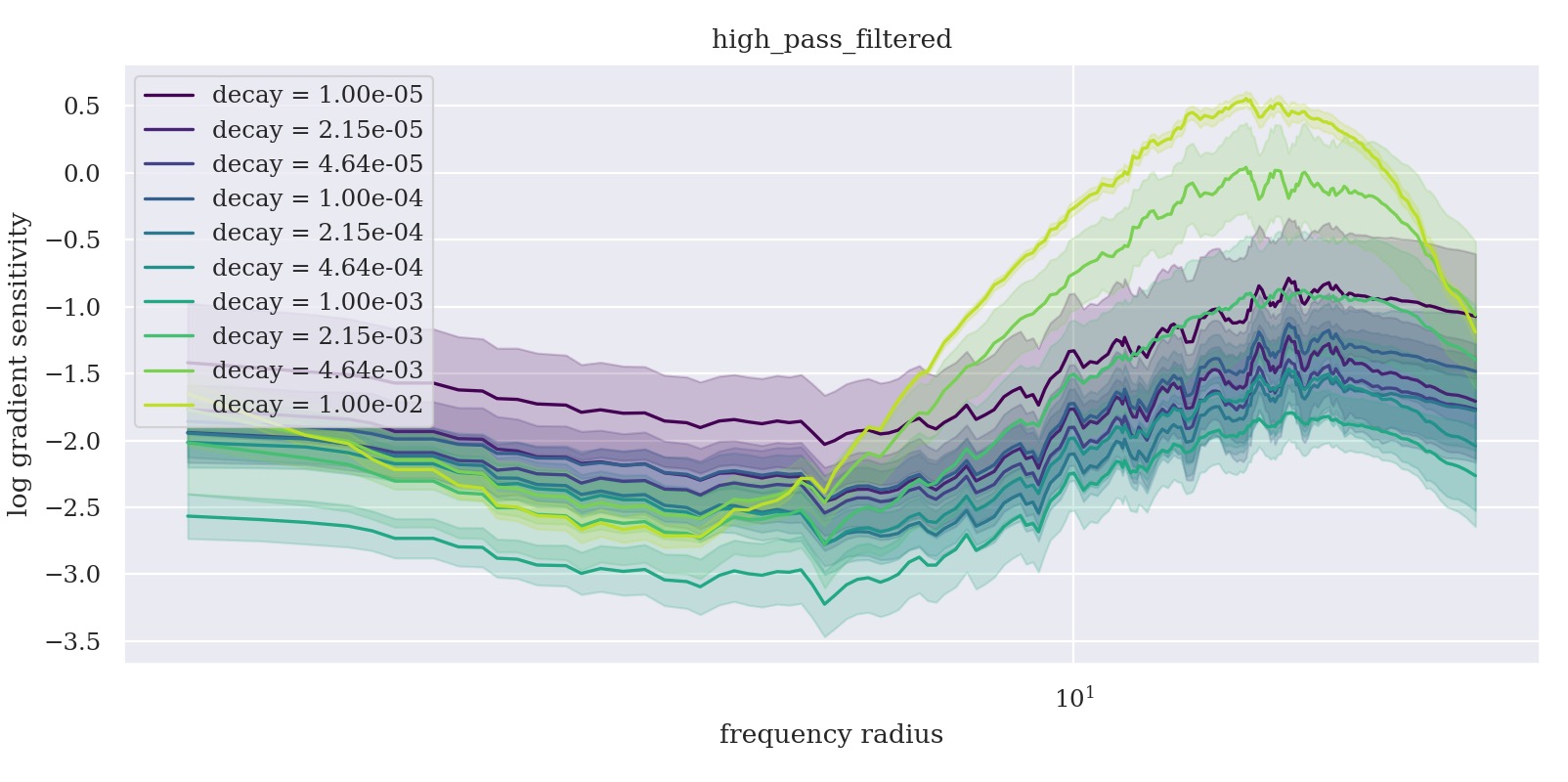}
      \caption{\footnotesize Myrtle CNN models trained with varying weight decay.}
   \end{subfigure}
   \begin{subfigure}{0.45\linewidth}
      \centering
      \includegraphics[width=\linewidth]{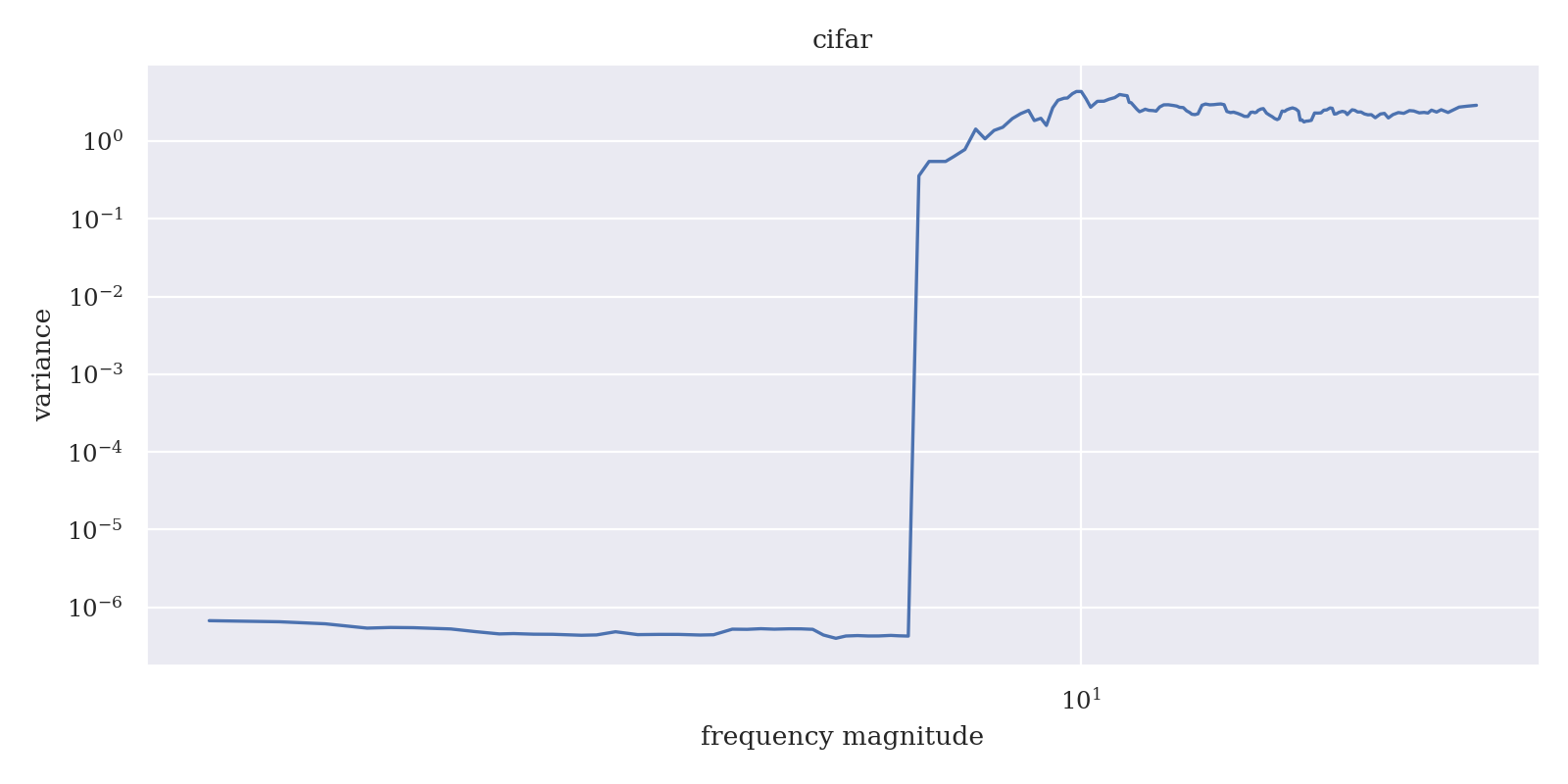}
   \end{subfigure}
   \caption{\footnotesize Radial averages \(E[\nrm{\nabla_{x}f(x)^T \hat{e}_{cij}}\, | \,
   \nrm{(i, j)} = r]\) of frequency sensitivities of CNNs trained on hpf-CIFAR10, post-processed by  smoothing by averaging with \(3\) neighbors
   on either side and taking logarithms. In particular, we do \emph{do not} divide \(E[\nrm{\nabla_{x}f(x)^T \hat{e}_{cij}}\, | \,
   \nrm{(i, j)} = r]\) by its integral with respect to \(r\). \textbf{Bottom right}: frequency statistics of
   hpf-CIFAR10 for comparison.}\label{fig:cifar-experiments-unnormalized}
\end{figure}

To gain some insight into how model frequency sensitivity evolves over the
course of training, in \cref{fig:untrained-nets} we compute frequency
sensitivity curves before training, i.e. at random initialization and omit the
preprocessing step of dividing each curve by its integral. In
\cref{fig:cifar-experiments-unnormalized} we show the unnormalized frequency
sensitivity curves of the same models after training on
hpf-CIFAR10.\footnote{The reason for avoiding normalization is that whereas in
most experiments our concern was \emph{relative} frequency sensitivity of a
given trained model accross frequency bands (i.e., is the given model more
sensitive to low frequency perturbations), in
\cref{fig:untrained-nets,fig:cifar-experiments-unnormalized} we aim to
quantitatively compare low-frequency sensitivity before and after training.} 

Linear CNNs do behave as predicted by the analysis in \cref{sec:sens2low}: in
particular as \cref{eq:princ-comp-converg} suggests their low frequency
sensitivity appears to shrink slowly if at all over the course of training. In
contrast most of the non-linear networks violate the qualitative behavior of
\cref{eq:princ-comp-converg}, with low frequency sensitivity after training on
hpf-CIFAR10 \emph{exceeding} what was present at random initialization.

\subsection{Model training}
\label{sec:model-arch}

In all experiments we optimize using stochastic gradient descent (SGD) with
momentum 0.9, using a ``reduce on plateau'' learning rate schedule where the
learning rate is multiplied by \(0.1\) after ``patience'' epochs without a
\(1\%\) improvement in validation accuracy, where ``patience'' is some fixed
integer (i.e., a hyperparameter); we use patience = \(20\) throughout. This
schedule proceeds until either a minimum learning rate (in all of our
experiments, \(10^{-6}\)) or a maximum number of epochs is hit, at which point
training stops. We use the PyTorch library \cite{torch} on a cluster environment
with Nvidia GPUs. 

All models  are trained on CIFAR10 except the ImageNette VGGs and AlexNets. The
ImageNette VGGs are optimized as above, with the exception that we use
distributed training on 8 GPUs (the batch size of 256 corresponds to a batch of
size 32 on each GPU). The AlexNets are trained using an unsupervised alignment
and uniformity objective as described in \cite[\S 4, \S A]{baradad2021learning};
we use the official implementation of \cite{baradad2021learning}.
For CIFAR10 and ImageNette, we use the same hyperparameters when training on
natural and high pass filtered images.

For each model architecture and choice of hyperparameters,\footnote{\emph{Except
for} the AlexNets trained on WMM data.} and each training
dataset, we train 5 models from different random weight initializations.
Efficiently training this many CNNs was facilitated by the excellent FFCV
library \cite{leclerc2022ffcv}.

Full tables of training hyperparameters, as well as validation accuracies of the
resulting models, are available alongside our code at
\href{github.com/pnnl/frequency_sensitivity/model_details}{github.com/pnnl/frequency\_sensitivity/model\_details}.

 \section{\(L = 2\): LASSO parameter shrinking and selection}
 \label{sec:lasso}
 
 When \(C_l=1\) for all \(l\) and \(L=1\), we may simplify
 \cref{eq:deep-ridge-alt,eq:deep-ridge-alt-tibs} to
 \begin{equation}
    \label{eq:lasso-obj}
    \min_{\hat{v}} \frac{1}{N} \lvert Y - \hat{X} \hat{v} \rvert_2^2 + 2 \lambda \lvert \hat{v} \rvert_1
 \end{equation}
 where \(\hat{X}\) and \(Y \) are as in \cref{sec:1layer}. The optimality
 criterion for \cref{eq:deep-ridge-alt} becomes (see e.g. \cite{tibshiraniSparsityLassoFriendsa})
 \begin{equation}
    \label{eq:lasso-obj-crit}
    \frac{1}{N} (\hat{X}^T \hat{X} \hat{v} - \hat{X}^T Y) + \lambda \nabla \nrm{\hat{v}}_1 = 0,
 \end{equation}
 where \(\nabla \nrm{\hat{v}}_1\) is the \emph{sub-gradient} of the
 \(\ell_1\)-norm:
 \begin{equation}
    \label{eq:subgrad}
    (\nabla \nrm{\hat{v}}_1)_i  \begin{cases}
       = \sign(\hat{v}_i) & \text{  if  } \hat{v}_i \neq 0 \\
       \in [-1,1] & \text{  if  } \hat{v}_i = 0.
    \end{cases}
 \end{equation}
 When \(\lambda = 0\) this again reduces to the unpenalized least squares
 solution \(\hat{v}_{\mathrm{LS}}:= (\hat{X}^T \hat{X})^{-1} \hat{X}^T Y\), and
 substituting this in \cref{eq:lasso-obj-crit} we obtain
 \begin{equation}
    \label{eq:lasso-obj-crit2}
    \frac{1}{N} (\hat{X}^T \hat{X} \hat{v} - \hat{X}^T \hat{X} \hat{v}_{\mathrm{LS}}) + \lambda \nabla \nrm{\hat{v}}_1 = 0.
 \end{equation}
 If we again make the assumption that \(\Sigma = \frac{1}{N}X^T X\) is
 ``diagonal'' as in \cref{eq:diagonality-assumption}, \cref{eq:lasso-obj-crit}
 simplifies to
 \begin{equation}
    \label{eq:lasso-obj-crit3}
    \tau_{cij}(\hat{v}_{cij} - \hat{v}_{\mathrm{LS}, cij}) + \lambda \frac{d \nrm{\hat{v}_{cij}}}{d \hat{v}_{cij}} = 0 \text{  for all  } cij,
 \end{equation}
 where strictly speaking \(\frac{d \nrm{\hat{v}_{cij}}}{d \hat{v}_{cij}}\) is a
 subgradient as in \cref{eq:subgrad}. From this we conclude
 \begin{equation}
    \hat{v}_{cij} \begin{cases}
       = \hat{v}_{\mathrm{LS}, cij} - \frac{\lambda}{\tau_{cij}} \sign(\hat{v}_{cij}), & \text{  if  } \hat{v}_{cij} \neq 0 \\
       \in [\hat{v}_{\mathrm{LS}, cij} -  \frac{\lambda}{\tau_{cij}} , \hat{v}_{\mathrm{LS}, cij} + \frac{\lambda}{\tau_{cij}}  ] & \text{  if  } \hat{v}_{cij} = 0
    \end{cases} \text{  for all  } cij
 \end{equation}
 (the second case is equivalent to: if the LASSO solution \(\hat{v}_{cij} = 0\),
 it must be that the least squares solution satisfies
 \(\nrm{\hat{v}_{\mathrm{LS}, cij}} \leq  \frac{\lambda}{\tau_{cij}} \)). In the
 case where \(\sign(\hat{v}_{cij}) = \sign(\hat{v}_{\mathrm{LS}, cij})\), we
 obtain a particularly nice conclusion:
 \begin{equation}
    \label{eq:particularly-nice}
    \nrm{\hat{v}_{cij}} = \nrm{\hat{v}_{\mathrm{LS}, cij}} -  \frac{\lambda}{\tau_{cij}}  \text{  for all  } cij
 \end{equation}
 \begin{proposition}[{Data-dependent frequency sensitivity, \(L=2\)}]
    \label{prp:data-dep-freq-L1}
    With the notations and assumptions introduced above, the magnitude of the directional
    derivative of \(f\) with respect to the \(cij\)-th Fourier component is
    linear in \( \frac{\lambda}{\tau_{cij}} \) with slope \(-1\).
 \end{proposition}
 If we again plug in the empirically determined power law observed in natural imagery, \(\tau_{cij} \approx
 \frac{\gamma}{\nrm{i}^\alpha + \nrm{j}^\beta}\), \cref{eq:particularly-nice} becomes
 \begin{equation}
    \label{eq:dead-leaves-shrinkage2}
    \nrm{\hat{v}_{cij}} = \nrm{\hat{v}_{\mathrm{LS}, cij}} -  \frac{\lambda}{\gamma} (\nrm{i}^\alpha + \nrm{j}^\beta) \text{  for all  } cij.
 \end{equation}
 Here, the sensitivity decreases monotonically with respect to both frequency
 magnitude and the regularization coefficient \(\lambda\). Note that compared to
 \cref{eq:dead-leaves-shrinkage} from \cref{sec:reg-CNN-training}, \cref{eq:dead-leaves-shrinkage2} implies a greater
 shrinking effect when  \(\nrm{\hat{v}_{cij}} < 1 \), and less shrinking when
 \(\nrm{\hat{v}_{cij}} > 1 \).\footnote{This remark applies generally to LASSO
 vs. ridge regression, and is perhaps most easily explained by comparing the
 derivatives of \(\nrm{x}\) and \(x^2\).} We conjecture that shrinkage due to \(\nrm{\hat{v}_{cij}} < 1 \) is the dominant effect, due to the initial distribution
 of weights \(w\) in modern neural networks, which are often sampled
 from uniform or normal distributions with variance \(\ll 1\) \cite[e.g.,
 according to He initialization;][]{He2016DeepRL}.
 
 \section{Proofs}
 \label{sec:pfs}
 
 \subsection{Proof of \cref{lem:DFTofCNN,lem:bilinearity,lem:parseval-convolution}}
 
 \begin{proof}[Proof of \cref{lem:parseval-convolution}]
    For a direct proof of the convolution part, see \cite[Lem.
    C.2]{kiani2022projunn}.
    
    We reduce to the ``single channel'' cases of these formulae (\(C_l = C_{l-1}
    = 1 \)), which we take to be well known. For the purposes of legibility, in
    this proof we denote the DFT (\cref{eq:DFT,eq:DFT2}) by \(\mathcal{F}\).
 
    For the convolution part we must apply the DFT to \cref{eq:conv-layer}. Let
    \( w_{c, \dots, d} \) denote the single channel tensor obtained by fixing the
    input/output channels to indices \(c, d\) and similarly let \(x_{d, \dots}\)
    be the single channel tensor fixing the input channel to index \(d\). Then by
    interchanging the order of sums 
    \begin{equation}
       \begin{split}
          (w \ast x)_{c i j} &= \sum_{m+m' = i, n+ n' =j} \Big(\sum_{d} w_{cmnd} x_{dm' n'}\Big) \\
          & =   \sum_{d} \Big( \sum_{m+m' = i, n+ n' =j} w_{cmnd} x_{dm' n'}\Big)\\
          &=  \sum_{d} \Big( (w_{c,\dots,d} \ast x_{d,\dots})_{ij}\Big) \\
       \end{split}
    \end{equation} 
    In other words, \( (w \ast x)_{c,\dots} = \sum_{d} w_{c,\dots,d} \ast
    x_{d,\dots} \) Now linearity of the DFT gives 
    \begin{equation}
       \mathcal{F}((w \ast x)_{c,\dots}) = \sum_d \mathcal{F}(w_{c,\dots,d} \ast
       x_{d,\dots}) = \sum_d \mathcal{F}(w_{c,\dots,d} ) \cdot \mathcal{F}( x_{d,\dots})
    \end{equation}
    where on the right hand side we've applied the standard single channel
    convolution theorem. But this means 
    \begin{equation}
       \mathcal{F}(w \ast x)_{cij} = \sum_d \mathcal{F}(w)_{cijd} \mathcal{F}(x)_{dij}
    \end{equation} as desired.
 
    Our proof of the Parseval identity is similar: we start with
    \cref{eq:contraction}, that is, \((w^{T} x)_k = \sum_{l, m, n} w_{kmnl} x_{lmn}\).
    With notation as above, we can write 
    \begin{equation}
       (w^{T} x)_k = \sum_{l, m, n} w_{kmnl} x_{lmn} = \sum_l w_{k,\dots, l}^T x_{l, \dots}
    \end{equation}
    where each term on the right hand side is an ordinary inner product of single
    channel signals. Taking the Parseval identity for these as well known, we get
    \begin{equation}
       (w^{T} x)_k = \sum_l \mathcal{F}(w)_{k,\dots, l}^T \mathcal{F}(x)_{l, \dots} = \sum_{l,m,n} \mathcal{F}(w)_{kmnl}^T \mathcal{F}(x)_{lmn}
    \end{equation}
    as desired.
 \end{proof}
 
 \begin{proof}[Proof of \cref{lem:DFTofCNN,lem:bilinearity}]
    We proceed by induction on \(L\). In the base case where \(  L = 0 \), \cref{lem:DFTofCNN} is equivalent to Parseval's theorem \cref{eq:parseval}. So, suppose \(L>0 \). We may decompose \(f \) as a convolution followed by a linear CNN of the form \cref{eq:simpleCNN}, say \(g \), with one fewer layers. Explicitly, \( g \) has weights \(w_2, \dots, w_{L} \) and
 \begin{equation}
     f(x) = g(w_1 \ast x).
 \end{equation}
 By inductive hypothesis, we may assume that
 \begin{equation}
     g(w_1 \ast x) = \hat{g}(\widehat{w_1 \ast x}) = \hat{w}^{L,T}\big((\prod_{l = 1}^{L-1} \hat{w}_{L-l, \dots ij})\widehat{w_1 \ast x}_{:, ij} \big)
 \end{equation}
 Applying the convolution theorem \cref{eq:convthm} to obtain
 \begin{equation}
     \widehat{w_1 \ast x}_{: ij} = \hat{w}_{1, \dots, ij} \hat{x}_{:, ij}
 \end{equation}
 completes the proof.
 \end{proof}
 
 \begin{proof}[Proof of \cref{lem:bilinearity}]
    For each weight \(\hat{w}^l\) let \(\hat{w}^l_{ij}\) denote the \(C_l \times C_{l-1} \) matrix obtained by fixing the spatial indices of \(\hat{w}^l\)  to \(ij\). Unpacking definitions,
    \begin{equation}
        \begin{split}
            \big( \hat{w}^{L-1} \cdot \hat{w}^{L-1} \cdots \hat{w}^1 \cdot \hat{x} \big)_{ij} &= \hat{w}^{L-1}_{ij} \cdot \hat{w}^{L-1}_{ij} \cdots \hat{w}^1_{ij} \cdot \hat{x}_{ij} \text{ and so } \\
            \hat{w}^{L, T}\big( \hat{w}^{L-1} \cdot \hat{w}^{L-1} \cdots \hat{w}^1 \cdot \hat{x} \big)_c &= \sum_{i, j} \sum_d (\hat{w}^{L}_{ij})_{cd} \big(\hat{w}^{L-1}_{ij} \cdot \hat{w}^{L-1}_{ij} \cdots \hat{w}^1_{ij} \cdot \hat{x}_{ij}\big)_d
        \end{split}
    \end{equation}
    We can recognize the inside sum as performing a matrix product of \( \hat{w}^{L}_{ij} \) with \( \hat{w}^{L-1}_{ij} \cdot \hat{w}^{L-1}_{ij} \cdots \hat{w}^1_{ij} \cdot \hat{x}_{ij} \). Since matrix multiplication is associative, we can just as well multiply the \(\hat{w}^l_{ij}\) first and \emph{then} act on the vector \(\hat{x}_{ij} \). Thus
    \begin{equation}
        \sum_{i, j} \sum_d (\hat{w}^{L}_{ij})_{cd} \big(\hat{w}^{L-1}_{ij} \cdot \hat{w}^{L-1}_{ij} \cdots \hat{w}^1_{ij} \cdot \hat{x}_{ij}\big)_d = \sum_{i, j} \sum_d \big(\hat{w}^{L}_{ij} \cdot \hat{w}^{L-1}_{ij} \cdot \hat{w}^{L-1}_{ij} \cdots \hat{w}^1_{ij} \big)_{cd} \cdot (\hat{x}_{ij})_d
    \end{equation}
    Now we can recognize the right hand side as \( (\hat{w}^{L} \cdot \hat{w}^{L-1} \cdots \hat{w}^1)^T \hat{x} \), as claimed.
 \end{proof}
 
 \subsection{Proofs of \cref{thm:generalized-matrix-holder} and \cref{lem:non-com-generalized-holder}}
 
 Recall we aim to prove:
 the optimization problem
 \begin{equation}
    \label{eq:deep-ridg-obj-recall}
    \min_{\hat{w}} \cL(((\hat{w}^{L} \cdot \hat{w}^{L-1} \cdot \hat{w}^{L-1} \cdots \hat{w}^1)^T \hat{x}^n)_{n=1}^N, (y^n)_{n=1}^N) + \lambda \sum_{l, i, j} \nrm{\hat{w}^l_{ij}}_2^2
 \end{equation}
 is equivalent to the following optimization problem for the product \(\hat{v}_{\dots ij} = \prod_{l = 0}^{L} \hat{w}_{L-l, \dots ij} \):
 \begin{equation}
    \label{eq:deep-ridge-alt-recall}
     \min_{\hat{v}} \cL((\hat{v}^T \hat{x}^n)_{n=1}^N, (y^n)_{n=1}^N) + \lambda L \sum_{i, j}(\nnrm{ \hat{v}_{ij}}_{\frac{2}{L}}^S)^{\frac{2}{L}}.
 \end{equation}
 where the minima runs over the space of tensors \(\hat{v}\) such that each
 matrix \(\hat{v}_{ij}\) has rank at most \(\min\{C, C_1, \dots, C_{L-1}, K\}\).
 We proceed by a series of reductions; as a first step we observe
 \begin{equation}
    \begin{split}
       &\min_{\hat{w}} \cL(((\hat{w}^{L} \cdot \hat{w}^{L-1} \cdot \hat{w}^{L-1} \cdots \hat{w}^1)^T \hat{x}^n)_{n=1}^N, (y^n)_{n=1}^N)  + \lambda \sum_{l, i, j} \nrm{\hat{w}^l_{ij}}_2^2\\
       =& \min_{\hat{v}} \,\, \min_{ \hat{w}^{L} \cdot \hat{w}^{L-1} \cdot \hat{w}^{L-1} \cdots \hat{w}^1 = \hat{v}} \cL((\hat{v}^T \hat{x}^n)_{n=1}^N, (y^n)_{n=1}^N)+ \lambda \sum_{l, i, j} \nrm{\hat{w}^l_{ij}}_2^2,
    \end{split}
 \end{equation}
 and hence \cref{thm:generalized-matrix-holder} will follow if we can prove
 \begin{equation}
    \label{eq:reg-terms}
    \sum_{i, j}(\nnrm{ \hat{v}_{ij}}_{\frac{2}{L}}^S)^{\frac{2}{L}} =
 \min \frac{1}{L} \sum_{l, i, j} \nrm{\hat{w}^l_{ij}}_2^2,
 \end{equation}
 where the \(\min\) on the right hand side runs over all \(\hat{w}\) such that
 \(\hat{w}^{L} \cdot \hat{w}^{L-1} \cdot \hat{w}^{L-1} \cdots \hat{w}^1 = \hat{v}\). We can
 make life slightly simpler by noticing \cref{eq:reg-terms} decomposes over the
 \(i, j\) index, and will follow from
 \begin{equation}
    \label{eq:reg-terms-decomp}
    (\nnrm{ \hat{v}_{ij}}_{\frac{2}{L}}^S)^{\frac{2}{L}} =
 \min \frac{1}{L} \sum_{l} \nrm{\hat{w}^l_{ij}}_2^2 \text{  for all  } i, j,
 \end{equation}
 which we will show in \cref{lem:generalized-srebro} --- note that at this point
 we are arriving at a statement about norms of matrix products and can dispense with the baggage of \(i, j\) indices and \(\hat{}\)s. To formally state that
 lemma, we introduce a convenient definition.
 \begin{definition}
    \label{def:comp-mat}
    A sequence of matrices \(A_1 \in M(m_1 \times n_1, \CC), \dots, A_{L} \in
    M(m_{L} \times n_{L}, \CC)\) is \textbf{composable} if and only if
    \begin{equation}
       m_{l} = n_{l+1} \text{ for } l = 1, \dots, L - 1
    \end{equation}
 \end{definition}
 In other words, \(A_1, \dots, A_{L}\) is composable if and only if the product
 \(A_{L} \cdots A_1\) makes sense.
 \begin{lemma}
    \label{lem:generalized-srebro}
    If \(B \in M(m \times n, \CC)\) is a matrix with complex entries and \(L \in
    \NN\) is a non-negative integer, then
    \begin{equation}
       \label{eq:srebro-thing}
       (\nnrm{B}_{\frac{2}{L}}^S)^{\frac{2}{L}} =  \min (\prod_l \nrm{A_l}_2^2)^{\frac{1}{L}}
        = \min \frac{1}{L}\sum_{l} \nrm{A_l}_2^2,
    \end{equation}
    where both minima are taken over all composable sequences of complex matrices
    \(A_1, \dots, A_{L}\) such that \(A_{L} \cdots A_1 = B\).
 \end{lemma}
 We first deal with the elementary aspects of \cref{lem:generalized-srebro}: it
 will suffice to show that whenever \(A_{L} \cdots A_1 = B\),
 \begin{equation}
    \label{eq:str-ineq}
    (\nnrm{B}_{\frac{2}{L}}^S)^{\frac{2}{L}} \leq   (\prod_l \nrm{A_l}_2^2)^{\frac{1}{L}}
        \leq \frac{1}{L}\sum_{l} \nrm{A_l}_2^2 \text{  \emph{and that}  }
 \end{equation}
 \begin{equation}
    \label{eq:suitable-choice}
    (\nnrm{B}_{\frac{2}{L}}^S)^{\frac{2}{L}} = \frac{1}{L}\sum_{l} \nrm{A_l}_2^2
 \end{equation}
 for \emph{some}  composable sequence \( A_1, \dots, A_{L} \)  such that \(
 A_{L} \cdots A_1 = B \). As noted in \cite{Gunasekar2018} the second inequality of \cref{eq:str-ineq} is
 simply the arithmetic-geometric mean inequality applied to \(\nrm{A_1}_2^2,
 \dots \nrm{A_{L}}_2^2 \in \RR_{\geq 0}\). Furthermore, we can obtain
 \cref{eq:suitable-choice} using the singular value decomposition of \(B\): let
 \(B = USV^{\ast}\) where, letting \(r := \min\{m, n\}\), \(U \in U(r, m)\) and
 \(V \in U(n, r)\) are unitary and \(S =
 \diag(\lambda_1, \dots, \lambda_r)\), where \(\lambda_i \geq 0 \) for all \(i\).
 We may decompose \(B\) into \(L\) factors like
 \begin{equation}
    \label{eq:suitable-choice2}
    B = (U S^{\frac{1}{L}}) \cdot (\prod_{i = 1}^{L-1} S^{\frac{1}{L}}) \cdot (S^{\frac{1}{L}} V^\ast),
 \end{equation}
 and by unitary invariance of the \(C_2\) (a.k.a. Frobenius) norm,
 \begin{equation}
    \label{eq:suitable-choice3}
    \begin{split}
       &\nrm{U S^{\frac{1}{L}}}_2^2 +\sum_{i = 1}^{L-1} \nrm{S^{\frac{1}{L}}}_2^2 + \nrm{S^{\frac{1}{L}} V^\ast}_2^2 =\nrm{S^{\frac{1}{L}}}_2^2 +\sum_{i = 1}^{L-1} \nrm{S^{\frac{1}{L}}}_2^2 + \nrm{S^{\frac{1}{L}}}_2^2 \\
       &= L \nrm{S^{\frac{1}{L}}}_2^2.
    \end{split}
 \end{equation}
 We now note that \emph{by \cref{def:cp-norm}},
 \begin{equation}
    \label{eq:suitable-choice4}
    \begin{split}
       (\nnrm{B}_{\frac{2}{L}}^S)^{\frac{2}{L}} &= \sum_i \nrm{\lambda_i}^{\frac{2}{L}} \text{ and on the other hand  } \\
       \nrm{S^{\frac{1}{L}}}_2^2 &= \sum_i \nrm{\lambda_i}^{\frac{1}{L} \cdot 2},
    \end{split}
 \end{equation}
 so that combining \cref{eq:suitable-choice3,eq:suitable-choice4} gives
 \cref{eq:suitable-choice} for the composable sequence of
 \cref{eq:suitable-choice2}.
 
 It remains to prove the first inequality of \cref{eq:str-ineq} --- this is a special case of the non-commutative generalized H\"older inequality
 \cref{lem:non-com-generalized-holder}. In fact we will prove a slightly more
 general statement --- to state it we need a couple more definitions:
 \begin{definition}
    For a (not necessarily square) complex matrix \( A \in M(m \times n, \CC)\),
    \begin{equation}
        \nrm{A} := \sqrt{A^\ast A}
    \end{equation}
    where \(A^\ast \) is the conjugate transpose of \(A\).
 \end{definition}
 The matrix \(\nrm{A}\) is Hermitian and positive semi-definite, and often
 referred to as the \emph{polar} part of \(A\).\footnote{E.g. in the polar
 decomposition \cite[\S 8]{Higham2008}.} Given any Hermitian matrix \(H \) and
 complex number \(z \), we may form the matrix \(H^z \); explicitly it can be
 defined as \(U \diag(\lambda_i^z ) U^\ast\) where \(H = U \diag(\lambda_i)
 U^\ast \) is a diagonalization of \(H\). We next define unitary invariant norms
 on spaces of matrices. For technical reasons to be encountered shortly, we
 actually  introduce \emph{families of norms} compatible with the natural
 inclusions \(M(m \times n, \CC) \subseteq M(m' \times n', \CC)\) for \(m' \geq
 m, n'\geq n\).
 \begin{definition}[{cf. \cite[\S IV]{bhatiaMatrixAnalysis1996}}]
    A \textbf{compatible family of matrix norms} is a function 
    \begin{equation}
       \nnrm{-}: \coprod_{m, n \in \NN} M(m\times n, \CC) \to \RR_{\geq 0}
    \end{equation}
    such that \begin{enumerate}[(i)]
       \item the restriction of \(\nnrm{-}\) to \(M(m\times n, \CC)\) is a norm
       (in the sense of functional analysis) for all \(m, n \in \NN\), and 
       \item whenever \(m' \geq m, n'\geq n\) and \(\iota: M(m \times n, \CC) \subseteq
       M(m' \times n', \CC)\) is the ``upper left block inclusion'' sending 
       \begin{equation}
          A \mapsto \begin{bmatrix}
             A & 0 \\
             0 & 0 \\
          \end{bmatrix}
       \end{equation}
       we have \(\nnrm{\iota(A)} = \nnrm{A}\).
    \end{enumerate}
    A compatible family of norms is \textbf{unitary
    invariant} if and only if for any \( A \in M(m \times n, \CC) \) and any
    unitary matrices \(U \in U(m) \) and \(V \in U(n) \)
    \begin{equation}
       \nnrm{U A V^T} = \nnrm{A}.
    \end{equation}
 \end{definition}
 We will show below that for any \(p > 0\) the Schatten \(p\)-norms form a
 unitary invariant compatible family.
 \begin{lemma}
    \label{lem:non-com-generalized-holder-recall}
    If \(B \in M(m \times n, \CC)\) is a matrix with complex entries, \(A_1,
    \dots, A_{L}\)  is a composable sequence of complex matrices such that
    \(A_L \cdots A_1 = B\)  and \(\sum_i
    \frac{1}{p_1} = \frac{1}{r}\) where \(p_1,
    \dots, p_{L} >0\) are positive real numbers,  then for every unitary
    invariant compatible family of norms \( \nnrm{-}\),
    \begin{equation}
        \nnrm{\nrm{B}^r}^{\frac{1}{r}} \leq \prod_i \nnrm{\nrm{A_i}^{p_i}}^{\frac{1}{p^i}}.
    \end{equation}
  \end{lemma}
 We note that this  non-commutative generalized H\"older inequality is
 ``non-commutative'' since we work with products of matrices as opposed to inner
 products of vectors, and ``generalized'' since we consider \(\ell_p\) exponents
 \(p_i\) where \(\sum \frac{1}{p_i} > 1\). To be specific,
 \cref{lem:non-com-generalized-holder} is the case of
 \cref{lem:non-com-generalized-holder-recall} where the unitary invariant
 compatible family of norms 
 \(\nnrm{-}\) consists of the \(C_1 \) norms\footnote{Which is simply the sum of the
 singular values (sometimes called the \emph{trace norm}).} and \(p_i = 2\) for
 all \(i\). Here we are implicitly using the relationship between Schatten norms
 \begin{equation}
    \nnrm{A}_p^S = (\nnrm{\nrm{A}^p}_1^S)^{\frac{1}{p}} \text{  for all  } A \in M(m \times n, \CC), p>0.
 \end{equation}
 As mentioned in \cref{sec:reg-CNN-training}, when \( L=2 \)
 \cref{lem:non-com-generalized-holder-recall} is \cite[Exercise
 IV.2.7]{bhatiaMatrixAnalysis1996}. In the case of Schatten norms, it is
 essentially derived in the course of the proof of \cite[Thm.
 1]{dai2021representation}. Below, we solve \cite[Exercise
 IV.2.7]{bhatiaMatrixAnalysis1996} and show that
 the case \(L>2\) follows by induction.\footnote{We beg forgiveness for posting a
 Springer GTM exercise solution on the internet.}
 
 We first address a subtle difference between
 \cref{lem:non-com-generalized-holder-recall} and the setup of
 \cite{bhatiaMatrixAnalysis1996} (and indeed most work on matrix analysis).
 \cite{bhatiaMatrixAnalysis1996} considers square matrices throughout, whereas in
 \cref{lem:non-com-generalized-holder-recall} we allow all matrices to be
 rectangular --- this is essential in our applications, since neural networks
 have variable width (even if we made our CNNs of \cref{sec:DFT-CNN} ``constant
 width'' by requiring \(C_1 = C_2 = \dots = C_{L-1}\), the first and last layers
 would still change width in general). Thankfully, there is a simple trick that
 allows us to reduce to the case of square matrices. Say \(A_l \in M(m_l \times
 n_l)\) as in \cref{def:comp-mat}, and let
 \begin{equation}
    N := \max \{m_1, \dots, m_L, n_1, \dots, n_L \}
 \end{equation}
 Since \(N \geq m_l, n_l\) by definition, for each matrix \(A_l\) we may define a
 new block diagonal matrix
 \begin{equation}
    \tilde{A}_l := \begin{bmatrix}
       A_l & 0 \\
       0 & 0
    \end{bmatrix}
 \end{equation}
 (that is, we push \(A_l\) to the top left corner). Note that by
 \cref{def:comp-mat} and the hypotheses of
 \cref{lem:non-com-generalized-holder-recall} \(m = m_L, n= n_1\) and a
 straightforward inspection of the mechanics of block diagonal matrix
 multiplication shows
 \begin{equation}
    \tilde{A}_L \cdot \tilde{A}_{L-1} \cdots \tilde{A}_1 = \begin{bmatrix}
       B & 0 \\
       0 & 0 \\
    \end{bmatrix} =: \tilde{B}
 \end{equation}
 Now \(\tilde{A}_1, \dots, \tilde{A}_L \) and \(\tilde{B} \) are all square, and
 to reduce to the square case it will suffice to argue that 
 \begin{equation}
    \nnrm{\nrm{\tilde{A}_i}^{p_i}} = \nnrm{\nrm{A_i}^{p_i}}^{\frac{1}{p^i}} \text{ for all \(i\) and }  \nnrm{\nrm{\tilde{B}}^r} = \nnrm{\nrm{B}^r}
 \end{equation}
 \emph{Since by hypothesis \(\nnrm{-}\) is a compatible family of unitary invariant
 norms}, it will suffice to show that 
 \begin{equation}
    \nrm{\tilde{A}_i}^{p_i} = \begin{bmatrix}
       \nrm{A_i}^{p_i} & 0 \\
       0 & 0 \\
    \end{bmatrix}
 \end{equation}
 and this is follows from the identities 
 \begin{equation}
    \begin{bmatrix}
       A & 0 \\
       0 & 0 \\
    \end{bmatrix}^\ast
    \begin{bmatrix}
       A & 0 \\
       0 & 0 \\
    \end{bmatrix}
    =\begin{bmatrix}
       A^\ast A & 0 \\
       0 & 0 \\
    \end{bmatrix} \text{  and  }
    \begin{bmatrix}
       H & 0 \\
       0 & 0 \\
    \end{bmatrix}^z 
    = \begin{bmatrix}
       H^z & 0 \\
       0 & 0 \\
    \end{bmatrix}
 \end{equation}
 valid for any complex matrix \(A\), Hermitian matrix \(H\) and complex number
 \(z\), the proofs of which we omit. 
 
 Before continuing with the proof of
 \cref{lem:non-com-generalized-holder-recall}, we pause to verify that Schatten
 norms are indeed a unitary invariant compatible family. In doing so we prove a
 lemma that will be of further use in the sequel.
 
 \begin{definition}
    A \textbf{compatible family of guage functions} is a function \(\Phi:
    \coprod_n \RR^n \to \RR_{\geq 0}\) such that 
    \begin{enumerate}[(i)]
       \item for each \(n \in \NN\) the restriction of \(\Phi\) to \(\RR^n\) is a
       norm (in the sense of functional analysis) and 
       \item whenever \(n' \geq n\) and \(\iota: \RR^n \to \RR^{n'}\) is the
       inclusion mapping \((x_1, \dots, x_n) \mapsto (x_1, \dots, x_n, 0,\dots,
       0)\), we have \(\Phi(\iota(x)) = \Phi(x)\).
    \end{enumerate}
    A \textbf{compatible family of guage functions} is \emph{symmetric} if and
    only if for each \(n \in \NN\) the restriction of \(\Phi\) to \(\RR^n\) is
    invariant under the action of matrices of the form \(P D \) where \(P\) is a
    permutation and \(D\) is diagonal, with diagonal entries in \(\{\pm 1\}\). 
 \end{definition}
 
 \begin{lemma}[{cf. \cite[Thm. IV.2.1]{bhatiaMatrixAnalysis1996}}]
    \label{lem:nrm-guage}
    There is a natural one-to-one correspondence between unitary invariant
    compatible families of matrix norms and symmetric compatible families of
    guage functions.
 \end{lemma}
 \begin{proof}
    Given a unitary invariant compatible family of matrix norms \(\nnrm{-}\), a
    symmetric compatible family of guage functions \(\Phi\) can be defined using
    the maps 
    \begin{equation}
       \RR^n \xrightarrow{\diag} M(n\times n, \CC) \xrightarrow{\nnrm{-}} \RR_{\geq 0};
    \end{equation}
    compatibility of \(\Phi\) comes from compatibility of \(\nnrm{-}\) and the
     identity ``\(\diag\circ\iota = \iota \circ \diag\)'' (suitably interpreted),
     and symmetry of \(\Phi\) follows from the identity \(\diag(PD x) =
     (PD)\diag(x)(PD)^\ast\) for matrices \(PD\) as above, the fact such matrices
     \(PD\) are unitary and unitary invariance of \(\nnrm{-}\).
 
     Conversely, given a symmetric compatible family of guage functions \(\Phi\)
     one may define a unitary invariant compatible family of matrix norms
     \(\nnrm{-}\) as \(\nnrm{A} := \Phi(s(A))\)  where \(s(A)\) denotes the
     singular values of \(A\). Compatibility comes from the fact that the
     singular values of a block diagonal matrix 
     \[\begin{bmatrix}
       A_1 & 0 \\
       0 & A_2
     \end{bmatrix}\]
     are the concatenation of \(s(A_1)\) and \(s(A_2)\), and unitary invariance
     follows from the fact that singular values unitary invariant up to permutations
     (and \(\Phi\) is symmetric). The proof that the maps \(\nnrm{-}: M(m \times
     n, \CC) \to \RR_\geq 0\) are indeed \emph{norms} is as in \cite[Thm. IV.2.1]{bhatiaMatrixAnalysis1996}.
 
     It can be verified that these maps are mutual inverses --- we omit this
     final step.
 \end{proof}
 
 \begin{corollary}
    For any \(p>0\) the Schatten \(p\)-norms form a unitary invariant
    compatible family of matrix norms.
 \end{corollary}
 \begin{proof}
    By \emph{definition}, \(\nnrm{A}_p^S = (\sum_i s(A)_i^p)^{\frac{1}{p}}\),
    i.e. the \(\ell_p\) norm of the singular values of \(A\). By
    \cref{lem:nrm-guage} it suffices to show that the \(\ell_p\)-norms form a
    symmetric compatible family --- this is straightforward and omitted.
 \end{proof}
 
 We now resume proving \cref{lem:non-com-generalized-holder-recall}, first in the
 case \(L=2\) (later we will prove the general case by induction). Recall that at this
 point we have reduced to the case where all matrices in sight are \(n\times n\)
 for some fixed \(n\in \NN\), so in particular we are dealing with a fixed
 unitary invariant norm (no further need for compatible families). By the above
 lemma, \(\nnrm{A} = \Phi(s(A))\) for some symmetric  guage
 function \(\Phi\), for all \(A \in M(n \times n, \CC)\). By \cite[Thm.
 IV.2.5]{bhatiaMatrixAnalysis1996}, 
 \begin{equation}
    s(A_1A_2)^r <_w s(A_1)^r s(A_2)^r
 \end{equation}
 where \(<_w\) denotes weak submajorization. Now the ``strongly isotone''
 property of the symmetric guage function \(\Phi\) implies
 \begin{equation}
    \label{eq:isotone-thing}
    \Phi(s(A_1A_2)^r) \leq \Phi(s(A_1)^r s(A_2)^r)
 \end{equation}
 We need a generalized H\"older inequality for symmetric guage functions. 
 \begin{lemma}[{\cite[Ex. IV.1.7]{bhatiaMatrixAnalysis1996}}]
    \label{lem:gen-hold-sym-guage}
    If \(p_1, p_2 >0\) and \(\frac{1}{p_1} + \frac{1}{p_2} = \frac{1}{r}\)  then for
    every symmetric guage function \(\Phi: \RR^n \to \RR_{\geq 0}\) and every
    \(x, y \in \RR^n\) we have 
    \begin{equation}
       \Phi(\mathrm{abs}(x \cdot y)^r)^{\frac{1}{r}} \leq \Phi(\mathrm{abs}(x)^{p_1})^{\frac{1}{p_1}}\Phi(\mathrm{abs}(y)^{p_2})^{\frac{1}{p_2}}
    \end{equation}
    where \(\mathrm{abs}\) denotes the coordinatewise absolute value.
 \end{lemma}
 \begin{proof}
    Apply the regular \(r=1\) H\"older inequality \cite[Thm.
    IV.1.6]{bhatiaMatrixAnalysis1996} to the vectors \(\tilde{x} = \mathrm{abs}(x)^r,
    \tilde{y}=\mathrm{abs}(y)^r\) (coordinatewise \(r\)-th powers) with the exponents
    \(\tilde{p_1} = p_1/r, \tilde{p_2} = p_2/r\) (note that \(\frac{1}{\tilde{p_1}} +
    \frac{1}{\tilde{p_2}} = \frac{r}{p_1} +\frac{r}{p_2} = 1\)) to obtain
    \begin{equation}
       \Phi(\mathrm{abs}(x \cdot y)^r) = \Phi(\mathrm{abs}(\tilde{x} \cdot \tilde{y})) 
       \leq \Phi(\mathrm{abs}(\tilde{x})^{\tilde{p_1}})^{\frac{1}{\tilde{p_1}}} 
       \Phi(\mathrm{abs}(\tilde{y})^{\tilde{p_2}})^{\frac{1}{\tilde{p_2}}}
       = \Phi(\mathrm{abs}(x)^{p_1})^{\frac{r}{p_1}}\Phi(\mathrm{abs}(y)^{p_2})^{\frac{r}{p_2}}
    \end{equation}
    where in the last equality we have just used the definitions of \(\tilde{x},
    \tilde{y}, \tilde{p_1}\) and \(\tilde{p_2}\). Taking \(r\)-th roots completes the
    proof.
 \end{proof}
 Now applying \cref{lem:gen-hold-sym-guage} to \cref{eq:isotone-thing} gives
 \begin{equation}
    \Phi(s(A_1A_2)^r)^{\frac{1}{r}} \leq \Phi(s(A_1)^r s(A_2)^r)^{\frac{1}{r}} \leq \Phi(s(A_1)^{p_1})^{\frac{1}{p_1}}\Phi(s(A_2)^{p_2})^{\frac{1}{p_2}}
 \end{equation}
 (the first inequality is just taking \(r\)-th roots of \cref{eq:isotone-thing},
 the second is applying \cref{lem:gen-hold-sym-guage}). Using the identity
 \(s(\nrm{A}^r) = s(A)^r\), we finally obtain
 \begin{equation}
    \nnrm{\nrm{A_1 A_2}^r}^{\frac{1}{r}} \leq \nnrm{\nrm{A_1}^{p_1}}^{\frac{1}{p_1}} \nnrm{\nrm{A_2}^{p_2}}^{\frac{1}{p_2}}
 \end{equation}
 which is \cref{lem:non-com-generalized-holder-recall} when \(L=2\).
 
 Now suppose \(L > 2 \) and assume by inductive hypothesis that
 \cref{lem:non-com-generalized-holder-recall} holds for all smaller values of
 \(L\). Define 
 \begin{equation}
    \frac{1}{q} = \sum_{i=1}^{L-1} \frac{1}{p_i}
 \end{equation}
 (note that \(\frac{1}{p_L} + \frac{1}{q} = \frac{1}{r}\)). By the \( L= 2\) case of \cref{lem:non-com-generalized-holder-recall} 
 \begin{equation}
    \nnrm{\nrm{A_L \cdot (A_{L-1}\cdots A_1)}^r}^{\frac{1}{r}} \leq \nnrm{\nrm{A_L}^{p_L}}^{\frac{1}{p_L}} \nnrm{\nrm{A_{L-1}\cdots A_1}^q}^{\frac{1}{q}}
 \end{equation}
 and by inductive hypothesis
 \begin{equation}
    \nnrm{\nrm{A_{L-1}\cdots A_1}^q}^{\frac{1}{q}} \leq \prod{i=1}^{L-1} \nnrm{\nrm{A_i}^{p_i}}^{\frac{1}{p_i}}.
 \end{equation}

%% file: freq_sens.bib
@inproceedings{abello2021,
  author    = {Abello, Antonio A. and Hirata, Roberto and Wang, Zhangyang},
  booktitle = {2021 IEEE/CVF Conference on Computer Vision and Pattern Recognition Workshops (CVPRW)},
  title     = {Dissecting the High-Frequency Bias in Convolutional Neural Networks},
  year      = {2021},
  volume    = {},
  number    = {},
  pages     = {863-871},
  doi       = {10.1109/CVPRW53098.2021.00096}
}

@inproceedings{baradad2021learning,
  title     = {Learning to See by Looking at Noise},
  author    = {Manel Baradad and Jonas Wulff and Tongzhou Wang and Phillip Isola and Antonio Torralba},
  booktitle = {Advances in Neural Information Processing Systems},
  editor    = {A. Beygelzimer and Y. Dauphin and P. Liang and J. Wortman Vaughan},
  year      = {2021},
  url       = {https://openreview.net/forum?id=RQUl8gZnN7O}
}

@inproceedings{berhnard2021,
  author    = {Bernhard, Rémi and Moëllic, Pierre-Alain
               and Mermillod, Martial and Bourrier, Yannick and Cohendet, Romain and Solinas,
               Miguel and Reyboz, Marina},
  booktitle = {2021 International Joint Conference on
               Neural Networks (IJCNN)},
  title     = {Impact of Spatial Frequency Based Constraints
               on Adversarial Robustness},
  year      = {2021},
  volume    = {},
  number    = {},
  pages     = {1-8},
  doi       = {10.1109/IJCNN52387.2021.9534307}
}

@article{dixmierFormesLineairesAnneau1953a,
  title   = {{Formes lin\'eaires sur un anneau d'op\'erateurs}},
  author  = {Dixmier, Jacques},
  year    = {1953},
  journal = {Bulletin de la Soci\'et\'e Math\'ematique de France},
  volume  = {81},
  pages   = {9--39},
  doi     = {10.24033/bsmf.1436},
  langid  = {french}
}

@inproceedings{geirhos2018imagenettrained,
  title     = {ImageNet-trained {CNN}s are biased towards texture; increasing shape bias improves accuracy and robustness.},
  author    = {Robert Geirhos and Patricia Rubisch and Claudio Michaelis and Matthias Bethge and Felix A. Wichmann and Wieland Brendel},
  booktitle = {International Conference on Learning Representations},
  year      = {2019},
  url       = {https://openreview.net/forum?id=Bygh9j09KX}
}

@book{Goodfellow-et-al-2016,
  title     = {Deep Learning},
  author    = {Ian Goodfellow and Yoshua Bengio and Aaron Courville},
  publisher = {MIT Press},
  note      = {\url{http://www.deeplearningbook.org}},
  year      = {2016}
}

@inproceedings{guoLowFrequencyAdversarial2019,
  title     = {Low {{Frequency Adversarial Perturbation}}},
  booktitle = {{{UAI}}},
  author    = {Guo, Chuan and Frank, Jared S. and Weinberger, Kilian Q.},
  year      = {2019}
}

@article{He2016DeepRL,
  title   = {Deep Residual Learning for Image Recognition},
  author  = {Kaiming He and X. Zhang and Shaoqing Ren and Jian Sun},
  journal = {2016 IEEE Conference on Computer Vision and Pattern Recognition (CVPR)},
  year    = {2016},
  pages   = {770-778}
}

@book{Higham2008,
  author    = {Nicholas J. Higham},
  title     = {Functions of Matrices: {Theory} and Computation},
  publisher = {Society for Industrial and Applied Mathematics},
  address   = {Philadelphia, PA, USA},
  year      = {2008},
  pages     = {xx+425},
  isbn      = {978-0-898716-46-7}
}

@article{leeOcclusionModelsNatural2001,
  title      = {Occlusion {{Models}} for {{Natural Images}}: {{A Statistical Study}} of a {{Scale-Invariant Dead Leaves Model}}},
  shorttitle = {Occlusion {{Models}} for {{Natural Images}}},
  author     = {Lee, Ann B. and Mumford, David and Huang, Jinggang},
  year       = {2001},
  month      = jan,
  journal    = {International Journal of Computer Vision},
  volume     = {41},
  number     = {1},
  pages      = {35--59},
  issn       = {1573-1405},
  doi        = {10.1023/A:1011109015675},
  langid     = {english}
}

@article{maiya2022a,
  title  = {A Frequency Perspective of Adversarial Robustness},
  author = {Shishira Maiya and Max Ehrlich and Vatsal Agarwal and Ser-Nam Lim and Tom Goldstein and Abhinav Shrivastava},
  year   = {2022},
  url    = {https://openreview.net/forum?id=7gRvcAulxa}
}

@conference{mathieu2014,
  title    = {Fast training of convolutional networks through FFTS: International Conference on Learning Representations (ICLR2014), CBLS, April 2014},
  author   = {Michael Mathieu and Mikael Henaff and Yann LeCun},
  year     = {2014},
  month    = jan,
  day      = {1},
  language = {English (US)},
  note     = {2nd International Conference on Learning Representations, ICLR 2014 ; Conference date: 14-04-2014 Through 16-04-2014}
}

@online{pageHowTrainYour2018,
  title        = {How to {{Train Your ResNet}}},
  author       = {Page, David},
  date         = {2018-09-24T17:11:20+00:00},
  url          = {https://myrtle.ai/learn/how-to-train-your-resnet/},
  urldate      = {2022-05-09},
  langid       = {american},
  organization = {{Myrtle}}
}

@inproceedings{pmlr-v70-chen17d,
  title     = {Strong {NP}-Hardness for Sparse Optimization with Concave Penalty Functions},
  author    = {Yichen Chen and Dongdong Ge and Mengdi Wang and Zizhuo Wang and Yinyu Ye and Hao Yin},
  booktitle = {Proceedings of the 34th International Conference on Machine Learning},
  pages     = {740--747},
  year      = {2017},
  editor    = {Precup, Doina and Teh, Yee Whye},
  volume    = {70},
  series    = {Proceedings of Machine Learning Research},
  month     = {06--11 Aug},
  publisher = {PMLR},
  url       = {https://proceedings.mlr.press/v70/chen17d.html}
}

@inproceedings{Pratt2017FCNNFC,
  title     = {FCNN: Fourier Convolutional Neural Networks},
  author    = {Harry Pratt and Bryan M. Williams and Frans Coenen and Yalin Zheng},
  booktitle = {ECML/PKDD},
  year      = {2017}
}

@article{sharmaEffectivenessLowFrequency2019,
  title        = {On the {{Effectiveness}} of {{Low Frequency Perturbations}}},
  author       = {Sharma, Yash and Ding, G. and Brubaker, Marcus A.},
  year         = {2019},
  journaltitle = {IJCAI},
  doi          = {10.24963/ijcai.2019/470}
}

@article{tibshiraniEquivalencesSparseModels,
  title    = {Equivalences {{Between Sparse Models}} and {{Neural Networks}}},
  author   = {Tibshirani, Ryan J},
  pages    = {8},
  url      = {https://www.stat.cmu.edu/~ryantibs/papers/sparsitynn.pdf},
  year     = {2021},
  langid   = {english}
}

@article{tibshiraniSparsityLassoFriendsa,
  title  = {Sparsity, the {{Lasso}}, and {{Friends}}},
  author = {Tibshirani, Ryan and Wasserman, Larry},
  pages  = {34},
  langid = {english}
}

@conference{vasilache2015,
  title    = {Fast convolutional nets with fbfft: A GPU performance evaluation},
  author   = {Nicolas Vasilache and Jeff Johnson and Michael Mathieu and Soumith Chintala and Serkan Piantino and Yann LeCun},
  year     = {2015},
  month    = jan,
  day      = {1},
  language = {English (US)},
  note     = {3rd International Conference on Learning Representations, ICLR 2015 ; Conference date: 07-05-2015 Through 09-05-2015}
}

@inproceedings{yin2019,
  author    = {Dong Yin and
               Raphael Gontijo Lopes and
               Jonathon Shlens and
               Ekin Dogus Cubuk and
               Justin Gilmer},
  editor    = {Hanna M. Wallach and
               Hugo Larochelle and
               Alina Beygelzimer and
               Florence d'Alch{\'{e}}{-}Buc and
               Emily B. Fox and
               Roman Garnett},
  title     = {A Fourier Perspective on Model Robustness in Computer Vision},
  booktitle = {Advances in Neural Information Processing Systems 32: Annual Conference
               on Neural Information Processing Systems 2019, NeurIPS 2019, December
               8-14, 2019, Vancouver, BC, Canada},
  pages     = {13255--13265},
  year      = {2019},
  url       = {https://proceedings.neurips.cc/paper/2019/hash/b05b57f6add810d3b7490866d74c0053-Abstract.html},
  bibsource = {dblp computer science bibliography, https://dblp.org}
}

@article{Zhu2021GoingDI,
  title   = {Going Deeper in Frequency Convolutional Neural Network: A Theoretical Perspective},
  author  = {Xiaohan Zhu and Zhen Cui and Tong Zhang and Yong Li and Jian Yang},
  journal = {ArXiv},
  year    = {2021},
  volume  = {abs/2108.05690}
}

@inproceedings{Gunasekar2018,
  title = {Implicit Bias of Gradient Descent on Linear Convolutional Networks},
  booktitle = {Advances in Neural Information Processing Systems},
  author = {Gunasekar, Suriya and Lee, Jason D and Soudry, Daniel and Srebro, Nati},
  editor = {Bengio, S. and Wallach, H. and Larochelle, H. and Grauman, K. and {Cesa-Bianchi}, N. and Garnett, R.},
  year = {2018},
  volume = {31},
  publisher = {{Curran Associates, Inc.}}
}

@inproceedings{Hastie2001TheEO,
  title={The Elements of Statistical Learning: Data Mining, Inference, and Prediction, 2nd Edition},
  author={Trevor J. Hastie and Robert Tibshirani and Jerome H. Friedman},
  booktitle={Springer Series in Statistics},
  year={2001}
}

@inproceedings{lawrenceImplicitBiasLinear2022,
  title = {Implicit {{Bias}} of {{Linear Equivariant Networks}}},
  booktitle = {Proceedings of the 39th {{International Conference}} on {{Machine Learning}}},
  author = {Lawrence, Hannah and Kiani, Bobak and Georgiev, Kristian G. and Dienes, Andrew K.},
  year = {2022},
  month = jun,
  pages = {12096--12125},
  publisher = {{PMLR}},
  issn = {2640-3498},
  langid = {english}
}

@inproceedings{
hendrycks2018benchmarking,
title={Benchmarking Neural Network Robustness to Common Corruptions and Perturbations},
author={Dan Hendrycks and Thomas Dietterich},
booktitle={International Conference on Learning Representations},
year={2019},
url={https://openreview.net/forum?id=HJz6tiCqYm},
}

@InProceedings{pmlr-v97-recht19a,
  title = 	 {Do {I}mage{N}et Classifiers Generalize to {I}mage{N}et?},
  author =       {Recht, Benjamin and Roelofs, Rebecca and Schmidt, Ludwig and Shankar, Vaishaal},
  booktitle = 	 {Proceedings of the 36th International Conference on Machine Learning},
  pages = 	 {5389--5400},
  year = 	 {2019},
  editor = 	 {Chaudhuri, Kamalika and Salakhutdinov, Ruslan},
  volume = 	 {97},
  series = 	 {Proceedings of Machine Learning Research},
  month = 	 {09--15 Jun},
  publisher =    {PMLR},
  url = 	 {https://proceedings.mlr.press/v97/recht19a.html}
}

@misc{mosaicml2022composer,
    author = {The Mosaic ML Team},
    title = {composer},
    year = {2021},
    howpublished = {\url{https://github.com/mosaicml/composer/}},
}

@inproceedings{yunUnifyingViewImplicit2021,
  title = {A Unifying View on Implicit Bias in Training Linear Neural Networks},
  booktitle = {International {{Conference}} on {{Learning Representations}}},
  author = {Yun, Chulhee and Krishnan, Shankar and Mobahi, Hossein},
  year = {2021},
  month = mar,
  langid = {english}
}

@misc{lyuGradientDescentMaximizes2020b,
  title = {Gradient {{Descent Maximizes}} the {{Margin}} of {{Homogeneous Neural Networks}}},
  author = {Lyu, Kaifeng and Li, Jian},
  year = {2020},
  month = dec,
  number = {arXiv:1906.05890},
  eprint = {1906.05890},
  eprinttype = {arxiv},
  primaryclass = {cs, stat},
  publisher = {{arXiv}},
  archiveprefix = {arXiv}
}

@InProceedings{pmlr-v178-xiao22a,
  title = 	 {Eigenspace Restructuring: A Principle of Space and Frequency in Neural Networks},
  author =       {Xiao, Lechao},
  booktitle = 	 {Proceedings of Thirty Fifth Conference on Learning Theory},
  pages = 	 {4888--4944},
  year = 	 {2022},
  editor = 	 {Loh, Po-Ling and Raginsky, Maxim},
  volume = 	 {178},
  series = 	 {Proceedings of Machine Learning Research},
  month = 	 {02--05 Jul},
  publisher =    {PMLR},
  url = 	 {https://proceedings.mlr.press/v178/xiao22a.html}
}

@misc{xiaoLearningCurves,
  doi = {10.48550/ARXIV.2205.14846},
  
  url = {https://arxiv.org/abs/2205.14846},
  
  author = {Xiao, Lechao and Pennington, Jeffrey},
  
  title = {Precise Learning Curves and Higher-Order Scaling Limits for Dot Product Kernel Regression},
  
  publisher = {arXiv},
  
  year = {2022},
  
  copyright = {Creative Commons Attribution 4.0 International}
}

@inproceedings{
dai2021representation,
title={Representation Costs of Linear Neural Networks: Analysis and Design},
author={Zhen Dai and Mina Karzand and Nathan Srebro},
booktitle={Advances in Neural Information Processing Systems},
editor={A. Beygelzimer and Y. Dauphin and P. Liang and J. Wortman Vaughan},
year={2021},
url={https://openreview.net/forum?id=3oQyjABdbC8}
}

@InProceedings{pmlr-v178-jagadeesan22a,
  title = 	 {Inductive Bias of Multi-Channel Linear Convolutional Networks with Bounded Weight Norm},
  author =       {Jagadeesan, Meena and Razenshteyn, Ilya and Gunasekar, Suriya},
  booktitle = 	 {Proceedings of Thirty Fifth Conference on Learning Theory},
  pages = 	 {2276--2325},
  year = 	 {2022},
  editor = 	 {Loh, Po-Ling and Raginsky, Maxim},
  volume = 	 {178},
  series = 	 {Proceedings of Machine Learning Research},
  month = 	 {02--05 Jul},
  publisher =    {PMLR},
  url = 	 {https://proceedings.mlr.press/v178/jagadeesan22a.html}
}

@book{bhatiaMatrixAnalysis1996,
  title = {Matrix {{Analysis}}},
  author = {Bhatia, Rajendra},
  year = {1996},
  month = nov,
  publisher = {{Springer Science \& Business Media}},
  googlebooks = {F4hRy1F1M6QC},
  isbn = {978-0-387-94846-1},
  langid = {english}
}

@inproceedings{torchvision,
author = {Marcel, S\'{e}bastien and Rodriguez, Yann},
title = {Torchvision the Machine-Vision Package of Torch},
year = {2010},
isbn = {9781605589336},
publisher = {Association for Computing Machinery},
address = {New York, NY, USA},
url = {https://doi.org/10.1145/1873951.1874254},
doi = {10.1145/1873951.1874254},
booktitle = {Proceedings of the 18th ACM International Conference on Multimedia},
pages = {1485–1488},
numpages = {4},
location = {Firenze, Italy},
series = {MM '10}
}

@TECHREPORT{Krizhevsky09learningmultiple,
    author = {Alex Krizhevsky},
    title = {Learning multiple layers of features from tiny images},
    institution = {},
    year = {2009}
}

@inproceedings{deng2009imagenet,
  title={Imagenet: A large-scale hierarchical image database},
  author={Deng, Jia and Dong, Wei and Socher, Richard and Li, Li-Jia and Li, Kai and Fei-Fei, Li},
  booktitle={2009 IEEE conference on computer vision and pattern recognition},
  pages={248--255},
  year={2009},
  organization={Ieee}
}

@InProceedings{Simonyan15,
  author       = "Karen Simonyan and Andrew Zisserman",
  title        = "Very Deep Convolutional Networks for Large-Scale Image Recognition",
  booktitle    = "International Conference on Learning Representations",
  year         = "2015",
}

@incollection{torch,
title = {PyTorch: An Imperative Style, High-Performance Deep Learning Library},
author = {Paszke, Adam and Gross, Sam and Massa, Francisco and Lerer, Adam and Bradbury, James and Chanan, Gregory and Killeen, Trevor and Lin, Zeming and Gimelshein, Natalia and Antiga, Luca and Desmaison, Alban and Kopf, Andreas and Yang, Edward and DeVito, Zachary and Raison, Martin and Tejani, Alykhan and Chilamkurthy, Sasank and Steiner, Benoit and Fang, Lu and Bai, Junjie and Chintala, Soumith},
booktitle = {Advances in Neural Information Processing Systems 32},
editor = {H. Wallach and H. Larochelle and A. Beygelzimer and F. d\textquotesingle Alch\'{e}-Buc and E. Fox and R. Garnett},
pages = {8024--8035},
year = {2019},
publisher = {Curran Associates, Inc.},
url = {http://papers.neurips.cc/paper/9015-pytorch-an-imperative-style-high-performance-deep-learning-library.pdf}
}

@article{hacohen,
  author  = {Guy Hacohen and Daphna Weinshall},
  title   = {Principal Components Bias in Over-parameterized Linear Models, and its Manifestation in Deep Neural Networks},
  journal = {Journal of Machine Learning Research},
  year    = {2022},
  volume  = {23},
  number  = {155},
  pages   = {1--46},
  url     = {http://jmlr.org/papers/v23/21-0991.html}
}

@inproceedings{Rahaman2019OnTS,
  title={On the Spectral Bias of Neural Networks},
  author={Nasim Rahaman and Aristide Baratin and Devansh Arpit and Felix Dr{\"a}xler and Min Lin and Fred A. Hamprecht and Yoshua Bengio and Aaron C. Courville},
  booktitle={ICML},
  year={2019}
}

@article{jo,
  author    = {Jason Jo and
               Yoshua Bengio},
  title     = {Measuring the tendency of CNNs to Learn Surface Statistical Regularities},
  journal   = {CoRR},
  volume    = {abs/1711.11561},
  year      = {2017},
  url       = {http://arxiv.org/abs/1711.11561},
  eprinttype = {arXiv},
  eprint    = {1711.11561},
  bibsource = {dblp computer science bibliography, https://dblp.org}
}

@inproceedings{
kiani2022projunn,
title={proj{UNN}: efficient method for training deep networks with unitary matrices},
author={Bobak Kiani and Randall Balestriero and Yann LeCun and Seth Lloyd},
booktitle={Advances in Neural Information Processing Systems},
editor={Alice H. Oh and Alekh Agarwal and Danielle Belgrave and Kyunghyun Cho},
year={2022},
url={https://openreview.net/forum?id=nEJMdZd8cIi}
}

@misc{leclerc2022ffcv,
    author = {Guillaume Leclerc and Andrew Ilyas and Logan Engstrom and Sung Min Park and Hadi Salman and Aleksander Madry},
    title = {ffcv},
    year = {2022},
    howpublished = {\url{https://github.com/libffcv/ffcv/}},
    note = {commit xxxxxxx}
}

@InProceedings{pmlr-v119-wang20k,
  title = 	 {Understanding Contrastive Representation Learning through Alignment and Uniformity on the Hypersphere},
  author =       {Wang, Tongzhou and Isola, Phillip},
  booktitle = 	 {Proceedings of the 37th International Conference on Machine Learning},
  pages = 	 {9929--9939},
  year = 	 {2020},
  editor = 	 {III, Hal Daumé and Singh, Aarti},
  volume = 	 {119},
  series = 	 {Proceedings of Machine Learning Research},
  month = 	 {13--18 Jul},
  publisher =    {PMLR},
  url = 	 {https://proceedings.mlr.press/v119/wang20k.html}
}

@misc{vggcifar,
  author = {Fu, Cheng-Yang},
  title = {pytorch-vgg-cifar10},
  year = {2017},
  publisher = {GitHub},
  journal = {GitHub repository},
  howpublished = {\url{https://github.com/chengyangfu/pytorch-vgg-cifar10}},
}

@misc{imagenette,
  author = {FastAI},
  title = {imagenette},
  year = {2019},
  publisher = {GitHub},
  journal = {GitHub repository},
  howpublished = {\url{https://github.com/fastai/imagenette}},
}

@inproceedings{fouriermix,
  author    = {Jiachen Sun and
               Akshay Mehra and
               Bhavya Kailkhura and
               Pin{-}Yu Chen and
               Dan Hendrycks and
               Jihun Hamm and
               Z. Morley Mao},
  editor    = {Shai Avidan and
               Gabriel J. Brostow and
               Moustapha Ciss{\'{e}} and
               Giovanni Maria Farinella and
               Tal Hassner},
  title     = {A Spectral View of Randomized Smoothing Under Common Corruptions:
               Benchmarking and Improving Certified Robustness},
  booktitle = {Computer Vision - {ECCV} 2022 - 17th European Conference, Tel Aviv,
               Israel, October 23-27, 2022, Proceedings, Part {IV}},
  series    = {Lecture Notes in Computer Science},
  volume    = {13664},
  pages     = {654--671},
  publisher = {Springer},
  year      = {2022},
  url       = {https://doi.org/10.1007/978-3-031-19772-7\_38},
  doi       = {10.1007/978-3-031-19772-7\_38},
  bibsource = {dblp computer science bibliography, https://dblp.org}
}

@inproceedings{
diffenderfer2021a,
title={A Winning Hand: Compressing Deep Networks Can Improve Out-of-Distribution Robustness},
author={James Diffenderfer and Brian R Bartoldson and Shreya Chaganti and Jize Zhang and Bhavya Kailkhura},
booktitle={Advances in Neural Information Processing Systems},
editor={A. Beygelzimer and Y. Dauphin and P. Liang and J. Wortman Vaughan},
year={2021},
url={https://openreview.net/forum?id=YygA0yppTR}
}

@article{fukushimaNeocognitronSelforganizingNeural1980,
  title = {Neocognitron: {{A}} Self-Organizing Neural Network Model for a Mechanism of Pattern Recognition Unaffected by Shift in Position},
  shorttitle = {Neocognitron},
  author = {Fukushima, Kunihiko},
  year = {1980},
  month = apr,
  journal = {Biological Cybernetics},
  volume = {36},
  number = {4},
  pages = {193--202},
  issn = {1432-0770},
  doi = {10.1007/BF00344251},
  langid = {english}
}

@article{collobertWav2LetterEndtoEndConvNetbased2016,
  title = {{{Wav2Letter}}: An {{End-to-End ConvNet-based Speech Recognition System}}},
  shorttitle = {{{Wav2Letter}}},
  author = {Collobert, Ronan and Puhrsch, Christian and Synnaeve, Gabriel},
  year = {2016},
  month = sep,
  journal = {arXiv:1609.03193 [cs]},
  eprint = {1609.03193},
  eprinttype = {arxiv},
  primaryclass = {cs},
  archiveprefix = {arXiv}
}

@inproceedings{krizhevskyImageNetClassificationDeep2012,
  title = {{{ImageNet Classification}} with {{Deep Convolutional Neural Networks}}},
  booktitle = {Advances in {{Neural Information Processing Systems}}},
  author = {Krizhevsky, Alex and Sutskever, Ilya and Hinton, Geoffrey E},
  year = {2012},
  volume = {25},
  publisher = {{Curran Associates, Inc.}}
}

@article{lecunBackpropagationAppliedHandwritten1989,
  title = {Backpropagation {{Applied}} to {{Handwritten Zip Code Recognition}}},
  author = {LeCun, Y. and Boser, B. and Denker, J. S. and Henderson, D. and Howard, R. E. and Hubbard, W. and Jackel, L. D.},
  year = {1989},
  month = dec,
  journal = {Neural Computation},
  volume = {1},
  number = {4},
  pages = {541--551},
  issn = {0899-7667},
  doi = {10.1162/neco.1989.1.4.541}
}

@inproceedings{karpathyLargescaleVideoClassification2014,
  title = {Large-Scale {{Video Classification}} with {{Convolutional Neural Networks}}},
  booktitle = {Proceedings of the {{IEEE Conference}} on {{Computer Vision}} and {{Pattern Recognition}}},
  author = {Karpathy, Andrej and Toderici, George and Shetty, Sanketh and Leung, Thomas and Sukthankar, Rahul and {Fei-Fei}, Li},
  year = {2014},
  pages = {1725--1732}
}
